\documentclass[11pt,letterpaper]{article}
\pdfoutput=1 
\usepackage{comment}
\usepackage[height=8in]{geometry}
\usepackage{amsmath, amssymb, amsthm, mathabx}
\usepackage{graphicx, color}
\usepackage{array, multirow}
\usepackage[font=small,labelfont=bf]{caption} 
\usepackage{lipsum}
\usepackage{amsfonts}
\usepackage{tabularx}
\usepackage{booktabs}
\usepackage{graphicx}
\usepackage{epstopdf}
\usepackage{algorithmic}
\usepackage{mathtools}
\usepackage{tikz}
\ifpdf
  \DeclareGraphicsExtensions{.eps,.pdf,.png,.jpg}
\else
  \DeclareGraphicsExtensions{.eps}
\fi

\AtBeginDocument{%
	\setlength{\oddsidemargin}{\dimexpr(\paperwidth-\textwidth)/2-1in}%
	\setlength{\evensidemargin}{\oddsidemargin}%
	\setlength{\topmargin}{%
		\dimexpr(\paperheight-\textheight)/2-\headheight-\headsep-1in}%
}

\usepackage[shortlabels]{enumitem}

\usepackage[sort]{natbib}

\setcitestyle{numbers,square,comma}
\usepackage{hyperref}
\hypersetup{colorlinks=true,
			urlcolor=blue,
			linkcolor=blue,
			citecolor=blue,
			bookmarksdepth=paragraph}
\usepackage[nameinlink]{cleveref}
\crefname{equation}{}{}
\crefname{section}{section}{sections}
\crefname{figure}{figure}{figures}
\crefname{table}{table}{tables}
\crefname{example}{example}{examples}
\crefname{proposition}{proposition}{propositions}
\Crefname{section}{Section}{Sections}
\Crefname{figure}{Figure}{Figures}
\Crefname{table}{Table}{Tables}
\Crefname{definition}{Definition}{Definitions}
\Crefname{theorem}{Theorem}{Theorems}
\Crefname{remark}{Remark}{Remarks}
\Crefname{example}{Example}{Examples}
\Crefname{proposition}{Proposition}{Propositions}
\numberwithin{equation}{section}

\newtheorem{theorem}{Theorem}[section]
\newtheorem{lemma}{Lemma}[section]

\newtheorem{proposition}{Proposition}[section]
\theoremstyle{plain}
\newtheorem{definition}{Definition}[section]
\newtheorem{remark}{Remark}[section]

\newtheorem{innercustomgeneric}{\customgenericname}
\providecommand{\customgenericname}{}
\newcommand{\newcustomtheorem}[2]{%
  \newenvironment{#1}[1]
  {%
   \renewcommand\customgenericname{#2}%
   \renewcommand\theinnercustomgeneric{##1}%
   \innercustomgeneric
  }
  {\endinnercustomgeneric}
}

\newcustomtheorem{customthm}{Theorem}

\NewDocumentCommand{\Martina}{mo}{
    \IfValueF{#2}{
                        {{
                            \textcolor{red}{ 
                            \textbf{M:}
                            \textit{{#1}}
                            }
                        }}
        }
    \IfValueT{#2}{
                        \marginnote{{\scriptsize
                            \textcolor{red}{ 
                            \textbf{M:}
                            \textit{{#1}}
                            }
                        }}
        }
                    }

\DeclareMathOperator*{\sinc}{sinc}

\title{Theoretical guarantees for the advantage of GNNs over NNs in generalizing bandlimited functions on Euclidean cubes}

\author{A. Martina Neuman$^1$, Rongrong Wang $^2$, Yuying Xie $^3$}

\author{A. Martina Neuman\thanks{Faculty of Mathematics, University of Vienna, Kolingasse 14, 1090 Vienna, Austria; E-mail: \texttt{neumana53@univie.ac.at}}, Rongrong Wang\thanks{Department of Computational Mathematics Science and Engineering and Department of Mathematics, Michigan State University, 428 S Shaw Lane, East Lansing, 48824, Michigan, USA; E-mail: \texttt{wangron6@msu.edu}}, and Yuying Xie \thanks{Department of Computational Mathematics Science and Engineering and Department of Statistics and Probability, Michigan State University, 428 S Shaw Lane, East Lansing, 48824, Michigan, USA; E-mail: \texttt{xyy@msu.edu}}.}

\date{}

\begin{document}

\maketitle

\begin{abstract}
Graph Neural Networks (GNNs) have emerged as formidable resources for processing graph-based information across diverse applications. While the expressive power of GNNs has traditionally been examined in the context of graph-level tasks, their potential for node-level tasks, such as node classification, where the goal is to interpolate missing node labels from the observed ones, remains relatively unexplored. In this study, we investigate the proficiency of GNNs for such classifications, which can also be cast as a function interpolation problem. Explicitly, we focus on ascertaining the optimal configuration of weights and layers required for a GNN to successfully interpolate a band-limited function over Euclidean cubes. Our findings highlight a pronounced efficiency in utilizing GNNs to generalize a bandlimited function within an $\varepsilon$-error margin. Remarkably, achieving this task necessitates only $O_d((\log\varepsilon^{-1})^d)$ weights and $O_d((\log\varepsilon^{-1})^d)$ training samples. We explore how this criterion stacks up against the explicit constructions of currently available Neural Networks (NNs) designed for similar tasks. Significantly, our result is obtained by drawing an innovative connection between the GNN structures and classical sampling theorems. In essence, our pioneering work marks a meaningful contribution to the research domain, advancing our understanding of the practical GNN applications.
\end{abstract}



\section{Introduction} \label{sec:intro}
Graph Neural Networks (GNNs) serve as powerful instruments for processing data associated with graphs and have found widespread applications in various domains such as applied chemistry \cite{fung2021benchmarking, gilmer2017neural, jiang2021could}, biology \cite{hu2021spagcn, song2021dstg, song2021scgcn, wen2022graph}, recommender systems \cite{ gao2022graph, huang2021mixgcf, wu2020graph}, and social networking \cite{chen2017stochastic, fan2019graph, mandal2021graph, tan2019deep}. GNNs harness the inherent topological structure of graphs that are either derived from prior domain knowledge or constructed as proximity graphs based on input features. Such adaptability positions them as valuable in tasks that involve predictions, a fact that is supported by empirical observations. \cite{ala2020improving, ma2021deep}. These findings indicate that GNNs exhibit a remarkable ability to leverage graph information, surpassing the performance of conventional Neural Networks (NNs) in practical predictive scenarios. However, the theoretical foundations of this capability have remained largely uncharted. Additionally, current theoretical investigations of GNNs have predominantly revolved around their ability to distinguish between different graphs, akin to graph-oriented tasks such as the Weisfeiler-Lehman isomorphism test \cite{weisfeiler1968reduction}. Hence, there exists a notable gap in our knowledge of their potential to approximate ``smooth" functions defined on individual graphs, a task that is inherently node-centric. 

In this paper, we engage in an original research initiative to investigate the expressiveness of GNNs in generalizing \textit{bandlimited} functions over Euclidean domains, thus addressing a previously unexplored dimension in our understanding. Specifically, we aim to constructively identify the necessary size requirement for a GNN to effectively generalize an arbitrary bandlimited function over a compact Euclidean domain within a predefined and quantifiable error margin. This pursuit serves a dual purpose: it sets an upper error threshold when GNNs are applied to node classification tasks and thus simultaneously establishes a lower limit for the optimal generalization performance achievable by GNNs. In order to ensure a clear and meaningful notion of the generalization error, we make a crucial assumption: the test data, while drawn from the same distribution as the training data, comprises entirely new data that the trained network has never encountered. To provide context for this assumption and to elucidate the idea of ``smooth" functions on a graph, we adopt a popular perspective from manifold learning that both the training and test data in the graph domain can be regarded as samples originating from an underlying continuous manifold, as discussed in \cite{zemel2004proximity}. From this perspective, the error analysis can then be done on the data manifold outside the training set, and our ultimate objective is to learn across this continuous latent space. While our study primarily focuses on the advantages of GNN models within the simplest manifolds, namely Euclidean cubes, it is essential to recognize that locally differential manifolds can be transformed into these cubes \cite{guillemin2010differential}. This suggests that our findings are likely to hold relevance for understanding GNN capabilities in more complex manifold settings.

The expressive power of NNs, on the other hand, has been extensively explored in a myriad of research endeavors, as exemplified in the literature \cite{inoue2019expressive, kileel2019expressive, lu2017expressive, raghu2017expressive, zhong2017recovery, zweig2021functional}. Within the framework of general approximation capabilities, various manifestations of NNs employing different nonlinear activations have demonstrated their capacity to approximate a broad spectrum of functions with arbitrary precision \cite{cybenko1989approximation, hornik1991approximation, hornik1989multilayer, pinkus1999approximation}. The issue of approximability concerning network size is explored in works such as \cite{candes1999harmonic, chui1992approximation, devore1989optimal, eldan2016power, maiorov1999lower}. In the domain of expressing smooth functions, the classic study by \cite{mhaskar1996neural} established that specific NN constructions, featuring a single hidden layer and of order $O(\varepsilon^{-s})$, achieve $\varepsilon$-approximation for $C^{s}$ functions that are analytically extendable to a polyellipse. (Note that any bandlimited function on a $d$-dimensional Euclidean space extends holomorphically to the corresponding complex space of the same dimension.) Contemporary research on the ability of Rectified Linear Unit (ReLU) NNs to handle functions with high degrees of smoothness is reflected in studies such as \cite{chen2019note, montanelli2021deep, petersen2018optimal, yarotsky2017error}.

In the specific context of NNs' powers to learn bandlimited functions, many works have been done, such as \cite{chen2019note, montanelli2021deep, opschoor2022exponential, wang2018exponential}. In \cite{chen2019note, montanelli2021deep}, it was confirmed, constructively, that deep ReLU Neural Networks (ReLU DNNs) possess the capability to approximate bandlimited functions on Euclidean cubes with an $\varepsilon$-precision. Notably, each specific DNN construction demands $O((\log\varepsilon^{-1})^2)$ layers but a minimum of $\Omega(\varepsilon^{-2})$ weights in the worst case scenario. While these findings align with the emphasis on \textit{dimension-independent} outcomes in their respective papers, they are mitigated by a notable constraint, that is, a polynomial error rate concerning the requisite number of weights, particularly when assuming fixed dimensions. In $\varepsilon$-approximating real-valued functions that are complex-analytically continuable to an open ellipse on $[-1,1]^{d}$, \cite{opschoor2022exponential} introduced a \textit{dimension-dependent} ReLU DNN structure with a favorable logarithmic error rate $O_d((\log\varepsilon^{-1})^{d+1})$ for the required number of weights and a \textit{highly efficient} $O_d(\log\varepsilon^{-1}\log((\log\varepsilon^{-1})^{d+1}))$ rate for the number of layers. 
In parallel, our findings reveal a GNN capable of approximating a bandlimited function on $[-1,1]^{d}$ within an $\varepsilon$-margin that requires only
\begin{equation*}
    O_d((\log\varepsilon^{-1})^{d}) \text{ weights } \quad\text{ and }\quad O_d((\log\varepsilon^{-1})^{d}) \text{ layers},
\end{equation*}
confirming that the additional logarithmic factor in the number of requisite weights documented in \cite{opschoor2022exponential} for NNs would be superfluous. 
Our GNN design is \textit{non-traditional in that the filtering step follows the nonlinear activation}. 
Although the construction requires a considerably deeper network, it achieves an expressiveness level comparable to, or even surpassing, that of NNs, particularly in theoretically worst-case scenarios concerning network weights. Furthermore, it is specifically built on a set of precisely $O((\log\varepsilon^{-1})^{d})$ samples, which are drawn from sampled functional values and concurrently serve as the weights for the network. This arrangement enables us to frame our problem in the generalization context, in which our GNN adeptly interpolates a function based on $O((\log\varepsilon^{-1})^{d})$ instances of the function itself. Consequently, it demonstrates a robust ability to predict values reliably for test data. This distinctive feature broadens the applicability of our work to semi-supervised learning scenarios.

In summary, our contributions to the research on the application of GNNs can be outlined as follows:
\begin{itemize}
    \item We formulate a GNN designed to approximate bandlimited functions on \textit{Euclidean cubes}, demonstrating a competitive rate in the number of required weights compared to traditional NNs. 
    \item Our GNN construction, derived from an interpolation process using sampled functional values, is explicit, offering its use as a pre-trained network and circumventing additional learning steps.
\end{itemize}
In terms of machine learning theory, our work pioneers the incorporation of sampling theory as a novel approach to investigating the expressive power of neural networks; particularly:
\begin{itemize}
    \item Our construction arises from the insight that a \textit{regularized} Whittaker-Kotel'nikov-Shannon sampling theory principle can be realized within the framework of GNNs.
\end{itemize}
This enriches the existing theoretical toolkit, which includes techniques such as Taylor approximation, Legendre approximation, Chebyshev approximation, and Barron's theorem. Further, our sampling-based approach relies exclusively on knowledge of functional sampled values, which stands in contrast to the prevalent Taylor-approximation-based framework found in various nonasymptotic studies on NNs, such as \cite{oono2019approximation, petersen2018optimal, yarotsky2017error}. In these latter approaches, constructions hinge on acquiring pointwise high-ordered derivative values of target functions, which are difficult to obtain with real-world data. Finally, we also want to note that, on the mathematical analysis and approximation theory front, our work produces several noteworthy contributions:
\begin{itemize}
    \item We prove a truncated regularized sampling principle for bandlimited functions and elucidate an interesting convergence property associated with the Fourier transform derived from a truncated Gaussian.
    \item We provide upper bounds for integrals of \textit{factorial polynomials} within the closed Newton-Cotes quadrature formula, a result not previously available in the existing literature.
\end{itemize}

The paper is organized as follows. In \S\ref{sec:Preliminaries}, we establish the groundwork for our discussion, which begins with an introduction to essential mathematical notations and definitions. Following that is \S\ref{sec:GNNIntro}, which is divided into two parts. The first part, \S\ref{sec:basicsGNN}, focuses on the fundamentals of GNNs, and the second part, \S\ref{sec:GNNarch} is dedicated to the explanation of our GNN modeling ideas. In \S\ref{sec:mathbasics} we provide the fundamental sampling theory crucial to our work, paving the way for the presentation of our three main results at the beginning of \S\ref{sec:Results}. We discuss the significance of these results in \S\ref{sec:Ramifications} and, to enhance clarity and simplify our analysis, offer a mock proof in \S\ref{sec:mockproof}. The subsequent sections, \S\ref{sec:sampthmpf}, \S\ref{sec:proxyGNNthm}, \S\ref{sec:actualGNNpf}, contain the proofs of our main results. A comprehensive summary along with a discussion of potential improvements in terms of the number of layers, is given in \S\ref{sec:Discussion}. Lastly, additional technical details are addressed in the appendix. 

\section{Preliminaries}\label{sec:Preliminaries}

We introduce the symbols, notations, and conventions that will be used consistently throughout this paper. 

We denote $\mathbb{Z}_{+}$ to be the set of positive integers and adopt the definition that $\mathbb{N}$ includes all the nonnegative integers. 

Let $A,B\geq 0$. By $A\lesssim B$, we mean there exists a universal constant $C>0$ such that $A\leq CB$, and by $A\asymp B$, we mean $A\lesssim B$ and $B\lesssim A$. If $a$ is a tuning parameter, then the expression $A\lesssim_a B$ implies that there exists a constant depending on $a$, $C(a)>0$, such that $A\leq C(a)B$. Related is the notion of magnitude $O(\tau)$, for some quantity $\tau>0$, by which we mean another quantity whose magnitude is at most $C\tau$, for some $C>0$. Similarly, if $C=C(a)>0$, for some parameter $a$, then we denote this as $O_a(\tau)$.

When $a$ is a scalar value, i.e. $a\in\mathbb{C}$, then $|a|$ denotes the norm value of $a$. When $S$ is an abstract set, then $|S|$ denotes the cardinality of $S$. Furthermore, we distinguish between Euclidean vectors and scalar values using the notation $\vec{a}$ and $a$, respectively. In cases when the nature of the quantity, whether vector-valued or scalar, is uncertain, we simply employ the general notation $a$. Additionally, we use the symbol $\cdot$ to denote either scalar multiplication, as in $a\cdot b$ where at least one of $a, b$ is a scalar, or vector inner product in $\vec{a}\cdot\vec{b}$. 

\paragraph{Restriction, $\delta$ and $\mathrm{ReLU}$.} Let $d\in\mathbb{Z}_+$ and let $S\subset\mathbb{R}^d$. Then $\chi_{S}$ denotes the restriction function on $S$, i.e., $\chi_{S}(\vec{x})=1$ if $\vec{x}\in S$ and $\chi_{S}(\vec{x})=0$ if $\vec{x}\in\mathbb{R}^d\setminus S$. \\
Let $\vec{X}\in\mathbb{R}^d$. By $\delta_{\vec{X}}$ we mean the delta function activated at $\vec{X}$; that is, $\delta_{\vec{X}}(\vec{x})=1$ if $\vec{x}=\vec{X}$ and $\delta_{\vec{X}}(\vec{x})=0$ otherwise. When $\vec{X}=\vec{0}$, we simplify $\delta_{\vec{0}} = \delta$.\\
Now suppose $d=1$, then the ReLU function, shifted by $X\in\mathbb{R}$, is expressed as, 
\begin{equation*}
    \mathrm{ReLU}_{X}(x) := \mathrm{ReLU}(x-X) := (x-X)\cdot\chi_{\{(x-X)\geq 0\}},
\end{equation*}
where we adopt that $\mathrm{ReLU}_0(x) = \mathrm{ReLU}(x) = x\cdot\chi_{\{x\geq 0\}}$. 

\paragraph{Support.} Let $d\in\mathbb{Z}_+$ and $f:\mathbb{R}^{d}\to\mathbb{R}$ be a measurable function. Then the \textit{essential support} of $f$, denoted $supp(f)$, is the smallest closed set $S\subset\mathbb{R}^d$, with respect to inclusion, such that $f(x)=0$ a.e. $x$ if $x\not\in S$.

\paragraph{Lebesgue spaces on $\mathbb{R}^d$.} Let $d\in\mathbb{Z}_+$ and $f:\mathbb{R}^{d}\to\mathbb{R}$ be a measurable function. Let $1\leq p\leq\infty$. Then for $1\leq p<\infty$, we say $f\in L^p(\mathbb{R}^d)$ if 
\begin{equation*} 
    \|f\|_p := \bigg(\int_{\mathbb{R}^d} |f(\vec{x})|^p \,d\vec{x}\bigg)^{1/p} <\infty,
\end{equation*}
and $f\in L^\infty(\mathbb{R}^d)$ if
\begin{equation*} 
    \|f\|_\infty := ess\sup_{\vec{x}\in\mathbb{R}^d} |f(\vec{x})| <\infty.
\end{equation*}
Here, $ess\sup$ stands for ``essential supremum", taken over a.e. $\vec{x}$. The norm notation $\|f\|_p$ is written with an understanding that the domain dimension $d$ of $\mathbb{R}^d$ is specified by the context. 

\paragraph{Fourier transform.} Let $f:\mathbb{R}^{d}\to\mathbb{R}$. We define its Fourier transform, $\mathcal{F}f$, as,
\begin{equation} \label{foudef1} 
    \mathcal{F}f(\vec{w})=\hat{f}(\vec{w}):=\int_{\mathbb{R}^{d}} f(\vec{x})e^{-i\vec{w}\cdot\vec{x}}\,d\vec{x}.
\end{equation}
The inverse Fourier transform of $f$, denoted $\mathcal{F}^{-1}f$, is 
\begin{equation} \label{foudef2}
    \mathcal{F}^{-1}f(\vec{x})=\widecheck{f}(\vec{x}):=\frac{1}{(2\pi)^{d}}\int_{\mathbb{R}^{d}} f(\vec{w})e^{i\vec{w}\cdot\vec{x}}\,d\vec{w}.
\end{equation}
Let $\mathcal{I}f:=f(-\cdot)$. Then \eqref{foudef2} is a part of the Fourier inversion theorem \cite[Theorem 8.26]{folland1999real} that says,
\begin{equation*} 
    \mathcal{F}^{-1}=\frac{1}{(2\pi)^{d}}\mathcal{F}\circ\mathcal{I}=\frac{1}{(2\pi)^{d}}\mathcal{I}\circ\mathcal{F}
\end{equation*}
whenever applicable. Moreover, by Plancherel's theorem \cite[Theorem~8.29]{folland1999real}
\begin{equation} \label{normcompare}
    \|f\|_2^2 = \frac{1}{(2\pi)^d}\|\hat{f}\|_2^2.
\end{equation}
Now suppose $f\equiv\chi_{[-\pi,\pi]^{d}}$ in \eqref{foudef2}. Then
\begin{equation} \label{sinc}
    \frac{1}{(2\pi)^{d}}\int_{[-\pi,\pi]^{d}}f(\vec{x})e^{i \vec{w}\cdot\vec{x}}\,d\vec{x}=\prod_{j=1}^{d}\frac{1}{2\pi}\int_{[-\pi,\pi]}e^{iw_{j}x_{j}}\,dx_{j}=\prod_{j=1}^{d}\sinc(w_{j})=:\sinc(\vec{w}).
\end{equation}
where $\sinc(x) := \frac{\sin(\pi x)}{\pi x}$. Note that, by definition, $\sinc(N)=0$ when $N\in\mathbb{Z}\setminus\{0\}$, a fact that will prove to be useful later in \S\ref{sec:sampthmpf}.

\subsection{Graph Neural Networks}\label{sec:GNNIntro}

\subsubsection{Basics} \label{sec:basicsGNN}

We represent a graph $G$ by a triple $G = (V,E,w)$. Here, $V$ denotes the set of graph nodes, and $E\subset V\times V$ the set of graph edges. Adopting a predetermined node labeling, we express $V$ as the set $V = \{v_1,v_2,\dots,v_{|V|}\}$. The edge weight function $w: V\times V\to [0,1]$ describes the connection strength among the nodes. That is, $w(v_j,v_k)> 0$ iff there exists an edge connecting $v_j, v_k$, i.e. $(v_j,v_k) \in E$. Associated with the edge weight function is the $|V|\times |V|$ graph adjacency matrix, symbolized by ${\bf A}$, whose $jk$-entry satisfies ${\bf A}_{jk}=w(v_j,v_k)$. When $G$ is a \textit{simple, undirected, unweighted} graph, it is denoted by a pair $G=(V,E)$. In this case, it is understood that $w(v_j,v_k)=1$ if $(v_j,v_k) \in E$ and $w(v_j,v_k)=0$ otherwise. Additionally, $w$ must meet two conditions: symmetry, $w(v_j,v_k) = w(v_k,v_j)$ and absence of self-loops, $w(v_j,v_j)=0$. The corresponding adjacency matrix ${\bf A}$ is then a symmetric matrix with $0,1$-entries and $0$'s on the diagonal. 

For each node $v\in V$, let it be associated with a feature vector $X_{v}\in\mathbb{R}^{d}$. Let ${\bf X}$ be the matrix that holds all the feature vectors as its columns, i.e. ${\bf X} \equiv \begin{bmatrix} (X_{v})_{v\in V}\end{bmatrix}$, or using the aforementioned predetermined node labeling, 
\begin{equation} \label{def:featuremat}
    {\bf X} = \begin{bmatrix} X_1 & X_2 & \cdots & X_{|V|}\end{bmatrix} \in\mathbb{R}^{d\times |V|}.
\end{equation}
Such feature matrix ${\bf X}$ is used as an input of a GNN, whose output is another feature matrix ${\bf Y} \equiv \begin{bmatrix} (Y_{v})_{v\in V}\end{bmatrix}$, where $Y_{v}$ is the output feature vector for the node $v$; alternatively,
\begin{equation*}
    {\bf Y} = \begin{bmatrix} Y_1 & Y_2 & \cdots & Y_{|V|}\end{bmatrix} \in\mathbb{R}^{m\times |V|}.
\end{equation*}
The input and output dimensions, respectively, $d, m$, are problem-dependent. For example, nodes can be assigned, as input feature vectors, coordinate vectors that depict their locations in space. Analytically, both the output and input of a GNN can be regarded as \textit{graph signals}, i.e. functions defined on the graph nodes.

In terms of structure, the defining characteristic of a GNN as a neural network lies in its architecture. Specifically, GNNs are built by combining multiple iterations of two distinct computational units, given as follows:
\begin{itemize}
    \item {\bf Multilayer Perceptron (MLP)}: An MLP is a two-step process of first applying an affine transformation and then a nonlinear activation function. For example, for a \textit{weight matrix} ${\bf W} \in\mathbb{R}^{m\times d}$, a \textit{bias vector} $\vec{b}\in\mathbb{R}^{m}$, and a nonlinear activation function $\rho$, an MLP unit acts symbolically on ${\bf X}$ in \eqref{def:featuremat} as 
    \begin{equation*} 
        {\bf X} \mapsto \rho({\bf W}{\bf X} + \vec{b}),
    \end{equation*}
    where $\rho$ is applied componentwise. These units are the fundamental building blocks of feedforward neural networks. Throughout this work we consider ReLU activation functions, for which the resulting GNN will be called a \textit{ReLU GNN}.
    \item {\bf Filter}: A graph filter unit is what distinguishes a GNN from a standard feedforward NN architecture. Abstractly, a graph filtering process is any process that takes in nodal features (e.g. {\bf X} in \eqref{def:featuremat}) \textit{and} graphical structure $(V,E,w)$ and outputs a new set of nodal features. There are two types of graph filters: spatial-based and spectral-based. The spatial filters explicitly leverage the graph connections to perform a feature-refining process, whereas the spectral ones utilize spectral graph theory to design filtering in the spectral domain. Some well-known spectral filters can be considered spatial filters \cite[Subsection~5.3]{ma2021deep}.
\end{itemize}

It is important to note that the order of activation, affine transformation, and graph filtering operations in a GNN can be tailored. For instance, the following order, in which filtering is applied before activation, is commonly recognized in practice: 
\begin{equation} \label{commondesign}
    \rho({\bf Filter}({\bf W}{\bf X} + \vec{b})).
\end{equation}
Another instance can be found in \cite{ruiz2020graphon}, where, to harness spectral theory, an implicit ordering was employed in which filtering and affine transformation in \eqref{commondesign} were interchanged. In this work, however, as to accommodate our intended application of sampling theory, we adopt the order
\begin{equation} \label{ourorder}
    {\bf Filter}(\rho({\bf W}{\bf X} + \vec{b})).
\end{equation}
Using this, a GNN operation can be mathematically expanded as
\begin{equation} \label{gnnex}
    {\bf Y}\equiv \rho_{L}({\bf W}^{(L)} \psi_{L-1}\circ\psi_{L-2} \circ\cdots\circ \psi_1({\bf X})+ b^{(L)}).
\end{equation}
Here, $L\in\mathbb{Z}_+$, and each $\psi_{j}$ in \eqref{gnnex} is an intermediate or hidden \textit{network layer}, which is a composition of an MLP and a filter unit, 
\begin{equation}\label{GNNlayer}
    \begin{split}
        &{\bf H}_{j} = \psi_{j}({\bf H}_{j-1}) = {\bf Filter} (\rho({\bf W}^{(j)}{\bf H}_{j-1} + b^{(j)})), \quad  j=1,\cdots, L-1,\\
        &{\bf H}_0 \equiv {\bf X}.
    \end{split}
\end{equation}
I.e., at the $j$th layer, the output of the previous layer ${\bf H}_{j-1}$, is fed to an MLP (with a weight matrix ${\bf W}^{(j)}$, bias vector $b^{(j)}$ and activation $\rho$) and then to a graph filter ${\bf Filter}$. The last $L$th layer is the output layer; note from \eqref{gnnex} that ${\bf Filter}$ is not performed here. The last activation $\rho_{L}$ is also optional; for example, one can take $\rho_{L}$ to be a softmax function to ensure that the output is bounded. The GNN \textit{network parameter} is the set of all utilized weights, which are entries of the weight matrices, all the bias vectors, and the number of layers, i.e. 
\begin{equation} \label{parameter}
    \{{\bf W}^{(j)}_{kl}, b^{(j)}, j=1,\cdots,L\}.
\end{equation}
The performance of a GNN is often contingent on its parameter. Note that, when filtering is not employed at any intermediate layer in \eqref{GNNlayer}, we obtain a standard feedforward NN structure, composed of MLPs.

\subsubsection{GNN architecture of choice} \label{sec:GNNarch}

In this study, we use (linear) spatial-based graph filters (see also Remark~\ref{rem:filter}). Such filtering exchanges information between graph nodes and is achieved by a \textit{graph filter kernel} (also referred to as a {\em graph shift operator} \cite{ruiz2020graphon}) ${\bf K} \in \mathbb{C}^{|V|\times|V|}$ which encodes the graph topology of $G$. Namely, take ${\bf X}$ in \eqref{def:featuremat}, then a filtering application on ${\bf X}$ takes the form 
\begin{equation}\label{GSO}
    {\bf Filter}: \quad {\bf X} \mapsto {\bf K}{\bf X}^{T}.
\end{equation} 
The choice of ${\bf K}$ can be determined by the user and is often problem-dependent. As will become evident shortly, we explore graph filter kernels of the form
\begin{equation}\label{GSOchoice}
    {\bf K} = {\bf G} \odot {\bf A},
\end{equation}
where $\odot$ is the Hadamard product, ${\bf A}$ is again the adjacency matrix of $G$, and ${\bf G}\in\mathbb{R}^{|V|\times|V|}$ is employed to localized the information propagation by ${\bf X}$ in \eqref{GSO}.

As outlined in the introduction \S\ref{sec:intro}, the central thesis of this study is to showcase the structural advantage of GNNs over regular NNs when it comes to generalizing bandlimited functions on a fundamental Euclidean domain $[-1,1]^d$. We accomplish this by presenting a GNN that can outperform the documented predictive capacity of NNs on a dense lattice covering the said domain. More specifically, we begin by providing a graph structure that is the lattice 
\begin{equation} \label{totalgrid}
    [-5,5]^{d}\cap\tau\mathbb{Z}^{d} := [-5,5]^{d}\cap\{\tau\vec{n}:\vec{n}\in\mathbb{Z}^{d}\},
\end{equation}
for some step size $\tau>0$ small such that $1/\tau\in\mathbb{Z}_+$. Our predictions are then carried out on the sub-lattice $[-1,1]^{d}\cap\tau\mathbb{Z}^{d}$. The reason for this selection is two-fold. Since our result is an existence result, we solely need to specify an appropriate graph choice. First, an integer lattice is convenient because it discretizes a Euclidean cube. The denser the lattice is, i.e. the smaller $\tau$ is, the closer it approximates the entire cube. The lattice points are the graph nodes, with each node $v$ being assigned its $d$-dimensional coordinate vector as an input feature $X_v$, while the lattice edges serve as the graph edges. Therefore, this is a proximity graph, in that nodes are connected based on the similarity of their associated features. In this context, a bandlimited function $f:\mathbb{R}^d\to\mathbb{R}$ restricted to the lattice $[-1,1]^d\cap\tau\mathbb{Z}^d$ can be treated as a graph signal. The output of our GNN, treated as another graph signal, is expected to proficiently learn the former. Second, this lattice graph selection plays a strategic role enabling us to connect between the discrete GNN structure and a continuous setting on which our analysis is based, as we now explain. Observe, on the lattice \eqref{totalgrid}, the filtering operation described in \eqref{GSO} yields a $k$th column, where $k=1,\cdots,d$, that can be expressed analytically as
\begin{equation} \label{koutput}
    \begin{bmatrix} \sum_{j=1}^{|V|} {\bf K}_{1j} (X_{v_j})_k \\ \vdots \\ \sum_{j=1}^{|V|} {\bf K}_{|V|j} (X_{v_j})_k \end{bmatrix}.
\end{equation}
Here, $(X_{v_j})_k$ represents the $k$th entry of the vector $X_{v_j}\in\mathbb{R}^d$. To gain further insight, we examine the $l$th entry of the column in \eqref{koutput}, which is $\sum_{j=1}^{|V|} {\bf K}_{lj} (X_{v_j})_k$. We select ${\bf K}$ in \eqref{GSOchoice} such that ${\bf K}_{lj}$ is a function of the Euclidean distance between $X_{v_l}$ and $X_{v_j}$; symbolically, we write ${\bf K}_{lj}=\mathcal{K}(X_{v_l}-X_{v_j})$, where $\mathcal{K}$ is a, real-valued, radial function on $\mathbb{R}^d$. Then
\begin{equation} \label{koutputconvo}
    \sum_{j=1}^{|V|} {\bf K}_{lj} (X_{v_j})_k = \sum_{j=1}^{|V|} \mathcal{K}(X_{v_l}-X_{v_j})(X_{v_j})_k.
\end{equation}
As previously mentioned, the choice of ${\bf K}$ depicted in \eqref{GSOchoice} is to localize the information aggregation and therefore limit the number of nodes $v_j$ in a vicinity of $v_l$ participating in the sum in \eqref{koutputconvo}. Further write $g(X_{v_j})=(X_{v_j})_k$. Then as the lattice \eqref{totalgrid} becomes denser, the discrete sum in \eqref{koutputconvo} approaches a continuous convolution. Heuristically:
\begin{equation} \label{locfil}
    \sum_{j=1}^{|V|} \mathcal{K}(X_{v_l}-X_{v_j})(X_{v_j})_k = \sum_{j=1}^{|V|} \mathcal{K}(X_{v_l}-X_{v_j})g(X_{v_j}) \approx \int_{\mathcal{N}(v_l)} \mathcal{K}(X_{v_l}-y)g(y)\,dy,
\end{equation}
where $\mathcal{N}(v_l)$ denotes a Euclidean neighborhood of $v_l$, the size of which is determined by the \textit{convolution kernel} $\mathcal{K}$. Note that \eqref{locfil} aligns with a popular perspective in machine learning that, when graphs are regarded as discretized manifolds, graph data converge to manifold data in the large-scale limit \cite{hein2007graph}. It also outlines our mathematical modeling objective, as follows.

We adopt a three-step modeling approach. We first consider an auxiliary continuous analog of \eqref{GNNlayer} in \textit{dimension one}, replacing the graph filtering with a convolution, namely:
\begin{equation} \label{proxygnnlayer}
    \begin{split}
        g_L(x) &= \rho_L({\bf W}^{(L)}g_{L-1}+b^{(L)})(x), \\
        g_j(x) &= \psi_{j}(g_{j-1})(x) = \mathcal{K}\ast (\rho({\bf W}^{(j)}g_{j-1} + b^{(j)}))(x), \quad j=1,\cdots,L-1\\
        g_0(x) &= x,
    \end{split}   
\end{equation} 
for $x\in\mathbb{R}$. It is important to note that any $g_j$ among the outputs $g_1, \cdots, g_{L-1}$ can be vector-valued functions. We then establish the interpolation power of this so-called \textit{proxy GNN} model \eqref{proxygnnlayer}. Following that, we derive an actual GNN model in \textit{dimension one} via numerical integration, with a small accompanying error. Subsequently we extend the model to higher dimensions and provide insights into its expressive capacity. This succinctly summarizes our analysis. 

\begin{figure}[htb!]
	\begin{center}
		\includegraphics[width = 5.5 in]{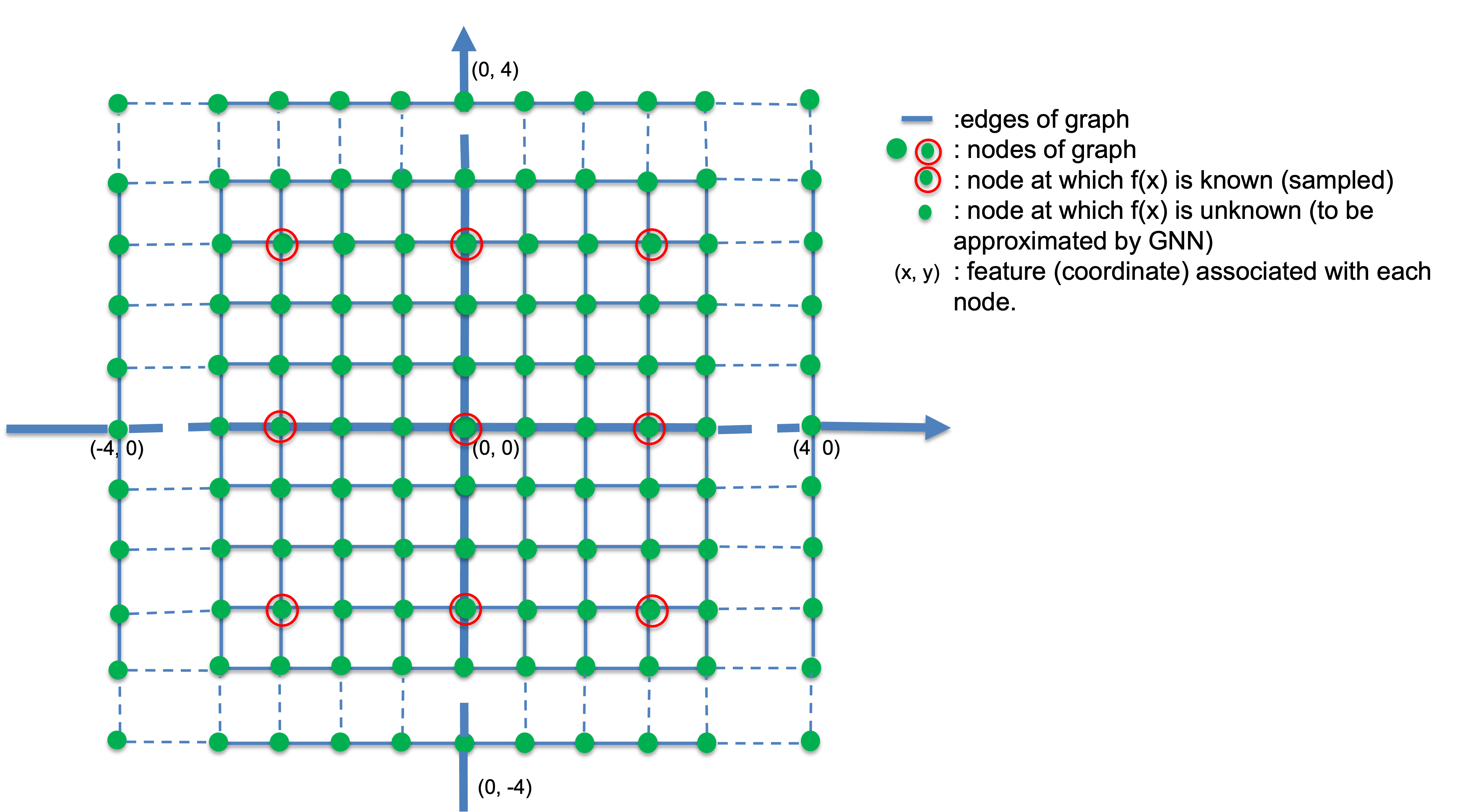}
		\caption{\small Toy example of graph used in GNN when learning a bandlimited function on $[-1, 1]^2$. }
		\label{fig:TCGA}
	\end{center}
\end{figure}

We will further explore the details of how \eqref{proxygnnlayer} lends itself to the application of sampling theory in \S\ref{sec:mockproof}, and additionally, provide an outline of the proof for the main results.

\subsection{Bandlimited functions and the basic mathematics of sampling} \label{sec:mathbasics}

To begin, let $\sigma>0$. We call $f:\mathbb{R}\to\mathbb{R}$ is a $\sigma$-bandlimited function if $f\in L^2(\mathbb{R})$ and if $supp(\hat{f})\subset [-\sigma,\sigma]$. In other words, we can express
\begin{equation} \label{bl}
    f(x)= \frac{1}{2\pi}\int_{-\sigma}^{\sigma} \hat{f}(t)e^{ixt}\,dt, \quad\forall x\in\mathbb{R}.
\end{equation}
For such $f$, we designate $f\in B^2_\sigma$. We remark that, typically, the symbol $B^2_\sigma$ refers to the \textit{Bernstein space} that comprises all bounded functions $g\in L^2(\mathbb{R})$ extendable to an entire function of \textit{exponential type} $\sigma$ on $\mathbb{C}$. For a more comprehensive definition, we refer the reader to \cite[Subsection~7.2.2]{pfander2015sampling}. Notably, the Paley-Wiener theorem \cite[Theorem~X]{paley1934fourier} establishes an equivalence between $\sigma$-bandlimited functions and those belonging to $B^2_\sigma$. Therefore, we employ the said symbol as a convenient shorthand for the former class of functions.

The preceding remark confirms that if $f\in B^2_\sigma$, then $f$ is a smooth function of all orders. Furthermore, $f\in L^\infty(\mathbb{R})$; indeed, by \eqref{normcompare} and the inversion formula \eqref{bl}
\begin{equation} \label{L2toLinfty}
    |f(x)| \leq\frac{\sqrt{2\sigma}}{2\pi}\|\hat{f}\|_2 = \sqrt{\frac{\sigma}{\pi}}\|f\|_2, \quad\forall x\in\mathbb{R}.
\end{equation}
It is also well-known that a bandlimited function $f\in B^2_\sigma$ can be recovered via its discretely sampled values, as outlined below.

\begin{theorem}[Whittaker-Kotel’nikov-Shannon sampling theorem] \label{WKSthm} 
\cite{shannon1949communication} Let $f\in B^2_\sigma$ for some $\sigma>0$. Then,
\begin{equation} \label{wks}
    f(x) = \sum_{n\in\mathbb{Z}}f(n\pi/\sigma)\frac{\sin\pi(\sigma x/\pi-n)}{\pi(\sigma x/\pi-n)} = \sum_{n\in\mathbb{Z}}f(n\pi/\sigma)\sinc(\sigma x/\pi-n)
\end{equation}
where the series converges uniformly on compact subsets of $\mathbb{R}$.
\end{theorem}

In terms of information, \eqref{wks} signifies a remarkable trade-off, in that, to acquire the values of the signal $f$ at every point on $\mathbb{R}$, one only needs to know its values sampled on a discrete lattice of $\mathbb{R}$. When dealing with $f\in B^2_\pi$, i.e. $\sigma=\pi$, the series presented in \eqref{wks} simplifies to a more familiar form
\begin{equation*}
    f(x) = \sum_{n\in\mathbb{Z}}f(n)\sinc(x-n).
\end{equation*}
We will use this formula frequently in \S\ref{sec:sampthmpf}. 

The concept of bandlimited functions can be generalized to high dimensions, as follows.

\begin{definition}[Bandlimited functions on $\mathbb{R}^d$] \label{def:bandlmtd} Let $d\in\mathbb{Z}_+$ and $\sigma>0$. Then a function $f:\mathbb{R}^d\to\mathbb{R}$ is said to be $[-\sigma,\sigma]^d$-bandlimited if it holds that
\begin{equation} \label{highdimsupp}
    supp(\hat{f})\subset [-\sigma,\sigma]^d. 
\end{equation} 
Such $f$ is denoted as $f\in B^2_\sigma$ if $d=1$ and as $f\in B^2_{[-\sigma,\sigma]^d}$ if $d>1$.
\end{definition}

Just as in the derivation \eqref{L2toLinfty}, it can be inferred from \eqref{foudef2}, \eqref{normcompare}, \eqref{highdimsupp} that 
\begin{equation*}
    \|f\|_\infty\leq \bigg(\frac{\sigma}{\pi}\bigg)^{d/2}\|f\|_2. 
\end{equation*}
A version of Theorem~\ref{WKSthm} tailored for high dimensions is also available.

\begin{theorem}[Parzen sampling theorem] \label{parzensampthm} \cite{parzen1956simple}
Let $d\in\mathbb{Z}_+$ and let $f:\mathbb{R}^d\to\mathbb{R}$ be such that $f\in B^2_{[-\sigma,\sigma]^d}$ for some $\sigma>0$. Then,
\begin{equation} \label{parzen}
    f(\vec{x}) = \sum_{\vec{n}\in\mathbb{Z}^{d}}f(\vec{n}\pi/\sigma)\prod_{j=1}^{d}\frac{\sin(\sigma x_{j}-n_{j}\pi)}{(\sigma x_{j}-n_{j}\pi)} =\sum_{\vec{n}\in\mathbb{Z}^{d}}f(\vec{n}\pi/\sigma)\sinc(\vec{x}\sigma/\pi - \vec{n})
\end{equation}
where the series converges uniformly on compact subsets of $\mathbb{R}^d$.
\end{theorem}

It is evident that Theorem~\ref{WKSthm} is a specific instance of Theorem~\ref{parzensampthm}. We conclude this subsection by offering a noteworthy remark and a practical proposition. First, the series \eqref{wks} (resp. \eqref{parzen}) in fact converge in $L^2(\mathbb{R})$ (resp. $L^2(\mathbb{R}^d)$) and uniformly on $\mathbb{R}$ (resp. $\mathbb{R}^d$). This is the consequence of Kluv\'anek's sampling theorem, presented as Theorem~\ref{Ksampthm} in \S\ref{appx:sampling}. Particularly, we will also explore how Theorem~\ref{parzensampthm} emerges from this theorem in the same subsection. Second, as another direct implication of the said Theorem~\ref{Ksampthm}, we can compute the total energy of a bandlimited function from its sampled values. We extract this result in the following lemma, which will play a pivotal role in our subsequent analysis. 

\begin{lemma} \label{lem:totaleng} Let $d\in\mathbb{Z}_+$ and $f\in B^2_{[-\sigma,\sigma]^d}$, for some $\sigma>0$. Then the following Gaussian quadrature formula holds
\begin{equation} \label{Gaussquad}
    \|f\|^2_2 = \frac{1}{(2\pi)^d}\|\hat{f}\|^2_2 = \sum_{\vec{n}\in\mathbb{Z}^d} |f(\vec{n}\pi/\sigma)|^2. 
\end{equation}
\end{lemma}

\section{Main results}\label{sec:Results}

We present here our three main contributions. First, we establish a reconstruction theorem based on uniform sampling, Theorem~\ref{thm:samp}. We then proceed to showcase a result on the expressiveness of the proxy GNN model in dimension one \eqref{proxygnnlayer} introduced in \S\ref{sec:GNNarch}, presented as Theorem~\ref{thm:proxyGNN}. Finally, leveraging this insight, we demonstrate the generalization capacity of GNNs on $[-1,1]^d$, encapsulated in Theorem~\ref{thm:actualGNN}. We conclude this section with an exploration of the implications of our findings in \S\ref{sec:Ramifications}, after which, we delve into a mock proof in \S\ref{sec:mockproof}. This mock proof serves a dual role: it untangles the analysis involved and provides a preview of how sampling theory fits into our framework. Detailed proofs are left to the sections that follow.

Let $h,r>0$ and $M\in\mathbb{Z}_{+}$. These are our parameters. Let $d\in\mathbb{Z}_+$. Let $h$ be small; for $\sigma<\pi/h$, we define the mapping $\mathcal{R}_{M,r,h}$ on $B^2_{[-\sigma,\sigma]^d}\subset B^2_{[-\pi/h,\pi/h]^d}$ such that
\begin{equation} \label{Op3}
    \begin{split}
        \mathcal{R}_{M,r,h}f(\vec{x}) &:= \sum_{\vec{n}\in\mathbb{Z}^d}f(h\vec{n})\sinc{}_{h}(\vec{x}-h\vec{n})\mathcal{G}_{M,r,h}(\vec{x}-h\vec{n})\\
        & := \sum_{\vec{n}\in\mathbb{Z}^d}f(h\vec{n})\sinc(h^{-1}\vec{x}-\vec{n})\mathcal{G}_{M,r}(h^{-1}\vec{x}-\vec{n}),
    \end{split}
\end{equation}
for $f\in B^2_{[-\sigma,\sigma]^d}$. Here, as indicated, $g_h$ denotes a scaled version, $g_h = g(h^{-1}\cdot)$, of a function $g$. The function $\sinc$ has been defined in \eqref{sinc}, and $\mathcal{G}_{M,r}(\vec{x}) := \prod_{j=1}^{d}\mathcal{G}_{M,r}(x_j)$, where
\begin{equation} \label{def:truncGaussian}
    \mathcal{G}_{M,r}(x):=
    \begin{cases}
        \mathcal{G}_{r}(x) &\text{ if } x\in [-M,M], \\ 
        0 &\text{ otherwise}
    \end{cases}
\end{equation} 
is the truncated version of the Gaussian $\mathcal{G}_{r}(x)=\exp(-\frac{x^2}{2r^2})$ with variance $r$. It is evident that $\mathcal{G}_{M,r}$ is continuously differentiable to an infinite order at every $x$, except for at the truncation points $x=\pm M$. The mapping described in \eqref{Op3} is a type of reconstruction mapping. Indeed, we illustrate in Theorem~\ref{thm:samp} below that, by selecting suitable parametric values for $M, h, r$ in relation to $\sigma$, $\mathcal{R}_{M,r,h}$ approximates the identity mapping $Id$ on $B^2_\sigma$. 

\begin{theorem}[Reconstruction from regularized sampling] \label{thm:samp} 
Let $d\in\mathbb{Z}_+$ and $f\in B^2_{[-\sigma,\sigma]^d}$, for some $\sigma>0$. Let $\mathcal{R}_{M,r,h}$ be as in \eqref{Op3}. Set $h=4/M$, where $M\in\mathbb{Z}_+$ is chosen sufficiently large so that $\sigma<\pi/(2h) = M\pi/8$, and set $r=\sqrt{M}$. Then
\begin{equation} \label{sampthmconc}
    \begin{split}
        \|\mathcal{R}_{M,r,h}f-f\|_2 &\lesssim_d \exp(-M/4)\|f\|_2\\
        \|\mathcal{R}_{M,r,h}f-f\|_\infty &\lesssim_d \sigma^{d/2}\exp(-M/4)\|f\|_2.
    \end{split}
\end{equation}
\end{theorem}

Theorem~\ref{thm:samp} is our first result in this paper and stands on its own as an interesting regularized sampling theorem. It follows from definition \eqref{def:truncGaussian} that the total number of sampling points used in \eqref{Op3} does not exceed $(2M+1)^d$. Therefore, it can be inferred from \eqref{sampthmconc} that $\mathcal{R}_{M,r,h}f$ offers an approximation to $f$ with a $d$th-root subgeometric rate of convergence in terms of the number of sampling points; such finding holds significant implications for our subsequent discussions on GNNs. The proof of Theorem~\ref{thm:samp} can be found in \S\ref{sec:sampthmpf}. 

We wish to note that a previous work \cite{liwen2004regularized} touched upon a similar result. However, it is important to acknowledge that their proof contained several errors and lacked a rigorous examination of the Fourier transform of $\sinc{}_h\cdot\mathcal{G}_{M,r,h}$, an oversight that we aim to rectify in the current study. For example, unlike what was claimed in \cite{liwen2004regularized}, the said Fourier transform is not nonnegative, an essential assumption for their result of \eqref{sampthmconc}. Nevertheless, we are able to uphold Theorem~\ref{thm:samp}, by substituting the nonnegativity requirement with a compelling demonstration of the following crucial Fourier convergence fact when $d=1$:
\begin{equation} \label{crucialconvergence}
    \sup_{w\in [-\sigma,\sigma]} \sum_{k\in\mathbb{Z}\setminus\{0\}} |\mathcal{F}(\sinc{}_h\cdot\mathcal{G}_{M,r,h})(w + 2k\pi)| <\infty.
\end{equation}
Moreover, the convergence in \eqref{crucialconvergence} is exponential in terms of $M$ with the parameter choices specified in Theorem~\ref{thm:samp}. Such an outcome, established as Proposition~\ref{prop:2crucialsizes} in \S\ref{sec:sampthmpf}, is a remarkable analytic fact, considering the slow decay of $\mathcal{F}(\sinc{}_h\cdot\mathcal{G}_{M,r,h})$ at infinity.

Next, we turn to a generalization result specific to the proxy GNN model \eqref{proxygnnlayer}, the underpinning theory for which stems from Theorem~\ref{thm:samp}.

\begin{theorem}[Generalization in dimension one using proxy GNN \eqref{proxygnnlayer}] \label{thm:proxyGNN}
Let $f\in B^2_\sigma$, for a bandwidth $\sigma>0$. Let $h=4/M$, where $M\in 4\mathbb{Z}_+$ be sufficiently large so that $\sigma<\pi/(2h) = M\pi/8$. Suppose that one has access to the following uniformly sampled values of $f$: $\{f(hn)\}_{n=-M/2}^{M/2}$. Then there exists a proxy GNN $\Psi_f$ specified by $f$, using $M+1$ weights that are precisely the sampled values, and two layers (including one filter layer and one MLP layer), such that
\begin{equation*}
    \|(f-\Psi_f)\cdot\chi_{[-1,1]}\|_2, \quad \|(f-\Psi_f)\cdot\chi_{[-1,1]}\|_\infty \lesssim M^{3/2}\exp(-M/32)\|f\|_2.
\end{equation*}
\end{theorem}

Another way to rephrase Theorem~\ref{thm:proxyGNN} is as follows.

\begin{theorem} \label{thm:proxy}
Let $f\in B^2_\sigma$, for a bandwidth $\sigma>0$. Then for every $\varepsilon>0$ small, there exists a proxy GNN $\Psi_f$ specified by $f$, with $O(\log\varepsilon^{-1})$ weights and two layers (including one filter layer and one MLP layer), such that
\begin{equation*} 
    \|(f-\Psi_f)\cdot\chi_{[-1,1]}\|_2, \quad \|(f-\Psi_f)\cdot\chi_{[-1,1]}\|_\infty
    \lesssim \varepsilon\|f\|_2.
\end{equation*}
In addition, the $O(\log\varepsilon^{-1})$ weights employed in constructing $\Psi_{f}$ consist of uniformly sampled values of $f$. 
\end{theorem}

Note that, although in the context of the presented language, Theorem~\ref{thm:proxyGNN} (and also Theorem~\ref{thm:proxy}) can be construed as an approximation outcome, it is a generalization result. It asserts that the proxy GNN can effectively generalize the function $f$ even when equipped with only sparse information about the function itself, specifically the sampled values $\{f(hn)\}_{n=-M/2}^{M/2}$. Nevertheless, we want to stress that transitioning between the two interpretations, namely approximation and generalization, should not entail any conflict. In evaluating the generalization power of the proxy GNN $\Psi_f$, it is perfectly reasonable for us to assume that we already have the ground truth profile of $f$ at our disposal. This perspective will guide our approach throughout the proof of Theorem~\ref{thm:proxyGNN} (hence of Theorem~\ref{thm:proxy}), which is given in \S\ref{sec:proxyGNNthm}. 

Although Theorem~\ref{thm:proxyGNN} addresses the proxy GNN's expressiveness, it still serves as a distinctive and interesting generalization result. More importantly, it lays the theoretical groundwork for the subsequent GNN outcome, which is essentially a discretized rendition of the proxy model. In addition, we will later use the latter to formulate a mock proof in \S\ref{sec:mockproof}, capitalizing on its simplified mathematics to streamline complexities and thus enhance the accessibility of our presentation.

We are now prepared to introduce the central highlight of our results, a theorem pertaining to the prowess of GNNs, as promised. The detailed proof of this theorem is provided in \S\ref{sec:actualGNNpf}.

\begin{theorem}[Generalization using GNN] \label{thm:actualGNN}
Let $d\in\mathbb{Z}_+$ and let $f\in B^2_{[-\sigma,\sigma]^d}$ for some $\sigma>0$. Let $h=4/M$, where $M\in 4\mathbb{Z}_+$ be sufficiently large so that $\sigma<\pi/(2h) = M\pi/8$. Suppose that one has access to the following uniformly sampled values of $f$: 
\begin{equation*}
    \{f(h\vec{n}): \vec{n}=(n_1,\cdots,n_d), n_j = -M/2,\cdots,M/2, j=1,\cdots, d\}.
\end{equation*}
Then there exists a ReLU GNN $\tilde{\Psi}_f$ specified by $f$, using $O_d(M^d)$ weights that are precisely the sampled values, and $O_d(M^d)$ layers (including one filter layer), such that, for any $\tau\leq 1/(5M)$ and $\vec{x}\in [-1,1]^d\cap\tau\mathbb{Z}^d$,
\begin{equation} \label{GNNthmconc}
    |f(\vec{x})-\tilde{\Psi}_f(\vec{x})| \lesssim_d M^{5d}\exp(-M/32)\|f\|_2.
\end{equation}
\end{theorem}

Similarly as before, we can also express Theorem~\ref{thm:actualGNN} in an alternative manner.

\begin{theorem} \label{thm:actual}
Let $d\in\mathbb{Z}_+$ and let $f\in B^2_{[-\sigma,\sigma]^d}$ for some $\sigma>0$. Then for every $\varepsilon>0$ small, there exists a ReLU GNN $\tilde{\Psi}_f$ specified by $f$, with $O_d((\log\varepsilon^{-1})^d)$ weights and $O_d((\log\varepsilon^{-1})^d)$ layers (including one filter layer), such that, for any $\tau\leq 1/(5M)$ and $\vec{x}\in [-1,1]^d\cap\tau\mathbb{Z}^d$,
\begin{equation*} 
    |f(\vec{x})-\tilde{\Psi}_f(\vec{x})| \lesssim_d \varepsilon\|f\|_2.
\end{equation*}
In addition, the $O_d((\log\varepsilon^{-1})^d)$ weights employed in constructing $\tilde{\Psi}_f$ consist of uniformly sampled values of $f$. 
\end{theorem}

Having now unveiled our main results, we proceed to illuminate their key implications and underscore their significance.

\subsection{Ramifications of the main results} \label{sec:Ramifications}

In summary, our primary objective is to validate the generalization power of GNNs compared to that documented for NNs, particularly over basic Euclidean cubes. This domain choice is deliberate, since, at a local level, a manifold is geometrically Euclidean. Therefore, our achievement in producing a GNN capable of interpolating bandlimited functions on a centered unit cube should be considered a critical initial step in showcasing the effectiveness of these sophisticated network structures on complex manifolds and their approximate graphs, establishing a baseline for their learning performance over such settings. 

We would like to highlight some flexibility afforded by the use of the Whittaker-Kotel'nikov-Shannon sampling theory. First, when functional values are sampled at a frequency surpassing the so-called \textit{Nyquist} rate (for a definition, see \cite[Subsection~2.1.A]{zayed2018advances}), our findings can be extended to lattices with missing nodes. 
This enables a clear and straightforward approach to simulate the effect of non-uniform sampling on a Euclidean cube.
\footnote{On a related note, \cite[Theorem~1.9]{benedetto2012modern} provides a reconstruction formula for non-uniform sampling that hinges on the existence of a suitable \textit{Fourier frame}. Such existence can be guaranteed by \cite[Theorem~I]{duffin1952class}, which essentially requires the sampled points to exhibit a certain density, a condition, however, naturally satisfied with oversampling.}
Second, it is also feasible to deploy our methodology to graphs embedded in a group structure, exploiting the Kluv'anek sampling theorem, Theorem~\ref{Ksampthm} in \S\ref{appx:sampling}. In addition, we should note that our specific context serves as an illustration and is not intended to capture the full spectrum of possibilities for more contemporary sampling applications in the study of neural networks. 

Lastly, we mention that, in a practical node classification task, a GNN undertakes semi-supervised learning, incorporating a small set of labeled data with a larger pool of unlabeled data for training. The objective is to predict labels for all nodes based on the observed ones. Typically, this involves adjusting the network parameter \eqref{parameter} through an optimization process, aiming to align the network output closely with the labeled data while maintaining a specified level of proximity measured by a predefined metric. We emphasize that our approach does not engage in any such learning. Instead, we directly derive the network parameter using tools from sampling theory, thus providing a pre-trained network through analytical means.

\subsection{Proof outline} \label{sec:mockproof}

Our road map to proving the central result, Theorem~\ref{thm:actualGNN}, goes as follows. First, the reconstruction result from uniform sampling, Theorem~\ref{thm:samp}, is pivotal to proving the generalization power of proxy GNN in dimension one:
\begin{equation} \label{samptoproxyGNN}
    \text{reconstruction from sampling in dimension one } \mapsto \quad \text{proxy GNN } (\text{Theorem}~\ref{thm:proxyGNN}).
\end{equation}
Although the proof of this step is given in \S\ref{sec:proxyGNNthm}, we will present a simplified version below, particularly highlighting how a sampling principle is embedded in the proxy GNN structure \eqref{proxygnnlayer}. In the next step, we will demonstrate that the continuous structure of the proxy GNN can be discretized, using the closed Newton-Cotes quadrature formula \cite[Subsection~9.3]{quarteroni2010numerical}, to form a GNN structure. This analysis will be done in \S\ref{sec:actualGNNpfdimone} and conducted in dimension one.
\begin{equation} \label{GNNindimone}
    \text{proxy GNN in dimension one } \xmapsto{\text{numerical integration}} \quad \text{GNN in dimension one.}
\end{equation}
To extend the one-dimensional version of Theorem~\ref{thm:actualGNN} to its $d$-dimensional counterpart, we introduce an approximate product structure that can be replicated by a ReLU NN, as Lemma~\ref{lem:product} in \S\ref{sec:actualGNNpfhighdims}. This lemma is a direct consequence of \cite[Proposition~3]{yarotsky2017error}. When combined with \eqref{GNNindimone}, this final step allows us to successfully conclude Theorem~\ref{thm:actualGNN} in all dimensions:
\begin{equation} \label{productstep}
    \text{GNN in dimension one} \quad
    \xmapsto{\text{approximate product}} \quad \text{GNN in dimension } d \,\, (\text{Theorem}~\ref{thm:actualGNN}).
\end{equation}
With that being outlined, we now return to \eqref{samptoproxyGNN}, as promised. 

\paragraph{A mock proof of \eqref{samptoproxyGNN}.} Suppose $f\in B^2_{\pi}$ for simplicity. Then from Theorem~\ref{WKSthm} 
\begin{equation} \label{series}
    f(x)=\sum_{n\in\mathbb{Z}}f(n)\sinc(x-n), \quad\forall x\in\mathbb{R}.
\end{equation}
Truncating the series in \eqref{series} into a finite sum from $n=-N$ to $n=N$, we note
\begin{equation} \label{endgame}
    \sum_{n=-N}^{N}f(n)\sinc(x-n)\overset{N\to\infty}{\longrightarrow}\sum_{n\in\mathbb{Z}}f(n)\sinc(x-n)=f(x).
\end{equation}
Suppose for a moment that the following integration-by-parts result holds 
\begin{equation} \label{intbp}
    \delta_{X}\ast\sinc(x) =\mathrm{ReLU}_{X}\ast (\sinc)''(x)
\end{equation} 
where $\mathrm{ReLU}_X$ and $\delta_{X}$ have been defined in \S\ref{sec:Preliminaries}. Then we can further deduce from \eqref{endgame} that
\begin{align}
    \label{endgamelab} \sum_{n=-N}^{N}f(n)\sinc(x-n) &=\bigg(\sum_{n=-N}^{N}f(n)\delta_{n}\bigg)\ast\sinc(x)\\
    \nonumber &=\sum_{n=-N}^{N}f(n)\mathrm{ReLU}_{n}\ast (\sinc)''(x)\overset{N\to\infty}{\longrightarrow}\sum_{n\in\mathbb{Z}}f(n)\sinc(x-n)=f(x).
\end{align}
Observe that \eqref{endgamelab} bears a resemblance to the proxy GNN structure delineated in \eqref{proxygnnlayer}. Indeed:
\begin{equation} \label{shannon_gnn}
    \begin{split}
        x \mapsto \{x-n\}_{n=-N}^{N} \xmapsto{\mathrm{act.}} \{\mathrm{ReLU}(x-n)\}_{n=-N}^{N}  &\xmapsto{(\sinc)''\ast} \{\mathrm{ReLU}_{n}\ast (\sinc)''(x)\}_{n=-N}^{N}\\
        &\mapsto \sum_{n=-N}^{N}f(n)\mathrm{ReLU}_{n}\ast (\sinc)''(x). 
    \end{split}
\end{equation}
Here, the activation used is the ReLU activation, and the convolution kernel $\mathcal{K}$ is $(\sinc)''$. The nonunital network weights are precisely $\{f(n)\}_{n=-N}^{N}$, the sampled functional values. Presenting it in this manner (\eqref{endgamelab}, \eqref{shannon_gnn}), we can clearly see how the sampling principle \eqref{series} manifests itself in the proxy model. However, to render \eqref{endgamelab} mathematically rigorous and computationally efficient, certain issues require attention. First, \eqref{intbp} may not hold for the $\sinc$ function, and Second, due to its heavy-tailed nature, the convergence in \eqref{series} is only of order $O(1/N)$, both of which necessitate the use of a function exhibiting better decay properties. To address these, we employ a \textit{scaled-truncated regularized} version of the series in \eqref{series}:
\begin{equation} \label{nuistseries}
    \sum_{n=-N}^{N}f(hn)\sinc{}_{h}\cdot\mathcal{G}_{M,r,h}(x-hn).
\end{equation}
The notations involved have been defined at the beginning of \S\ref{sec:Results}; particularly, $\mathcal{G}_{M,r,h}=\mathcal{G}_{M,r}(h^{-1}\cdot)$ in \eqref{def:truncGaussian}. The term ``regularized" pertains to the multiplication of $\sinc$ by a Gaussian in \eqref{nuistseries}, ``truncated" to the truncation of the Gaussian, and ``scaled" to the scaling applied to both $\sinc$ and the Gaussian. As indicated in Theorem~\ref{thm:proxyGNN}, we will assume that $M\in 4\mathbb{Z}_+$ and set $N=M/2$, $r=\sqrt{M}$, $h= 4/M$. We will now modify \eqref{endgamelab} to obtain an updated proxy GNN model as follows. 

Suppose $f\in B^2_\pi$ and that we have sampled 
\begin{equation*} 
    {\bf S}_f := \begin{bmatrix} f(-2) & f(-2+4/M) & \cdots & f(-4/M) & f(0) & f(4/M) &\cdots & f(2-4/M) & f(2) \end{bmatrix}.
\end{equation*}
We design a proxy GNN $\Psi_f$ that uses only the knowledge of ${\bf S}_f$ to predict $f$ on the entire interval $[-1,1]$. The model uses ReLU activations and the convolution 
\begin{equation}\label{frakF}
    (\sinc{}_h\cdot\mathcal{G}_{M,r,h})''\ast g(x) = \int g(x-y)(\sinc{}_h\cdot\mathcal{G}_{M,r,h})''(y)\,dy.
\end{equation}
Comparing \eqref{frakF} with \eqref{locfil}, one sees that the convolution kernel $\mathcal{K} = (\sinc{}_h\cdot\mathcal{G}_{M,r,h})''$. The model $\Psi_f$ is comprised of two hidden layers, a filter layer and an NN layer.

\begin{enumerate}
    \item {\bf Input:} The input for our proxy GNN is the function $g_0(x)= x$, $x\in\mathbb{R}$.
    \item {\bf First hidden layer:} Given the input $g_0$, the output function $g_1$ of the first layer is given by bias-shifting with respect to the sampling locations, then an application of component-wise ReLU activation, followed by a convolution:
    \begin{equation*} 
        \begin{split}
            g_0(x) &\equiv x \mapsto \begin{bmatrix} x+2 \\ x+2-4/M \\ \vdots \\ x-2+4/M \\ x-2 \end{bmatrix} \xmapsto{\mathrm{act.}} \begin{bmatrix} \mathrm{ReLU}(x+2) \\ \mathrm{ReLU}(x+2-4/M) \\ \vdots \\ \mathrm{ReLU}(x-2+4/M) \\ \mathrm{ReLU}(x-2)\end{bmatrix} \\
            & \xmapsto{(\sinc{}_h\cdot\mathcal{G}_{M,r,h})''\ast} \begin{bmatrix} \mathrm{ReLU}_{-2}\ast(\sinc{}_h\cdot\mathcal{G}_{M,r,h})''(x) \\ \mathrm{ReLU}_{-2+4/M}\ast(\sinc{}_h\cdot\mathcal{G}_{M,r,h})''(x) \\ \vdots \\ \mathrm{ReLU}_{-2+4/M}\ast(\sinc{}_h\cdot\mathcal{G}_{M,r,h})''(x) \\ \mathrm{ReLU}_{-2}\ast(\sinc{}_h\cdot\mathcal{G}_{M,r,h})''(x)\end{bmatrix} \equiv \vec{g}_1(x), \quad\forall x\in\mathbb{R}.
        \end{split}    
    \end{equation*}
    \item{\bf Second hidden layer:} Taking $\vec{g}_1$ as the input for the second layer, the output $g_2$ is given by
    \begin{equation} \label{summingstep}
         \vec{g}_1(x) \mapsto {\bf S}_f \vec{g}_1(x) = \sum_{n=-M/2}^{M/2} f(hn)\mathrm{ReLU}_{hn}\ast(\sinc{}_h\cdot\mathcal{G}_{M,r,h})''(x)\equiv g_2(x), \quad\forall x\in\mathbb{R}.
    \end{equation}
    \item{\bf Output:} The output of our GNN is the function $g_2(x)$, $x\in\mathbb{R}$.
\end{enumerate}

Note from \eqref{summingstep} that $\vec{g}_1$ used in the matrix multiplication performed in the second layer is a vector-valued function on $\mathbb{R}$, and the weight matrix ${\bf W}={\bf S}_f$. Having now elucidated \eqref{samptoproxyGNN} as well as presented our proxy GNN model explicitly, we give additional remarks concerning the output $g_2$. By a principle closely resembling the integration-by-parts principle exemplified in \eqref{intbp}, we obtain
\begin{equation} \label{outputg2}
    \begin{split}
        g_2(x) &=\sum_{n=-M/2}^{M/2} f(hn)\mathrm{ReLU}_{hn}\ast(\sinc{}_h\cdot\mathcal{G}_{M,r,h})''(x) \\
        &= \sum_{n=-M/2}^{M/2}f(hn)\sinc{}_{h}\cdot\mathcal{G}_{M,r,h}(x-hn) + \text{error}_1, \quad\forall x\in [-1,1].
    \end{split}   
\end{equation}
Then by virtue of Theorem~\ref{thm:samp} and the convenient parameter selection of $h, r$, we can assert that
\begin{equation} \label{nearlyblm}
    \sum_{n=-M/2}^{M/2}f(hn)\sinc{}_{h}\cdot\mathcal{G}_{M,r,h}(x-hn) = f(x) + \text{error}_2, \quad\forall x\in [-1,1].
\end{equation}
Furthermore, the displayed errors, $\text{error}_1, \text{error}_2$, can be shown to be exponentially decaying in terms of $M$. 

\begin{remark}
We give a brief discussion about the replication of product structure mentioned earlier in \eqref{productstep}. Let $\vec{x}=(x_1,\cdots,x_d)\in\mathbb{R}^d$, $\vec{n}=(n_1,\cdots,n_d)\in\mathbb{Z}^d$. Then in a straightforward manner, the $d$-dimensional version of $g_2$ from \eqref{outputg2} is
\begin{equation*} 
    \sum_{\vec{n}\in [-M/2,M/2]^{d}}f(h\vec{n}) \prod_{j=1}^{d} \mathrm{ReLU}_{n_{j}}\ast (\sinc{}_{h}\cdot\mathcal{G}_{M,r,h})''(x_{j}).
\end{equation*}
This indicates that the product we will need to approximate is the product of the one-dimensional discretizations described in step \eqref{GNNindimone}. 
\end{remark}

\begin{remark}[Nature of the graph filter] \label{rem:filter}
As suggested by \eqref{outputg2}, \eqref{nearlyblm}, our graph filter, can be considered a mixed spatial-spectral filter, in that it possesses both spatial and (nearly) spectral filtering properties. Its spatial filtering function is clear, due to the spatial truncation. Its spectral filtering function is more subtle, but it is evidenced in the fact that it is designed for the task of approximating bandlimited functions. 
\end{remark}

\begin{remark}[A notation convention] \label{rem:notconv}
A recurring theme in this paper is the establishment of results in one dimension, followed by an effort to generalize them to higher dimensions. As a result, we employ the convention that structures, when similar to each other except for their domain dimensions, will be denoted with the same symbols. Their distinction will rely on the context and whether their input argument is a scalar or a vector. This streamlined approach minimizes the proliferation of notations in this paper. An example has been given in \eqref{Op3}, where if $d=1$, then we write $\mathcal{R}_{M,r,h}f(x)$ and if $d>1$, $\mathcal{R}_{M,r,h}f(\vec{x})$. Another example would be the function $\mu_{M,r,h}$, whose version in one dimension, $\mu_{M,r,h}(x)$, is given in \eqref{def:muMrh}, and whose version in higher dimensions, $\mu_{M,r,h}(\vec{x})$ is given in \eqref{def:muhigh}. 
\end{remark}

\section{Proof of Theorem~\ref{thm:samp}} \label{sec:sampthmpf}

We initiate our proof with the pivotal case of $d=1$, since it carries immense importance and since higher dimensions will merely involve a straightforward extension of the one-dimensional outcome.

\subsection{Proof of Theorem~\ref{thm:samp} in dimension one} \label{sec:sampthmpfdim1}

Let $h,r>0$ and $M\in \mathbb{Z}_{+}$. Supposing $\sigma<\pi$, we define the mapping $\mathcal{R}_{M,r}$ acting on $f\in B^2_\sigma$ as follows,
\begin{equation} \label{Op2}
    \mathcal{R}_{M,r}f(x) :=\sum_{n\in\mathbb{Z}}f(n)\sinc(x-n)\mathcal{G}_{M,r}(x-n),
\end{equation}
where $\mathcal{G}_{M,r}$ is defined in \eqref{def:truncGaussian}. Observe the similarity between $\mathcal{R}_{M,r}$ and $\mathcal{R}_{M,r,h}$ in \eqref{Op3}; particularly, $\mathcal{R}_{M,r,h}$ becomes $\mathcal{R}_{M,r}$ if we select the scaling parameter $h=1$. For this reason, we call $\mathcal{R}_{M,r}, \mathcal{R}_{M,r,h}$ \textit{truncated regularized reconstruction} and \textit{scaled-truncated regularized reconstruction} mappings, respectively and aptly. Define the following error mappings associated with $\mathcal{R}_{M,r,h}, \mathcal{R}_{M,r}$ respectively,
\begin{equation} \label{def:errors}
    \mathcal{E}_{M,r,h} := \mathcal{R}_{M,r,h} - Id \quad\text{ and }\quad \mathcal{E}_{M,r} := \mathcal{R}_{M,r} - Id.
\end{equation}
To establish Theorem~\ref{thm:samp}, we want to demonstrate that the error $\mathcal{E}_{M,r,h}f$ decays in a, quantifiably, geometric fashion. However, since $\mathcal{R}_{M,r}$ (or $\mathcal{E}_{M,r}$) is expressively $\mathcal{R}_{M,r,h}$ (or $\mathcal{E}_{M,r,h}$) when there is no scaling involved, our approach is to initiate our analysis with the former mapping to streamline the notation and reasoning; i.e., we derive a version of Theorem~\ref{thm:samp}, Proposition~\ref{prop:sampthm}, that is tailored for $\mathcal{E}_{M,r}$, and subsequently modify it to encompass $\mathcal{E}_{M,r,h}$. Finally, we select the appropriate parameters $M,r,h$ and conclude Theorem~\ref{thm:samp}.

\begin{remark}
Much of the analysis presented here draws inspiration from \cite{liwen2004regularized}. We have opted to reconstruct some of their results in our analysis for several reasons. First, their convention differs slightly from ours. Second, some essential prerequisites and nuances that are necessary for their theory were left unaddressed. Third, we find discrepancies and inaccuracies in several places. Lastly, certain crucial points for our purpose were either not explicitly stated or insufficiently expounded upon in the said paper.
\end{remark}

As mentioned, we suppose $\sigma<\pi$ and focus on $\mathcal{E}_{M,r}$ in \eqref{def:errors}. Then according to Theorem~\ref{WKSthm} (and the remarks that follow), it holds for $f\in B^2_\sigma\subset B^2_\pi$ that
\begin{equation} \label{sigmaequalpicited}
    f(x) = \sum_{n\in\mathbb{Z}}f(n)\sinc(x-n),
\end{equation}
where the series converges compactly on $\mathbb{R}$ and in $L^2(\mathbb{R})$. Thus, since the Gaussian truncation $\mathcal{G}_{M,r}$ effectively limits the number of terms contributing to the series defining $\mathcal{R}_{M,r}f$ for each $x\in\mathbb{R}$, we can as well deduce that $\mathcal{R}_{M,r}f\in L^2(\mathbb{R})$ and consequently, $\mathcal{E}_{M,r}: B^2_\sigma\to L^2$. However, we will need a more robust conclusion. To achieve this, following the approach in \cite{liwen2004regularized}, we define for $w\in\mathbb{R}$,
\begin{align}
    \label{mults} \mu_{M,r}(w) &:=\frac{1}{2\pi}\int_{t\in w+[-\pi,\pi]}\widehat{\mathcal{G}_{M,r}}(t)\,dt=\frac{\mathcal{F}(\sinc\cdot\mathcal{G}_{M,r})(w)}{2\pi}\\
    \label{nults} \nu_{M,r}(w) &:=\frac{1}{2\pi}\int_{t\not\in w+[-\pi,\pi]}\widehat{\mathcal{G}_{M,r}}(t)\,dt.
\end{align}
Evidently, both $\mu_{M,r} \in L^2\cap L^{\infty}, \nu_{M,r}\in L^{\infty}$ are even, real-valued, continuous functions. By the Fourier inversion theorem \eqref{foudef2}
\begin{equation*}
    \frac{1}{2\pi}\int_{\mathbb{R}}\widehat{\mathcal{G}_{M,r}}(t)\,dt=\mathcal{G}_{M,r}(0)=1,
\end{equation*}
which, in conjunction with \eqref{mults}, \eqref{nults}, leads to
\begin{equation} \label{POU}
    \mu_{M,r}+\nu_{M,r}\equiv 1\quad\text{ and }\quad \sum_{n\in\mathbb{Z}\setminus\{0\}}\mu_{M,r}(w+2n\pi)=\nu_{M,r}(w).
\end{equation}
This says that $2n\pi$ translations of $\mu_{M,r}$ form a partition of unity on $\mathbb{R}$. The following quantities will prove to be crucial in later stages
\begin{equation*}
    \sup_{w\in [-\sigma,\sigma]} |\nu_{M,r}(w)| \quad\text{ and }\quad \sup_{w\in [-\sigma,\sigma]} \sum_{n\in\mathbb{Z}\setminus\{0\}} |\mu_{M,r}(w+2n\pi)|.
\end{equation*}
Therefore, we temporarily record the proposition below, whose proof will be given after the proof of Proposition~\ref{prop:sampthm}. 

\begin{proposition} \label{prop:2crucialsizes} Let $\sigma<\pi$. Let $\mu_{M,r}, \nu_{M,r}$ be as in \eqref{mults}, \eqref{nults}, respectively. Then it holds that
\begin{equation} \label{maxnu} 
    \sup_{w\in [-\sigma,\sigma]} |\nu_{M,r}(w)| = \max_{w\in [-\sigma,\sigma]} |\nu_{M,r}(w)| \lesssim \frac{\exp(-r^2(\pi-\sigma)^2/2)}{r(\pi-\sigma)} + \frac{r^2}{M}\exp\bigg(-\frac{M^2}{2r^2}\bigg),
\end{equation}
and that
\begin{align}
    \label{maxmu} \sup_{w\in [-\sigma,\sigma]} \sum_{k\in\mathbb{Z}\setminus\{0\}} |\mu_{M,r} (w+2k\pi)| &= \max_{w\in [-\sigma,\sigma]} \sum_{k\in\mathbb{Z}\setminus\{0\}} |\mu_{M,r}(w+2k\pi)| \\ 
    \nonumber &\lesssim \frac{\exp(-r^2(\pi-\sigma)^2/2)}{r(\pi-\sigma)} + c(r)\exp\bigg(-\frac{M^2}{4r^2}\bigg),
\end{align}
where $c(r):=\max\{1,\frac{1}{r},\frac{1}{r^2}\}$.
\end{proposition}

\begin{remark}
The result \eqref{maxmu} of Proposition~\ref{prop:2crucialsizes}, which was not established in \cite{liwen2004regularized}, is pivotal: without it, the framework presented in \cite{liwen2004regularized} would lack cohesion, highlighting the mentioned oversight.
\end{remark}

Given Proposition~\ref{prop:2crucialsizes} in our possession, with fixed $M,r$ and $\sigma<\pi$, we confirm a precise bound for the mapping $\mathcal{E}_{M,r}$. 

\begin{proposition} \label{prop:sampthm}
Let $\sigma<\pi$. Let $\mathcal{E}_{M,r}$ be as in \eqref{def:errors} and $f\in B^2_\sigma$. Then
\begin{align}
    \label{EfL2bd} \|\mathcal{E}_{M,r}f\|_2 &\lesssim \epsilon(\sigma,M,r)\|f\|_2\\
    \label{EfLinftybd} \|\mathcal{E}_{M,r}f\|_{\infty} &\lesssim \sqrt{\sigma}\,\epsilon(\sigma,M,r)\|f\|_2,
\end{align}
where
\begin{equation*} 
    \epsilon(\sigma,M,r) := \frac{\exp(-r^2(\pi-\sigma)^2/2)}{r(\pi-\sigma)} + \frac{r^2}{M}\exp\bigg(-\frac{M^2}{2r^2}\bigg) + c(r)\exp\bigg(-\frac{M^2}{4r^2}\bigg).
\end{equation*}
\end{proposition}

\begin{proof}
Since $f\in B^2_{\sigma}$, $\hat{f}$ is a continuous, compactly supported on $[-\sigma,\sigma]\subset [-\pi,\pi]$ with bounded variation. If we temporarily consider $\hat{f}$ a $2\pi$-periodic function, then by the Dirichlet-Jordan test \cite[Theorem~2.23]{chui1992introduction},
\begin{equation*} 
    \lim_{N\to\infty} 
    \bigg(\sum_{n=-N}^N f(n)e^{-inw}\bigg)=\sum_{n\in\mathbb{Z}}f(n)e^{-inw}=\sum_{n\in\mathbb{Z}}f(-n)e^{inw}=\hat{f}(w) 
\end{equation*}
uniformly on any compact subinterval of $[-\sigma,\sigma]$ and $[-\pi,-\sigma]\cup [\sigma,\pi]$. Therefore, we can write
\begin{equation} \label{blmtdFt}
    \hat{f}(w) = \sum_{n\in\mathbb{Z}}f(n)e^{-inw}\chi_{[-\pi,\pi]}(w).
\end{equation}
Alternatively \eqref{blmtdFt} can be deduced by applying Fourier transform to \eqref{sigmaequalpicited}; see also the derivation of \eqref{Kluscaledf} below. It follows that, if $w\in [-\pi,\pi]$, then
\begin{equation} \label{casepi}
    \sum_{n\in\mathbb{Z}}f(n)e^{-inw} = \hat{f}(w),
\end{equation}
 and if $w=u+2k\pi$ for some $u\in [-\pi,\pi]$ and $k\in\mathbb{Z}\setminus\{0\}$,
\begin{equation} \label{caseoutpi}
    \sum_{n\in\mathbb{Z}}f(n)e^{-inw} = \sum_{n\in\mathbb{Z}}f(n)e^{-in(u+2k\pi)} = \hat{f}(u).
\end{equation}
With $\mathcal{E}_{M,r}: B^2_\sigma\to L^2$ having been previously established, we can take the Fourier transform of $\mathcal{E}_{M,r}f$ to obtain for $w\in\mathbb{R}$,
\begin{equation} \label{hatEgenexpression}
    \widehat{\mathcal{E}_{M,r}f}(w) = (\hat{f}-\widehat{\mathcal{R}_{M,r}f})(w)=\sum_{n\in\mathbb{Z}}f(n)e^{-iwn}\bigg(\chi_{[-\pi,\pi]}(w)-\mu_{M,r}(w)\bigg),
\end{equation}
where the last equality above is resulted from \eqref{blmtdFt}. Then by virtue of \eqref{POU}, \eqref{casepi}, \eqref{caseoutpi}, \eqref{hatEgenexpression} and the fact that $supp(\hat{f})\subset [-\sigma,\sigma]\subset [-\pi,\pi]$, we acquire
\begin{equation} \label{hatEderivation}
    \widehat{\mathcal{E}_{M,r}f}(w)=
        \begin{cases}
            \hat{f}(w)\nu_{M,r}(w) &\text{ if } w\in [-\sigma,\sigma]\\
            -\hat{f}(w-2k\pi)\mu_{M,r}(w) &\text{ if } w - 2k\pi\in [-\sigma,\sigma], \text{ for some } k\in\mathbb{Z}\setminus\{0\}\\
            0 &\text{ otherwise }
        \end{cases}.
\end{equation} 
Consequently,
\begin{equation*} 
    \|\widehat{\mathcal{E}_{M,r}f}\|^2_2 =\int_{[-\sigma,\sigma]}|\hat{f}(w)|^2\bigg(\nu_{M,r}^2(w)+\sum_{k\in\mathbb{Z}\setminus\{0\}}\mu_{M,r}^2(w+2k\pi)\bigg)\,dw,
\end{equation*}
which, combined with \eqref{normcompare} and the $\ell^p$-embedding property, allows us to infer that
\begin{equation} \label{EMrL2}
    \|\mathcal{E}_{M,r}f\|_2\leq \bigg(\max_{w\in [-\sigma,\sigma]} |\nu_{M,r}(w)| + \max_{w\in [-\sigma,\sigma]} \sum_{k\in\mathbb{Z}\setminus\{0\}} |\mu_{M,r}(w+2k\pi)|\bigg)\|f\|_2.
\end{equation} 
Applying Proposition~\ref{prop:2crucialsizes} to the right-hand-side in \eqref{EMrL2}, we deduce \eqref{EfL2bd}. 

Further, it also follows from Proposition~\ref{prop:2crucialsizes} that
\begin{equation*}
    \|\widehat{\mathcal{E}_{M,r}f}\|_1 \leq \int_{[-\sigma,\sigma]}|\hat{f}(w)|\bigg( |\nu_{M,r}(w)| +\sum_{n\in\mathbb{Z}\setminus\{0\}} |\mu_{M,r}(w+2n\pi)| \bigg)\,dw <\infty.
\end{equation*}
Thus, by \eqref{EMrL2}, the Fourier inversion theorem \eqref{foudef2} and \eqref{normcompare} again, 
\begin{equation*} 
    \begin{split}
        \|\mathcal{E}_{M,r}f\|_{\infty} &=\sup_{w\in\mathbb{R}}\bigg|\frac{1}{2\pi}\int_{\mathbb{R}}\widehat{\mathcal{E}_{M,r}f}(t)e^{iwt}\,dt\bigg| \\
        &\leq \sqrt{\sigma/\pi}\bigg(\max_{w\in [-\sigma,\sigma]} |\nu_{M,r}(w)| + \max_{w\in [-\sigma,\sigma]} \sum_{k\in\mathbb{Z}\setminus\{0\}} |\mu_{M,r}(w+2k\pi)|\bigg)\|f\|_2.
    \end{split} 
\end{equation*}
Applying Proposition~\ref{prop:2crucialsizes} once more, this time to the right-hand-side above, yields \eqref{EfLinftybd}, as desired.
\end{proof}

We now provide a proof of Proposition~\ref{prop:2crucialsizes}. Following this, we will be ready to establish Theorem~\ref{thm:samp}.

\begin{proof}[Proof of Proposition~\ref{prop:2crucialsizes}]
We first attend to \eqref{maxnu}. By definition \eqref{mults} and \eqref{POU},
\begin{align}
    \nonumber \nu_{M,r}(w) &= 1-\frac{1}{2\pi}\int_{t\in w+[-\pi,\pi]}\widehat{\mathcal{G}_{M,r}}(t)\,dt\\ 
    \nonumber &= 1-\frac{1}{2\pi}\int_{t\in w+[-\pi,\pi]}\bigg(\hat{\mathcal{G}}_r(t)-\int_{y\not\in [-M, M]}\mathcal{G}_{r}(y)e^{-iyt}\,dy\bigg)dt\\
    \nonumber &=\frac{1}{2\pi}\int_{t\not\in w+[-\pi,\pi]}\hat{\mathcal{G}}_{r}(t)\,dt +\frac{1}{2\pi}\int_{t\in w+[-\pi,\pi]}\int_{y\not\in [-M, M]}\mathcal{G}_{r}(y)e^{-iyt}\,dydt\\
    \label{lastexp} &=\frac{1}{2\pi}\int_{t\not\in r(w+[-\pi,\pi])}\hat{\mathcal{G}}(t)\,dt +\frac{1}{\pi}\int_{y\not\in [-M/r,M/r]}\mathcal{G}(y)\frac{\sin(ry\pi)\cos(ryw)}{y}\,dy
\end{align}
for $w\in\mathbb{R}$. The first term in \eqref{lastexp} serves as a \textit{regularization error}, while the second, containing $M$ in its expression, serves as a \textit{truncation error}. Direct computing gives,
\begin{equation} \label{maxnu_M}
    \begin{split}
        \max_{[-\sigma,\sigma]} |\nu_{M,r}(w)| &\leq\max_{[-\sigma,\sigma]} \frac{1}{2\pi}\int_{t\not\in r(w+[-\pi,\pi])}\hat{\mathcal{G}}(t)\,dt + \int_{y\not\in [-M,M]}\mathcal{G}_{r}(y)\,dy \\
        &=: \max_{[-\sigma,\sigma]} e_r(w) + e_{M,r},
    \end{split}
\end{equation}
and that $\max_{[-\sigma,\sigma]} e_r(w)$ can only be attained at $w=0,\pm\sigma$. However, since $e_r$ is an even function, and $\frac{d^2}{dw^2} e_r(0)>0$, it remains that
\begin{equation} \label{maxatsigma}
    \begin{split}
        \max_{[-\sigma,\sigma]} e_r(w) = e_r(\sigma) &= \frac{1}{2\pi}\int_{r\pi-r\sigma}^\infty \exp\bigg(-\frac{t^2}{2}\bigg)\,dt + \frac{1}{2\pi} \int_{r\sigma+r\pi}^\infty \exp\bigg(-\frac{t^2}{2}\bigg)\,dt\\
        &\leq \frac{1}{\pi} \int_{r\pi-r\sigma}^\infty \exp\bigg(-\frac{t^2}{2}\bigg)\,dt.
    \end{split}
\end{equation}
To harness \eqref{maxatsigma} effectively, we make use of the following Mills' ratio \cite{boyd1959inequalities}
\begin{equation} \label{Mratio}
    \int_{x}^{\infty}\exp\bigg(-\frac{t^2}{2}\bigg)\,dt < \frac{\pi \exp(-x^2/2)}{\sqrt{(\pi-2)^2x^2+2\pi}+2x}, \quad \forall x>0;
\end{equation}
therefore \eqref{maxatsigma} yields
\begin{equation} \label{er}
    \max_{w\in [-\sigma,\sigma]} e_r(w)\lesssim \frac{\exp(-r^2(\pi-\sigma)^2/2)}{r(\pi-\sigma)}.
\end{equation}
Now regarding $e_{M,r}$ in \eqref{maxnu_M}, we also have as a direct result of \eqref{Mratio},
\begin{equation} \label{eMr}
    e_{M,r} =\int_{y\not\in [-M,M]}\mathcal{G}_{r}(y)\,dy = r\int_{y\not\in [-M/r,M/r]}\mathcal{G}(y)\,dy\lesssim\frac{r^2}{M}\exp\bigg(-\frac{M^2}{2r^2}\bigg).
\end{equation}
Combining \eqref{er}, \eqref{eMr}, we obtain \eqref{maxnu}. 

We turn to \eqref{maxmu} next. By definitions \eqref{sinc}, \eqref{mults}
\begin{equation} \label{strings}
    \begin{split}
        \sum_{k\in\mathbb{Z}\setminus\{0\}} |\mu_{M,r}(w+2k\pi)| &= \frac{1}{2\pi}\sum_{k\in\mathbb{Z}\setminus\{0\}} \bigg|\int_{w-\pi+2k\pi}^{w+\pi + 2k\pi} \int_{-M}^M\mathcal{G}_r(x)e^{-itx}\,dxdt\bigg|\\
        &= \frac{1}{2\pi}\sum_{k\in\mathbb{Z}\setminus\{0\}} \bigg|\int_{-M}^M\mathcal{G}_r(x)e^{-i(w+2k\pi)x}\int_{-\pi}^{\pi} e^{-itx}\,dtdx\bigg|\\
        &= \frac{1}{2\pi}\sum_{k\in\mathbb{Z}\setminus\{0\}} \bigg|\int_{-M}^M\mathcal{G}_r(x)e^{-i(w+2k\pi)x}\int_{-\pi}^{\pi} e^{-itx}\,dtdx\bigg|\\
        &= \sum_{k\in\mathbb{Z}\setminus\{0\}} \bigg|\int_{-M}^M\mathcal{G}_r(x)\cos((w+2k\pi)x)\sinc(x)\,dx\bigg|.
    \end{split}
\end{equation}
Observe that, had $M=\infty$ (i.e., no truncation), then working \eqref{strings} backward, we would get, for any $w'\in\mathbb{R}$
\begin{equation} \label{notruncnonneg}
    \bigg|\int_\mathbb{R} \mathcal{G}_r(x)\cos(w'x)\sinc(x)dx\bigg| = \frac{1}{2\pi} \bigg|\int_{w'-\pi}^{w'+\pi} \int_\mathbb{R} \mathcal{G}_r(x)e^{-itx}\,dxdt\bigg| = \frac{1}{2\pi}\int_{w'-\pi}^{w'+\pi} \hat{\mathcal{G}}_r(t)\,dt\geq 0,
\end{equation}
as the Fourier transform of a Gaussian is another Gaussian. Moreover, following the steps from \eqref{mults} to \eqref{POU} using this particular value of $M$ and introducing for $w\in\mathbb{R}$
\begin{equation*}
    \mu_r(w) :=\frac{1}{2\pi}\int_{t\in w+[-\pi,\pi]}\hat{\mathcal{G}}_r(t)\,dt \quad\text{ and }\quad
    \nu_r(w) :=\frac{1}{2\pi}\int_{t\not\in w+[-\pi,\pi]}\hat{\mathcal{G}}_r(t)\,dt,
\end{equation*}
we would also obtain
\begin{equation*}
    \sum_{k\in\mathbb{Z}\setminus\{0\}} \mu_r(w+2k\pi) = \nu_r(w),
\end{equation*}
which, by the nonnegativity of $\mu_r$ depicted in \eqref{notruncnonneg}, further implies
\begin{equation*}
    \begin{split}
        \sum_{k\in\mathbb{Z}\setminus\{0\}} \bigg|\int_\mathbb{R} \mathcal{G}_r(x)\cos((w+2k\pi)x)\sinc(x)dx\bigg| &= \sum_{k\in\mathbb{Z}\setminus\{0\}} |\mu_r(w+2k\pi)|\\
        &= \sum_{k\in\mathbb{Z}\setminus\{0\}} \mu_r(w+2k\pi) = \nu_r(w)\geq 0
    \end{split}
\end{equation*}
for every $w\in\mathbb{R}$. Therefore, through arguments analogous to those outlined from \eqref{lastexp} to \eqref{er}, we derive
\begin{equation*}
    \begin{split}
        \sup_{w\in [-\sigma,\sigma]} \sum_{k\in\mathbb{Z}\setminus\{0\}} \bigg|\int_\mathbb{R} \mathcal{G}_r(x)\cos((w+2k\pi)x)\sinc(x)dx\bigg| &= \max_{w\in [-\sigma,\sigma]} \nu_r(w)\\
        &\leq \max_{w\in [-\sigma,\sigma]} e_r(w) \lesssim \frac{\exp(-r^2(\pi-\sigma)^2/2)}{r(\pi-\sigma)}.
    \end{split}
\end{equation*}
Applying this finding to \eqref{strings} produces
\begin{multline} \label{nuseries}
    \sup_{w\in [-\sigma,\sigma]} \sum_{k\in\mathbb{Z}\setminus\{0\}} |\mu_{M,r}(w+2k\pi)| \\
    \lesssim \frac{\exp(-r^2(\pi-\sigma)^2/2)}{r(\pi-\sigma)} + \sup_{w\in [-\sigma,\sigma]} \sum_{k\in\mathbb{Z}\setminus\{0\}}\bigg|\int_M^\infty \mathcal{G}_r(x)\cos((w+2k\pi)x)\sinc(x)\,dx\bigg|,
\end{multline}
and so we are left with calculating the second term on the right-hand-side above. Note, through integration-by-parts,
\begin{align*}
    &\int_M^\infty \mathcal{G}_r(x)\cos((w+2k\pi)x)\sinc(x)\,dx \\
    &= - \int_M^\infty \frac{d}{dx}(\mathcal{G}_r(x)\sinc(x))\frac{\sin((w+2k\pi)x)}{w+2k\pi}\,dx + \mathcal{G}_r(x)\sinc(x)\frac{\sin((w+2k\pi)x)}{w+2k\pi}\bigg|_{x=M}^{x=\infty} \\
    &= - \int_M^\infty \frac{d}{dx}(\mathcal{G}_r(x)\sinc(x))\frac{\sin((w+2k\pi)x)}{w+2k\pi}\,dx\\
    &= - \int_M^\infty \frac{d^2}{dx^2}(\mathcal{G}_r(x)\sinc(x))\frac{\cos((w+2k\pi)x)}{(w+2k\pi)^2}\,dx + \frac{d}{dx}(\mathcal{G}_r(x)\sinc(x))\frac{\cos((w+2k\pi)x)}{(w+2k\pi)^2}\bigg|_{x=M},
\end{align*}
where the second equality is due to that $\sinc(M)=0$, as $M\in\mathbb{Z}_+$. For $w\in [-\sigma,\sigma]$, where $\sigma<\pi$, we take note that
\begin{equation} \label{boundarynterm}
    \sum_{k\in\mathbb{Z}\setminus\{0\}} \bigg|\frac{d}{dx}(\mathcal{G}_r(x)\sinc(x))\frac{\cos((w+2k\pi)x)}{(w+2k\pi)^2}\bigg|_{x=M}\bigg|\lesssim \exp\bigg(-\frac{M^2}{2r^2}\bigg),
\end{equation}
and that, through straightforward calculations, 
\begin{equation} \label{integralnterm}
    \sum_{k\in\mathbb{Z}\setminus\{0\}} \bigg| \int_M^\infty \frac{d^2}{dx^2}(\mathcal{G}_r(x)\sinc(x))\frac{\cos((w+2k\pi)x)}{(w+2k\pi)^2}\,dx\bigg| \lesssim c(r)\exp\bigg(-\frac{M^2}{4r^2}\bigg).
\end{equation}
By bringing together \eqref{nuseries}, \eqref{boundarynterm}, \eqref{integralnterm}, we arrive at \eqref{maxmu}, as needed.
\end{proof}

With all the preliminary steps completed, we are now in the position to prove Theorem~\ref{thm:samp} in dimension one, as promised. 

\begin{proof}[Proof of Theorem~\ref{thm:samp} in dimension one] Let us initially suppose $\sigma<\pi/h$ for some $h>0$ small. Then it can be verified from Theorem~\ref{WKSthm} that for $f\in B^2_\sigma\subset B^2_{\pi/h}$, 
\begin{equation*}
    f(x) = \sum_{n\in\mathbb{Z}}f(hn)\sinc{}_h(x-hn), \quad\forall x\in\mathbb{R},
\end{equation*}
whose Fourier transform yields
\begin{equation} \label{Kluscaledf}
    \hat{f}(w) = h\sum_{n\in\mathbb{Z}}f(hn)e^{-iwhn}\chi_{[-\pi, \pi]}(hw), \quad\forall w\in\mathbb{R}.
\end{equation}
Similarly as before, we derive for $w\in\mathbb{R}$
\begin{equation} \label{KluscaledRMrhf}
    \widehat{\mathcal{R}_{M,r,h}f}(w) = h\sum_{n\in\mathbb{Z}} f(hn)e^{-iwhn}\mu_{M,r,h}(w), 
\end{equation}
where 
\begin{equation} \label{def:muMrh}
    \mu_{M,r,h}(w) := \frac{1}{2\pi}\int_{t\in hw+[-\pi,\pi]} \widehat{\mathcal{G}_{M,r}}(t)\,dt = \frac{\mathcal{F}(\sinc\cdot\mathcal{G}_{M,r})(hw)}{2\pi} = \mu_{M,r}(hw).
\end{equation}
Therefore, as a consequence of \eqref{Kluscaledf}, \eqref{KluscaledRMrhf}, it holds that
\begin{equation*} 
    \widehat{\mathcal{E}_{M,r,h}f}(w) = h\sum_{n\in\mathbb{Z}}f(hn)e^{-iwhn}\bigg(\chi_{[-\pi, \pi]}(hw) - \mu_{M,r}(hw)\bigg),
\end{equation*}
from which we deduce the following fact akin to \eqref{hatEderivation}
\begin{equation*}
    \widehat{\mathcal{E}_{M,r,h}f}(w) = 
    \begin{cases}
        \hat{f}(w)\nu_{M,r}(hw) &\text{ if } w\in [-\sigma,\sigma]\\
        -\hat{f}(w-2k\pi/h)\mu_{M,r}(hw) &\text{ if } w-2k\pi/h \in [-\sigma, \sigma], \text{ for some } k\in\mathbb{Z}\setminus\{0\}\\
        0 &\text{ otherwise }
    \end{cases}.
\end{equation*}
Continuing along the lines of reasoning laid out in the proof of Proposition~\ref{prop:sampthm}, we then acquire
\begin{equation} \label{EshL2bdabs}
    \|\mathcal{E}_{M,r,h}f\|_2\leq \bigg(\max_{w\in [-\sigma,\sigma]} |\nu_{M,r}(hw)| + \max_{w\in [-\sigma,\sigma]} \sum_{k\in\mathbb{Z}\setminus\{0\}} |\mu_{M,r}(hw+2k\pi)|\bigg)\|f\|_2,
\end{equation} 
and
\begin{equation} \label{EshLinftybdabs}
    \|\mathcal{E}_{M,r,h}f\|_{\infty} \leq \sqrt{\sigma/\pi} \bigg(\max_{w\in [-\sigma,\sigma]} |\nu_{M,r}(hw)| + \max_{w\in [-\sigma,\sigma]} \sum_{k\in\mathbb{Z}\setminus\{0\}} |\mu_{M,r}(hw+2k\pi)|\bigg) \|f\|_2.
\end{equation}
Upon a close examination of the said proof, particularly \eqref{lastexp}, it is evident that
\begin{equation} \label{maxnuscaled}
    \max_{w\in [-\sigma,\sigma]} |\nu_{M,r}(hw)|\lesssim \frac{\exp(-r^2(\pi-h\sigma)^2/2)}{r(\pi-h\sigma)} + \frac{r^2}{M}\exp\bigg(-\frac{M^2}{2r^2}\bigg)
\end{equation}
and that
\begin{equation} \label{maxmuscaled}
    \max_{w\in [-\sigma,\sigma]} \sum_{k\in\mathbb{Z}\setminus\{0\}} |\mu_{M,r}(hw+2k\pi)| \lesssim \frac{\exp(-r^2(\pi-h\sigma)^2/2)}{r(\pi-h\sigma)} + c(r)\exp\bigg(-\frac{M^2}{4r^2}\bigg).
\end{equation}
Hence, by combining \eqref{EshL2bdabs}, \eqref{EshLinftybdabs}, \eqref{maxnuscaled}, \eqref{maxmuscaled}, we arrive at
\begin{equation} \label{hError}
    \|\mathcal{E}_{M,r,h}f\|_2 \lesssim \epsilon(\sigma,M,r,h)\|f\|_2 \quad\text{ and }\quad
    \|\mathcal{E}_{M,r,h}f\|_\infty \lesssim \sqrt{\sigma}\,\epsilon(\sigma,M,r,h)\|f\|_2,
\end{equation}
where
\begin{equation} \label{def:hbigE} 
    \epsilon(\sigma,M,r,h) := \frac{\exp(-r^2(\pi-h\sigma)^2/2)}{r(\pi-h\sigma)} + \frac{r^2}{M}\exp\bigg(-\frac{M^2}{2r^2}\bigg) + c(r)\exp\bigg(-\frac{M^2}{4r^2}\bigg).
\end{equation}  
To finish the proof, we focus on selecting the parameters $M$, $h$, $r$. Drawing insight from \eqref{def:hbigE} and the requirement that $\sigma<\pi/h$, we want to choose $h$ small enough so that $|\pi-h\sigma|\geq c$ for some $c>0$ of choice, and simultaneously, choose $M,r$ so that $M\approx r^2$. For simplicity, we can set $r=\sqrt{M}$ and $h=4/M$, with $M\in\mathbb{Z}_+$ large enough so that $\pi/2> 4\sigma/M$. With these parametric choices established, we conclude from \eqref{hError} that
\begin{equation*}
    \|\mathcal{E}_{M,r,h}f\|_2 \lesssim \exp(-M/4)\|f\|_2 \quad\text{ and }\quad \|\mathcal{E}_{M,r,h}f\|_\infty \lesssim \sqrt{\sigma}\exp(-M/4)\|f\|_2,
\end{equation*}
as wanted.
\end{proof}

\subsection{Proof of Theorem~\ref{thm:samp} in high dimensions}

Let it be noted that we are following the notation convention elucidated in Remark~\ref{rem:notconv} throughout this subsection.

Let $d\in\mathbb{Z}_+$ and $f\in B^2_{[-\sigma,\sigma]^d}$, where $\sigma<\pi/(2h)=M\pi/8$. We have the following result that is a direct consequence of Theorem~\ref{parzensampthm}:
\begin{equation} \label{parzenrecall}
    f(\vec{x}) =\sum_{\vec{n}\in\mathbb{Z}^{d}}f(h\vec{n})\sinc{}_h(\vec{x} -h\vec{n})=\sum_{\vec{n}\in\mathbb{Z}^{d}}f(h\vec{n})\prod_{j=1}^{d}\sinc(h^{-1}x_{j}-n_{j}), \quad\forall\vec{x}\in\mathbb{R}^d.
\end{equation}
Next, recall from \eqref{Op3} that
\begin{equation} \label{Op3high}
    \begin{split}
        \mathcal{R}_{M,r,h}f(\vec{x}) &=\sum_{\vec{n}\in\mathbb{Z}^{d}}f(h\vec{n})\sinc(h^{-1}\vec{x}-\vec{n})\mathcal{G}_{M,r}(h^{-1}\vec{x}-\vec{n})\\
        &= \sum_{\vec{n}\in\mathbb{Z}^{d}}f(h\vec{n})\sinc{}_{h}\cdot\mathcal{G}_{M,r,h}(\vec{x}-h\vec{n}),
    \end{split}   
\end{equation}
which, similar to \eqref{def:errors}, has a corresponding error mapping that is $\mathcal{E}_{M,r,h} = \mathcal{R}_{M,r,h}-Id$. By performing the Fourier transform on both \eqref{parzenrecall}, \eqref{Op3high}, we can derive, for $\vec{w}\in\mathbb{R}^d$
\begin{equation} \label{FThigh}
    \begin{split}
        \hat{f}(\vec{w}) &= h \sum_{\vec{n}\in\mathbb{Z}^d} f(h\vec{n})e^{ih\vec{w}\cdot\vec{n}}\chi_{[-\pi,\pi]^d}(h\vec{w})\\
        \widehat{\mathcal{R}_{M,r,h}f}(\vec{w}) &= h\sum_{\vec{n}\in\mathbb{Z}^d} f(h\vec{n})\mu_{M,r,h}(\vec{w}),
    \end{split}
\end{equation}
respectively, where 
\begin{equation} \label{def:muhigh}
    \mu_{M,r,h}(\vec{w}) :=\frac{1}{(2\pi)^{d}}\int_{\vec{t}\in h\vec{w}+[-\pi,\pi]^{d}}\widehat{\mathcal{G}_{M,r}}(\vec{t})\,d\vec{t}=\frac{\mathcal{F}(\sinc\cdot\mathcal{G}_{M,r})(h\vec{w})}{(2\pi)^d}.
\end{equation}
Hence, if we define
\begin{equation} \label{def:nuhigh}
    \nu_{M,r,h}(\vec{w}) :=\frac{1}{(2\pi)^{d}}\int_{\vec{t}\not\in h\vec{w}+[-\pi,\pi]^{d}}\widehat{\mathcal{G}_{M,r}}(\vec{t})\,d\vec{t},
\end{equation}
then by utilizing \eqref{FThigh} and drawing upon a similar line of reasoning as explained in \S\ref{sec:sampthmpfdim1}, we obtain
\begin{equation*}
    \widehat{\mathcal{E}_{M,r,h}f}(\vec{w}) = 
    \begin{cases}
        \hat{f}(\vec{w})\nu_{M,r,h}(\vec{w}) &\text{ if } \vec{w}\in [-\sigma,\sigma]^d\\
        -\hat{f}(\vec{w}-2\vec{k}\pi/h)\mu_{M,r,h}(\vec{w}) &\text{ if } \vec{w}-2\vec{k}\pi/h \in [-\sigma, \sigma]^d, \text{ for some } \vec{k}\in\mathbb{Z}^d\setminus\{\vec{0}\}\\
        0 &\text{ otherwise }
    \end{cases}.
\end{equation*}
It now readily follows that
\begin{equation} \label{EMrhhigh}
    \|\mathcal{E}_{M,r,h}f\|_2 \leq C(M,r,h)\|f\|_2, \quad\text{ and }\quad \|\mathcal{E}_{M,r,h}f\|_{\infty} \leq (\sigma/\pi)^{d/2} C(M,r,h) \|f\|_2,
\end{equation}
where
\begin{equation*}
    C(M,r,h) := \bigg(\max_{\vec{w}\in [-\sigma,\sigma]^d} |\nu_{M,r,h}(\vec{w})| + \max_{\vec{w}\in [-\sigma,\sigma]^d} \sum_{\vec{k}\in\mathbb{Z}^d\setminus\{\vec{0}\}} |\mu_{M,r,h}(\vec{w}+2\vec{k}\pi/h)|\bigg).
\end{equation*}
Note the parallel between \eqref{EMrhhigh} and \eqref{EshL2bdabs}, \eqref{EshLinftybdabs}. To manage \eqref{EMrhhigh}, we utilize the Gaussian nature of $\widehat{\mathcal{G}_{M,r}}$ to split the integrals in \eqref{def:muhigh}, \eqref{def:nuhigh} along the one-dimensional coordinates and acquire, respectively,
\begin{equation*} 
    \mu_{M,r,h}(\vec{w}) = \prod_{j=1}^d \bigg(\frac{1}{2\pi}\int_{t_j\in hw_j\in [-\pi,\pi]} \widehat{\mathcal{G}_{M,r}}(t_j)\,dt_j\bigg) = \prod_{j=1}^d \mu_{M,r}(hw_j),
\end{equation*}
and
\begin{equation*}
    |\nu_{M,r,h}(\vec{w})| \leq \sum_{j=1}^d \bigg|\frac{1}{2\pi}\int_{t_j\not\in hw_j\in [-\pi,\pi]} \widehat{\mathcal{G}_{M,r}}(t_j)\,dt_j\bigg|\prod_{k\not=j}\bigg|\frac{1}{2\pi}\int_\mathbb{R} \widehat{\mathcal{G}_{M,r}}(t_k)\,dt_k\bigg| = \sum_{j=1}^d |\nu_{M,r}(hw_j)|.
\end{equation*}
These two observations allow us to put the one-dimensional theory from \S\ref{sec:sampthmpfdim1} into action. In particular, recalling that $r=\sqrt{M}$, we can derive
\begin{equation} \label{nuboundhigh}
    \max_{\vec{w}\in [-\sigma,\sigma]^d}|\nu_{M,r,h}(\vec{w})| \lesssim_d \exp(-M/4).
\end{equation}
On the other hand, one sees that
\begin{align*} 
    &\max_{\vec{w}\in [-\sigma,\sigma]^d} \sum_{\vec{k}\in\mathbb{Z}^d\setminus\{\vec{0}\}} |\mu_{M,r,h}(\vec{w}+2\vec{k}\pi/h)| \\
    &\leq \sum_{j=1}^d \bigg(\max_{w_j\in [-\sigma,\sigma]} \sum_{k_j\in\mathbb{Z}\setminus\{0\}} |\mu_{M,r}(hw_j+ 2k_j\pi)|\bigg)\prod_{l\not=j} \bigg(\max_{w_l\in [-\sigma,\sigma]} \sum_{k_l\in\mathbb{Z}} |\mu_{M,r}(hw_l+ 2k_l\pi)|\bigg),
\end{align*}
where we can take note from the calculations in \eqref{strings} and the Mills' ratio \eqref{Mratio} to obtain 
\begin{equation*}
    \begin{split}
        \max_{w_l\in [-\sigma,\sigma]} \sum_{k_l\in\mathbb{Z}} |\mu_{M,r}(hw_l+ 2k_l\pi)| &\leq \max_{w_l\in [-\sigma,\sigma]} |\mu_{M,r}(hw_l)| + \max_{w_l\in [-\sigma,\sigma]} \sum_{k\in\mathbb{Z}\setminus\{0\}} |\mu_{M,r}(hw_l+ 2k_l\pi)|\\
        &\lesssim 1 + \int_M^{\infty} \mathcal{G}_r(x)\,dx + \exp(-M/4) \lesssim 1;
    \end{split}
\end{equation*}
therefore, we conclude
\begin{equation} \label{muboundhigh}
    \max_{\vec{w}\in [-\sigma,\sigma]^d} \sum_{\vec{k}\in\mathbb{Z}^d\setminus\{\vec{0}\}} |\mu_{M,r,h}(\vec{w}+2\vec{k}\pi/h)|\lesssim_d \exp(-M/4).
\end{equation}
Putting \eqref{nuboundhigh}, \eqref{muboundhigh} back in \eqref{EMrhhigh}, we finally arrive at 
\begin{equation*}
    \|\mathcal{E}_{M,r,h}f\|_2 \lesssim_d \exp(-M/4)\|f\|_2 \quad\text{ and }\quad \|\mathcal{E}_{M,r,h}f\|_\infty \lesssim_d \sigma^{d/2}\exp(-M/4)\|f\|_2,
\end{equation*}
culminating in the conclusion of Theorem~\ref{thm:samp}. \qed

\section{Proof of Theorem~\ref{thm:proxyGNN}} \label{sec:proxyGNNthm}

In this section and the next, we employ the following notations for the sake of compact presentation:
\begin{equation} \label{compactpresent}
    \mathfrak{s}(x) := \sinc(x) \quad\text{ and }\quad \rho(x) :=  \mathrm{ReLU}(x). 
\end{equation}
For instance, we denote the shifted ReLU at $X\in\mathbb{R}$ as $\rho_X$ and the scaled sinc function as $\mathfrak{s}_h$.

Let $f\in B^2_\sigma$. We have established in Proposition~\ref{prop:sampthm} that if 
\begin{equation} \label{Rrecall}
    \mathcal{R}_{M,r,h}f(x) = \sum_{n\in\mathbb{Z}}f(hn)\,\mathfrak{s}_h\cdot\mathcal{G}_{M,r,h}(x-hn), 
\end{equation}
where $h=4/M$, $r=\sqrt{M}$, and $M\in\mathbb{Z}_+$ is such that $\sigma< \pi/(2h)= M\pi/8$, then the followings hold 
\begin{align*}
    \|f - \mathcal{R}_{M,r,h}f\|_2 &\lesssim \exp(-M/4)\|f\|_2\\
    \|f - \mathcal{R}_{M,r,h}f\|_\infty &\lesssim \sqrt{\sigma}\exp(-M/4)\|f\|_2.
\end{align*}
Although the count of nonzero terms in \eqref{Rrecall} is contingent on the position of $x$, the series remains infinite a priori. Hence, to facilitate a practical integration of the sampling reconstruction formula into a neural network, we substitute $\mathcal{R}_{M,r,h}f$ with the following function, supposing $M\in 2\mathbb{Z}_+$,
\begin{equation} \label{tildeR}
    \widetilde{\mathcal{R}}_{M,r,h}f(x) := \sum_{n=-M/2}^{M/2} f(hn)\,\mathfrak{s}_h\cdot\mathcal{G}_{M,r,h}(x-hn).
\end{equation}
It is apparent from \eqref{outputg2} that we expect $\widetilde{\mathcal{R}}_{M,r,h}f$ to be nearly the output $g_2$ of the proxy GNN model $\Psi_f$ introduced in \S\ref{sec:mockproof}. What remains for us in this section is to show that the errors in both \eqref{outputg2}, \eqref{nearlyblm} indeed exhibit exponential decay concerning the parameter $M$. 

To confirm \eqref{nearlyblm}, we start by establishing the comparability between $\widetilde{\mathcal{R}}_{M,r,h}f$ and $\mathcal{R}_{M,r,h}f$ over $[-1,1]$. As explained in \S\ref{sec:Results}, we also have the flexibility to assume prior knowledge of the complete profile of $f$ over the said interval, thereby turning the generalization problem into an approximation one. We examine the following difference:
\begin{equation} \label{thediff}
    \begin{split}
        \mathfrak{F}_{M,r,h}(x) &:= \widetilde{\mathcal{R}}_{M,r,h}f(x) - \mathcal{R}_{M,r,h}f(x) \\
        &= \sum_{n=-M/2}^{M/2}f(hn)\,\mathfrak{s}_h\cdot\mathcal{G}_{M,r,h}(x-hn)-\sum_{n\in\mathbb{Z}} f(hn)\,\mathfrak{s}_h\cdot\mathcal{G}_{M,r,h}(x-hn). 
    \end{split}
\end{equation}
Observe that when $x\in [-1,1]$, the integer index $n$ in the second sum of \eqref{thediff} is restricted to a range of at most from $-\lceil 5M/4\rceil$ to $\lceil 5M/4\rceil$, whereas in the first sum, it spans the entire set $\{-M/2,\cdots,M/2\}$. Therefore, the difference function $\mathfrak{F}_{M,r,h}$ over $[-1,1]$ assumes the following form
\begin{equation} \label{thediffsimp}
    \mathfrak{F}_{M,r,h}(x)=\sum_{n=-M/2}^{-\lceil 5M/4\rceil } + \sum_{n=M/2}^{\lceil 5M/4\rceil}\tilde{\delta}_{n,x}f(hn)\,\mathfrak{s}_h\cdot\mathcal{G}_{M,r,h}(x-hn),
\end{equation}
where $\tilde{\delta}_{n,x}\in\{0,-1\}$ denotes a quantity whose values change with $n,x$. For example, suppose $M\in 4\mathbb{Z}_+$ and $x=1$, then $\tilde{\delta}_{n,1}=-1$ when 
\begin{equation*}
    n=-3M/4,\cdots,-M/2-1 \quad\text{ or }\quad n=M/2+1,\cdots,5M/4,
\end{equation*}
and $\tilde{\delta}_{n,1}=0$ when 
\begin{equation*}
    n= -M/2,\cdots, M/2 \quad\text{ or }\quad n=-5M/4,\cdots,-3M/4-1.
\end{equation*}
Hence, $\mathfrak{F}_{M,r,h}$ is a piecewise continuous function on $[-1,1]$. Using \eqref{thediffsimp}, we can establish a relatively small upper bound for its absolute value within this interval, as illustrated in the following lemma. 

\begin{lemma} \label{lem:diff}
Let $f\in B^2_\sigma$. Suppose $h=4/M$, $r=\sqrt{M}$ where $M\in 4\mathbb{Z}_+$ is sufficiently large so that $\sigma<\pi/h = M\pi/4$. Then the following holds for $x\in [-1,1]$
\begin{equation*} 
    |\mathfrak{F}_{M,r,h}(x)|\lesssim \sqrt{M}\exp(-M/32)\|f\|_2.
\end{equation*}
\end{lemma}

\begin{proof}
Let $x\in [-1,1]$. From \eqref{thediffsimp}, we deduce that
\begin{equation*}
    \begin{split}
        |\mathfrak{F}_{M,r,h}(x)| &\leq \exp(-(M/4)^2/2r^2)\bigg(\sum_{n=-M/2}^{-5M/4} + \sum_{n=M/2}^{5M/4} |f(hn)|\bigg)\\
        &\lesssim \sqrt{M}\exp(-M/32)\bigg(\sum_{n\in\mathbb{Z}} |f(hn)|^2\bigg)^{1/2}\\
        &= \sqrt{M}\exp(-M/32)\|f\|_2,
    \end{split}
\end{equation*}
where we have applied the Gaussian quadrature formula \eqref{Gaussquad} to the final step above. 
\end{proof}

We have verified, via the lemma, that by confining our consideration to the interval $[-1,1]$, with a sufficiently large $M$, $\widetilde{\mathcal{R}}_{M,r,h}f$ serves as a desirable replacement for $\mathcal{R}_{M,r,h}f$. Subsequently, we apply Theorem~\ref{thm:samp} to acquire \eqref{nearlyblm}. With this in place, we now attend to \eqref{outputg2}. In what follows, recall that we have introduced the $O$-notation in \S\ref{sec:Preliminaries}.

\begin{proposition} \label{prop:intbyparts} Let $f\in B^2_\sigma$. Suppose $h=4/M$, $r=\sqrt{M}$ where $M\in 2\mathbb{Z}_+$ is sufficiently large so that $\sigma <\pi/h = M\pi/4$. Then for every $x\in [-1,1]$
\begin{equation} \label{prop2conc}
    \begin{split}
        &\widetilde{\mathcal{R}}_{M,r,h}f(x) \\
        &= \sum_{n=-M/2}^{M/2}f(hn)\,\mathfrak{s}_h\cdot\mathcal{G}_{M,r,h}(x-hn) \\
        &= \bigg(\sum_{n=-M/2}^{M/2}f(hn)\rho_{hn}\bigg)\ast (\mathfrak{s}_h\cdot\mathcal{G}_{M,r,h})''(x) + O(M^{3/2}\exp(-M/2)\|f\|_2).
    \end{split}
\end{equation}
\end{proposition}

\begin{proof} We rely on the following lemma, whose proof is provided in \S\ref{sec:intbypartslemma}. 

\begin{lemma} \label{lem:intbyparts}
Let $X\in\mathbb{R}$. Let $\phi\in C^2(\mathbb{R})$ such that 
\begin{equation} \label{lemm5cond} 
    \phi(x)\to 0, \quad |\phi'(x)|=O((1+|x|)^{-2}), \quad \text{ and }\quad |\phi''(x)|=O((1+|x|)^{-3}) \quad\text{ as }\quad |x|\to\infty.
\end{equation}
Then for every $x\in\mathbb{R}$,
\begin{equation*} 
    \delta_{X}\ast\phi(x) =\mathrm{ReLU}_{X}\ast\phi''(x).
\end{equation*}
\end{lemma}

To see how Proposition~\ref{prop:intbyparts} is derived from Lemma~\ref{lem:intbyparts}, note that, if $x\in [-1,1]$ and $n\in\{-M/2,\cdots,M/2\}$, then we have $h^{-1}x - n = (M/4)x-n\in (-M,M)$. Hence 
\begin{equation} \label{GMrhandGrh}
    \mathcal{G}_{M,r,h}(x-hn)=\mathcal{G}_{M,r}(h^{-1}x-n)=\mathcal{G}_r(h^{-1}x-n)=\mathcal{G}_{r,h}(x-hn),
\end{equation}
and consequently
\begin{align}
    \nonumber \widetilde{\mathcal{R}}_{M,r,h}f(x) &= \sum_{n=-M/2}^{M/2}f(hn)\,\mathfrak{s}_h\cdot\mathcal{G}_{M,r,h}(x-hn) \\
    \nonumber &= \sum_{n=-M/2}^{M/2}f(hn)\bigg(\delta_{hn}\ast (\mathfrak{s}_h\cdot\mathcal{G}_{r,h})(x)\bigg)\\
    \label{firstconvo} &= \sum_{n=-M/2}^{M/2}f(hn)\bigg(\rho_{hn}\ast (\mathfrak{s}_h\cdot\mathcal{G}_{r,h})''(x)\bigg),
\end{align}
where we've applied Lemma~\ref{lem:intbyparts} to the last equality above. We now compare the expression in \eqref{firstconvo} with 
\begin{equation*}
    \bigg(\sum_{n=-M/2}^{M/2}f(hn)\rho_{hn}\bigg)\ast (\mathfrak{s}_h\cdot\mathcal{G}_{M,r,h})''(x)
\end{equation*}
in \eqref{prop2conc}. We expand each term $\rho_{hn}\ast (\mathfrak{s}_h\cdot\mathcal{G}_{r,h})''(x)$ as follows
\begin{align}
    \nonumber & \int_{-\infty}^{x-hn}(x-y-hn)(\mathfrak{s}_h\cdot\mathcal{G}_{r,h})''(y)\,dy \\
    \nonumber = & \int_{-\infty}^{-hM} + \int_{-hM}^{x-hn} (x-y-hn)(\mathfrak{s}_h\cdot\mathcal{G}_{r,h})''(y)\,dy\\
    \nonumber = & \rho_{hn}(x-y)(\mathfrak{s}_h\cdot\mathcal{G}_{r,h})'(y)|_{y=-hM} - \int_{hM}^{\infty}(\mathfrak{s}_h\cdot\mathcal{G}_{r,h})'(y)\,dy + \rho_{hn}\ast (\mathfrak{s}_h\cdot\mathcal{G}_{M,r,h})''(x)\\
    \label{expand1} = & \rho_{hn}(x-y)(\mathfrak{s}_h\cdot\mathcal{G}_{r,h})'(y)|_{y=-hM} + (\mathfrak{s}_h\cdot\mathcal{G}_{r,h})(hM) + \rho_{hn}\ast (\mathfrak{s}_h\cdot\mathcal{G}_{M,r,h})''(x),
\end{align}
where the second equality can be attributed to \eqref{GMrhandGrh} and definition \eqref{def:truncGaussian}. Next, observe that, for $x\in [-1,1]$, 
\begin{equation} \label{expand2}
    \begin{split}
        \bigg|\rho_{hn}(x-y)(\mathfrak{s}_h\cdot\mathcal{G}_{r,h})'(y)|_{y=-hM}\bigg| &=O(M\exp(-M/2))\\
        \bigg|(\mathfrak{s}_h\cdot\mathcal{G}_{r,h})(hM)\bigg| &=O(\exp(-M/2)).
    \end{split}   
\end{equation}
Furthermore, by the Cauchy-Schwarz inequality and \eqref{Gaussquad}, we get
\begin{equation} \label{expand3}
    \bigg|\sum_{n=-N}^{N}f(hn)O(M\exp(-M/2))\bigg| = O(M^{3/2}\exp(-M/2)\|f\|_2).
\end{equation}
Therefore, combining \eqref{expand1}, \eqref{expand2}, \eqref{expand3} leads to
\begin{multline*}
    \bigg(\sum_{n=-M/2}^{M/2}f(hn)\rho_{hn}\bigg)\ast (\mathfrak{s}_h\cdot\mathcal{G}_{r,h})''(x) = \bigg(\sum_{n=-M/2}^{M/2} f(hn)\rho_{hn}\bigg)\ast (\mathfrak{s}_h\cdot\mathcal{G}_{M,r,h})''(x)\\
    + O(M^{3/2}\exp(-M/2)\|f\|_2),
\end{multline*}
which is what we want.
\end{proof}

This proposition concludes our validation for \eqref{outputg2}. It should be now evident from what has been presented thus far, and indicated by both \eqref{outputg2}, \eqref{nearlyblm}, that, with the assumptions put in place in the premise of Theorem~\ref{thm:proxyGNN}, the output of the proxy GNN $\Psi_f$ provides a robust approximation of $f\in B^2_\sigma$ on the interval $[-1,1]$. Indeed, by combining the results of Theorem~\ref{thm:samp}, Lemma~\ref{lem:diff}, Proposition~\ref{prop:intbyparts}, we deduce
\begin{align*}
    \|(f-\Psi_f)\cdot\chi_{[-1,1]}\|_2 &\lesssim \bigg(\exp(-M/4)+ M^{3/2}\exp(-M/2) + \sqrt{M}\exp(-M/32)\bigg)\|f\|_2\\
    \|(f-\Psi_f)\cdot\chi_{[-1,1]}\|_\infty &\lesssim \bigg(\sqrt{\sigma}\exp(-M/4)+ M^{3/2}\exp(-M/2) + \sqrt{M}\exp(-M/32)\bigg)\|f\|_2,
\end{align*}
which, by recalling that $\sigma<\pi/(2h)=M\pi/8$, give the final conclusion of Theorem~\ref{thm:proxyGNN}. \qed

\section{Proof of Theorem~\ref{thm:actualGNN}} \label{sec:actualGNNpf}

We continue to apply the compact notations introduced in \eqref{compactpresent}. To recap Theorem~\ref{thm:proxyGNN}, we have developed a proxy GNN model $\Psi_f$ that is capable of generalizing a bandlimited function $f$ over $[-1,1]$, relying solely on sampled values of $f$ at discrete points within the interval. More concretely, we suppose $f\in B^2_\sigma$ and take $h=4/M$, $r=\sqrt{M}$, where $M\in 4\mathbb{Z}_+$ is large enough so that $\sigma<\pi/(2h)= M\pi/8$, to build 
\begin{equation} \label{proxyPsi}
   \Psi_f(x) = \bigg(\sum_{n=-M/2}^{M/2}f(hn)\rho_{hn}\bigg)\ast(\mathfrak{s}_{h} \cdot\mathcal{G}_{M,r,h})''(x), \quad\forall x\in\mathbb{R}.
\end{equation}
In line with our strategy outlined in \S\ref{sec:mockproof} to establish Theorem~\ref{thm:actualGNN}, we emphasize the following key steps. In dimension one, we perform a discretization of the convolution in \eqref{proxyPsi}. When confined to $x\in [-1,1]$, the integration range is a finite range, 
allowing us to convert the continuous convolution into a discrete graph filtering process. Importantly, it will follow that the graph structure we employ in dimension one corresponds to a dense lattice. 
At the conclusion of the discretization, we obtain a GNN output $\tilde{\Psi}_f$, which, much like $\Psi_f$, can generalize $f$ over a dense lattice of $[-1,1]$. Afterward, we leverage an approximate product to facilitate the extension of the one-dimensional result to higher dimensions. The GNN implementation, \eqref{totalGNNdimone} for dimension one, and \eqref{totalGNNhighdim} for higher dimensions, is then presented towards the conclusion of \S\ref{sec:actualGNNpfhighdims}. Readers who are specifically interested in observing such demonstration can proceed directly to the end of the said subsection. 

We now embark on addressing Theorem~\ref{thm:actualGNN} in dimension one. This corresponds to the step \eqref{GNNindimone} outlined in \S\ref{sec:mockproof}.

\subsection{Proof of Theorem~\ref{thm:actualGNN} in dimension one} \label{sec:actualGNNpfdimone}

Take $M\in 4\mathbb{Z}_+$. For $x\in [-1,1]$, $h=4/M$, $r=\sqrt{M}$, $n=-M/2,\cdots, M/2$, 
\begin{equation} \label{theint}
    \begin{split}
        \rho_{hn}\ast(\mathfrak{s}_h \cdot\mathcal{G}_{M,r,h})''(x)
        &=\int_{\mathbb{R}}\rho_{hn} (y)(\mathfrak{s}_h\cdot\mathcal{G}_{M,r,h})'' (x-y) \,dy \\
        &= \int_{hn}^{x+hM} (y-hn)(\mathfrak{s}_h\cdot\mathcal{G}_{M,r,h})'' (x-y) \,dy.
    \end{split}
\end{equation}
We discretize the integration range $[hn,x+hM]$ into a discrete lattice with some small step size $\tau>0$ to be chosen later. Suppose $x\in [-1,1]\cap\tau\mathbb{Z}$ be a point on this grid, i.e., $x=m\tau$ for some $m\in\mathbb{Z}$, and suppose for such $x$, we can discretize the integral in \eqref{theint} as
\begin{equation} \label{example}
    \rho_{hn} \ast(\mathfrak{s}_h\cdot\mathcal{G}_{M,r,h})''(x) \approx \sum_{l=0}^{L(n,x)} 
    a^{n}_{l}(x) \rho_{hn} (y_{l})(\mathfrak{s}_h\cdot\mathcal{G}_{M,r,h})'' (x-y_{l})
    + \mathcal{J}(n,x),
\end{equation}
for some choices of $a^{n}_{l}(x), y_{l}\in\mathbb{R}$, $L(n,x)\in\mathbb{Z}_{+}$ and a difference term $\mathcal{J}(n,x)$. If $f\in B^2_{\sigma}$, where $\sigma<\pi/(2h)$, then by defining the following function on $[-1,1]\cap\tau\mathbb{Z}$
\begin{equation} \label{1dimPsi}
    \tilde{\Psi}_f(x):= \sum_{n=-M/2}^{M/2}f(hn) \bigg(
    \sum_{l=0}^{L(n,x)}a^{n}_{l}(x) \rho_{hn}(y_{l})(\mathfrak{s}_h\cdot\mathcal{G}_{M,r,h})''(x-y_{l})\bigg),
\end{equation}
we can assert from \eqref{example} that for $x\in [-1,1]\cap\tau\mathbb{Z}$,
\begin{equation} \label{totalerr}
    \begin{split}
        |\Psi_f(x) - \tilde{\Psi}_f(x)| &\lesssim \sqrt{M}\max |\mathcal{J}(n,x)|\bigg(\sum_{n=-M/2}^{M/2}|f(hn)|^2\bigg)^{1/2}\\
        &= \sqrt{M}\max|\mathcal{J}(n,x)|\|f\|_2, 
    \end{split} 
\end{equation}
where the maximum is taken over $x\in [-1,1]\cap\tau\mathbb{Z}$, $n=-M/2,\cdots, M/2$, and the last inequality is a familiar application of \eqref{Gaussquad}. What \eqref{totalerr} says is that, $\tilde{\Psi}_f$ can serve as a suitable approximation for $\Psi_f$ on $[-1,1]\cap\tau\mathbb{Z}$, provided that the heuristic quantity
\begin{equation*}
    \sqrt{M}\max|\mathcal{J}(n,x)|
\end{equation*}
can be made small. Our proof of the theorem in dimension one will then be complete once we demonstrate that $\tilde{\Psi}_f$ can be realized using a GNN model. As mentioned, such network implementation will be given at the conclusion of \S\ref{sec:actualGNNpfhighdims}.

We return to \eqref{example}. To better quantify the approximation, we adopt the closed Newton-Cotes quadrature formula \cite[Subsection~9.3]{quarteroni2010numerical} as our numerical integration tool. Applying this formula to the right-hand-side integral in \eqref{theint}, and using the notations defined in \eqref{example}, we obtain
\begin{equation} \label{newcotes}
    \int_{hn}^{x+hM}(y-hn)(\mathfrak{s}_h\cdot\mathcal{G}_{M,r,h})'' (x-y) \,dy
    = \sum_{l=0}^{L(n,x)} (y_{l}-hn)(\mathfrak{s}_h\cdot\mathcal{G}_{M,r,h})'' (x-y_{l}) a^{n}_{l}(x) +\mathcal{J}(n,x),
\end{equation}
where $y_{l}:= hn+l\tau$, $\tau=(x+hM-hn)/L(n,x)$ for some $L(n,x)\in\mathbb{Z}_{+}$, and 
\begin{equation} \label{anl}
    a^{n}_{l}(x):= 
    \int_{hn}^{x+hM} \prod_{l\not= j=0}^{L(n,x)} \frac{u-y_{j}}{y_{l}-y_{j}}\,du.
\end{equation}
Here, $\big\{\prod_{l\not= j=0}^{L(n,x)}\frac{u-y_{j}}{y_{l}-y_{j}}\big\}_{l=0}^{L(n,x)}$ are the Lagrange polynomials \cite[Subsection~1.6]{insel2003linear} anchoring at $y_{l}$'s. Note that the discretization parameter $\tau$ is selected independently of $n, x$. In fact, it is uniformly chosen to create the grid $\tau\mathbb{Z}$ on the interval $[hn, x+hM]$. 
Therefore, 
\begin{equation} \label{tauandL}
    L(n,x)\tau = x+hM-hn,
\end{equation}
which signifies that $L(n,x)$ is the number of $\tau$-steps required to traverse from $hn$ to $x+hM$. Further, $L(n,x)$ behaves linearly with $n,x$. The largest value of $L(n,x)$ is $L_\mathrm{b}=L(-M/2,1)=7/\tau$, while the smallest is $L_\mathrm{s}=L(M/2,-1)=1/\tau=L_\mathrm{b}/7$. The subscripts $\mathrm{b}, \mathrm{s}$ stand for ``biggest", ``smallest", respectively.

Temporarily fixing $n=-M/2,\cdots,M/2$ and $x\in [-1,1]\cap\tau\mathbb{Z}$, we simplify $L(n,x)$ as $L$ onward. It follows from \eqref{tauandL} that the magnitude of $\tau$ is of $O(L^{-1})$. Now, going back to \eqref{newcotes}, the error $\mathcal{J}(n,x)$ can be expressed as \cite[Theorem~9.2]{quarteroni2010numerical}
\begin{equation} \label{Jtermeven}
    \mathcal{J}(n,x) := \frac{\tau^{L+3}}{(L+2)!}
    \bigg((\xi_\mathrm{e}-hn)(\mathfrak{s}_h\cdot\mathcal{G}_{M,r,h})'' (x-\xi_\mathrm{e}) \bigg)^{(L+2)}
    \int_0^{L}u^2(u-1)(u-2)\cdots (u-L)\,du,
\end{equation}
in case that $L$ is even, and
\begin{equation} \label{Jtermodd}
    \mathcal{J}(n,x) := \frac{\tau^{L+2}}{(L+1)!}
    \bigg( (\xi_\mathrm{o}-hn)(\mathfrak{s}_h\cdot\mathcal{G}_{M,r,h})'' (x-\xi_\mathrm{o}) \bigg)^{(L+1)}
    \int_0^{L}u(u-1)(u-2)\cdots (u-L)\,du,
\end{equation}
in case that $L$ is odd. Both $\xi_\mathrm{e},\xi_\mathrm{o}$ are some points in $(hn,x+hM)$. The subscripts $\mathrm{e}, \mathrm{o}$ stand for ``even" and ``odd", respectively. We proceed to establish an upper bound for $|\mathcal{J}(n,x)|$. Since both \eqref{Jtermeven}, \eqref{Jtermodd} are similar, we will focus our analysis on \eqref{Jtermeven}. In this context, we assume that $L$ is even. We expand the differentiation as
\begin{equation*} 
    \bigg( (\cdot-hn)(\mathfrak{s}_h\cdot\mathcal{G}_{M,r,h})''(x-\cdot) \bigg)^{(L+2)} =\sum_{l=0}^{L+2}{L+2\choose l}\bigg( (\cdot-hn)^{(L+2-l)}\cdot (\mathfrak{s}_h\cdot\mathcal{G}_{M,r,h})^{(l+2)}(x-\cdot) \bigg).
\end{equation*}
Given that the linear term only contributes $O(1)$ in magnitude, we can instead calculate
\begin{equation} \label{derterm}
    \sum_{l=0}^{L+2}{L+2\choose l} |(\mathfrak{s}_h\cdot\mathcal{G}_{M,r,h})^{(l+2)}(x-\cdot)|.
\end{equation}
Observe that $\mathcal{G}_{M,r,h}(x-y) =\mathcal{G}_{M,r}(h^{-1}(x-y)) =\mathcal{G}_{r}(h^{-1}(x-y))$, since $h^{-1}(x-y)\in supp(\mathcal{G}_{M,r})$ when $y\in (hn,x+hM)$. Therefore, we reduce our objective to evaluating 
\begin{equation*}
    \|(\mathfrak{s}_h\cdot\mathcal{G}_{r,h})^{(l+2)}\|_\infty.
\end{equation*}
Due to the Schwartz nature of $\mathcal{G}$, $\mathcal{F}((\mathfrak{s}_h\cdot\mathcal{G}_{r,h})^{(l+2)})$ exists and that $\|\mathcal{F}((\mathfrak{s}_h\cdot\mathcal{G}_{r,h})^{(l+2)})\|_1<\infty$, which enables us to invoke the Fourier inverse theorem \eqref{foudef2} and obtain, for $x\in\mathbb{R}$,
\begin{equation*}
    (\mathfrak{s}_h\cdot\mathcal{G}_{r,h})^{(l+2)}(x) = \frac{1}{2\pi}\int_{\mathbb{R}}\mathcal{F}((\mathfrak{s}_h\cdot\mathcal{G}_{r,h})^{(l+2)})(w)e^{iwx}\,dw;
\end{equation*}
hence
\begin{equation} \label{Ftranstep1}
    \|(\mathfrak{s}_h\cdot\mathcal{G}_{r,h})^{(l+2)}\|_{\infty} \lesssim \|\mathcal{F}((\mathfrak{s}_h\cdot\mathcal{G}_{r,h})^{(l+2)})\|_1.
\end{equation}
The Fourier transform of $(\mathfrak{s}_h\cdot\mathcal{G}_{r,h})^{(l+2)}$ is
\begin{equation*}
    \mathcal{F}((\mathfrak{s}_h\cdot\mathcal{G}_{r,h})^{(l+2)})(w) = h(iw)^{l+2}\mathcal{F}(\sinc{}\cdot\mathcal{G}_{r})(hw) = h(iw)^{l+2}\int_{hw+[-\pi,\pi]} \hat{\mathcal{G}}_{r}(u)\,du,
\end{equation*}
which implies
\begin{equation} \label{Ftranstep2}
    |\mathcal{F}((\mathfrak{s}_h\cdot\mathcal{G}_{r,h})^{(l+2)})(w)| \lesssim h|w|^{l+2}\int_{hw+[-\pi,\pi]} \hat{\mathcal{G}}_{r}(u)\,du=: h P_{l}(w). 
\end{equation}
Further,
\begin{align}
    \nonumber \| P_{l}\|_1 &= \int_{\mathbb{R}} |w|^{l+2}\bigg|\int_{hw+[-\pi,\pi]} \hat{\mathcal{G}}_{r}(u)\,du\bigg|\,dw=\int_{\mathbb{R}}\int_{h^{-1}(u+[-\pi,\pi])}|w|^{l+2}\hat{\mathcal{G}}_{r}(u)\,dwdu\\
    \nonumber &\lesssim \sqrt{M}\bigg(\int_{-\infty}^0\int_{h^{-1}(u+[-\pi,\pi])} + \int_0^{\infty} \int_{h^{-1}(u+[-\pi,\pi])}|w|^{l+2}\exp(-u^2M/2)\, dwdu\bigg)\\
    \nonumber &\lesssim \sqrt{M}h^{-l-3}\bigg(\int_{-\infty}^0 (\pi-u)^{l+2}\exp(-u^2M/2)\,du + \int_0^{\infty} (u+\pi)^{l+2}\exp(-u^2M/2)\,du\bigg)\\
    \label{l1} &\lesssim \sqrt{M}h^{-l-3} \int_{\pi}^{\infty} u^{l+2}\exp(-(u-\pi)^2M/2)\,du.
\end{align}
Recall from \eqref{derterm} that $l=0,\cdots, L+2$. We want to select $U=U(L,M)\in\mathbb{Z}_{+}$, smallest possible, such that, 
\begin{equation} \label{Uchoice}
    \forall u\geq U,\quad \log(u)\leq\frac{(u-\pi)^2M}{8L} \leq\frac{(u-\pi)^2M}{4(L+4)}\leq\frac{(u-\pi)^2M}{4(l+2)}.
\end{equation}
Assuming that we can define $L=L(M)$ strategically so that $L\asymp M$, then $U$ can be rendered finite. For example, we can select $L_\mathrm{s}\in\mathbb{Z}_{+}$ so that 
\begin{equation} \label{conditiononLs}
    M/(4L_\mathrm{s})=0.05,
\end{equation}
which gives
\begin{equation} \label{bigchoice}
    \log(u)\leq 0.025(u-\pi)^2=\frac{(u-\pi)^2M}{8L_\mathrm{s}}\quad\text{as soon as }\quad u\geq U_\mathrm{s}:=U(L_\mathrm{s},M):=14.
\end{equation}
Since $L_\mathrm{b}=7L_\mathrm{s}$, substituting $L=L_\mathrm{b}$ back in \eqref{Uchoice} produces
\begin{equation} \label{smallchoice}
    \log(u)\leq\frac{0.025(u-\pi)^2}{7}=\frac{(u-\pi)^2M}{8L_\mathrm{b}}\quad\text{as soon as }\quad u\geq U_\mathrm{b}:=U(L_\mathrm{b},M):=35.
\end{equation}
Therefore, we have ensured that with this choice of $L_\mathrm{s}$, $U=U(L,M)$ is finite for all corresponding $L_\mathrm{s}\leq L\leq L_\mathrm{b}$. Using such $U$ chosen in \eqref{Uchoice}, we rewrite the upper bound in \eqref{l1} as,
\begin{equation} \label{Ftranstep3}
    \begin{split}
        h\|P_{l}\|_1 &\lesssim \sqrt{M}h^{-l-2}\bigg(\int_{\pi}^{U} u^{l+2}\,du+\int_{U}^{\infty}\exp(-(u-\pi)^2M/4)\,du\bigg)\\
        &\lesssim \sqrt{M}h^{-l-2}\bigg(U^{l+3} +\sqrt{\frac{2\pi}{M}}\bigg).
    \end{split} 
\end{equation}
Combining \eqref{Ftranstep1}, \eqref{Ftranstep2}, \eqref{Ftranstep3} helps us conclude
\begin{equation} \label{linfbd}
    \|(\mathfrak{s}_h\cdot\mathcal{G}_{r,h})^{(l+2)}\|_{\infty}  \lesssim \sqrt{M}(M/4)^{l+2}\Big(U^{l+3}+M^{-1/2}\Big).
\end{equation}
We will integrate this information in a later step. To finalize an upper bound for $|\mathcal{J}(n,x)|$, we turn to control the integral presented in \eqref{Jtermeven}. We claim the following, whose proof is given in \ref{5.5}.

\begin{lemma} \label{lem3}
For every $L\in\mathbb{Z}_{+}$,
\begin{align*}
    \text{if } L \text{ is even}, \quad &\bigg|\int_0^{L}u^2(u-1)(u-2)\cdots (u-L)\,du\bigg|\lesssim (L+1)!\\
    \text{if } L \text{ is odd}, \quad &\bigg|\int_0^{L}u(u-1)(u-2)\cdots (u-L)\,du\bigg|\lesssim L!.
\end{align*}    
\end{lemma}

With this lemma and \eqref{linfbd} at our disposal, we can bound $|\mathcal{J}(n,x)|$ by a constant multiple of
\begin{equation} \label{Jmajor}
    \sqrt{M}\bigg(\frac{M}{4}\bigg)^2\frac{\tau^{L+3}}{L+2}\sum_{l=0}^{L+2}{L+2\choose l}\bigg(\frac{M}{4}\bigg)^{l+2}\Big(U^{l+3}+M^{-1/2}\Big)     \lesssim U^3\sqrt{M}\bigg(\frac{M}{4L}\bigg)^2 \bigg(\frac{UM}{4L}+\frac{1}{L}\bigg)^{L+2},
\end{equation}
where, we have recalled that $|\tau|=O(L^{-1})$. It is evident from \eqref{bigchoice}, \eqref{smallchoice} that the right-hand-side of \eqref{Jmajor} largest bound occurs when $L=L_\mathrm{s}$ is the smallest and $U=U_\mathrm{s}$. Taking this into account along with the heuristic evaluation in \eqref{totalerr}, we arrive at
\begin{equation} \label{finest}
    |\Psi_f(x) -\tilde{\Psi}_f(x)|\lesssim  U_\mathrm{s}^3M\bigg(\frac{M}{4L_\mathrm{s}}\bigg)^2\bigg(\frac{U_\mathrm{s}M}{4L_\mathrm{s}}+\frac{1}{L_\mathrm{s}}\bigg)^{L_\mathrm{s}+2}\|f\|_2.
\end{equation}
Since $U_\mathrm{s}=14$, the majorant in right-hand-side of \eqref{finest} can be replaced by a constant multiple of
\begin{equation} \label{discrep}
    M\bigg(\frac{14M}{4L_\mathrm{s}}\bigg)^{L_\mathrm{s}+4}\leq M (0.7)^{5M+4}\lesssim  2^{-M},\quad\forall M\geq 4.
\end{equation}
The step size is then $\tau = 1/L_\mathrm{s}=1/(5M)$. We remark that had we chosen a smaller ratio for $M/(4L_s)$ in \eqref{conditiononLs}, it would have led to a smaller $\tau$ and a reduced power in \eqref{discrep}, and consequently, a diminished discrepancy $|\Psi_f(x) -\tilde{\Psi}_f(x)|$ in \eqref{finest}. However, for the sake of concreteness, we opt to retain the value $\tau=1/(5M)$ in our proof.

We have assumed $L=L(n,x)$ to be even. However, in the case that it is odd, the analysis remains nearly identical. If we had used \eqref{Jtermodd} and thus analyzed
\begin{equation*} 
    \sum_{l=0}^{L+1}{L+1\choose l} |(\mathfrak{s}_h\cdot\mathcal{G}_{M,r,h})^{(l+2)}(x-\cdot)|
\end{equation*}
instead, \eqref{Uchoice} would still be applicable, and we would have reached
\begin{equation*} 
    M\bigg(\frac{10M}{4L_\mathrm{s}}\bigg)^{L_\mathrm{s}+3}\leq M (0.7)^{5M+3}\lesssim  2^{-M},\quad\forall M\geq 4.
\end{equation*}
Therefore, by Theorem~\ref{thm:proxyGNN}, we can conclude that, for $x\in [-1,1]\cap\tau\mathbb{Z}$
\begin{equation} \label{dimoneapprox}
    |\tilde{\Psi}_f(x)-f(x)| \leq |\tilde{\Psi}_f(x)-\Psi_f(x)|+|\Psi_f(x)-f(x)| \lesssim M^{3/2}\exp(-M/32)\|f\|_2,
\end{equation}
from which \eqref{GNNthmconc} follows. \qed

\subsection{Proof of Theorem~\ref{thm:actualGNN} in high dimensions} \label{sec:actualGNNpfhighdims}

We emphasize that we are adhering to the notation convention in Remark~\ref{rem:notconv} throughout this subsection.

Let $d\in\mathbb{Z}_+$ and let $f\in B^2_{[-\sigma,\sigma]^d}$, where $\sigma<\pi/(2h) = M\pi/8$, for some $M\in 4\mathbb{Z}_+$ large and $r=\sqrt{M}$. We define the following high-dimensional extension of \eqref{1dimPsi}
\begin{equation} \label{def:Psi*}
    \begin{split}
        \Psi^*_f(\vec{x}) &:= \sum_{\vec{n}\in [-M/2,M/2]^{d}}f(h\vec{n})\prod_{j=1}^{d}{\bf F}_{n_j}(x_j)\\
        &:= \sum_{\vec{n}\in [-M/2,M/2]^{d}}f(h\vec{n})\prod_{j=1}^{d}\bigg( \sum_{l_j=0}^{L(n_j,x_j)}a^{n_j}_{l_j}(x_j)\rho_{hn_j} (y_{l_j})(\mathfrak{s}_h\cdot\mathcal{G}_{M,r,h})'' (x_j-y_{l_j}) \bigg);
    \end{split}
\end{equation}
see also \eqref{newcotes}. We wish to claim that $\Psi^*_f$ poses a dependable replication of $f$ over $[-1,1]^d\cap\tau\mathbb{Z}^d$, $\tau=1/(5M)$. Recall that, as previously mentioned in the conclusion of \S\ref{sec:actualGNNpfdimone}, a smaller value for $\tau$ would also suffice. However, the complexity introduced by high dimensions necessitates the utilization of a neural network to replicate the product in \eqref{def:Psi*}. Therefore, $\Psi^*_f$ does not precisely align with the eventual output $\tilde{\Psi}_f$ of our ReLU GNN. Hence, our approach to proving Theorem~\ref{thm:actualGNN} will involve performing the following three tasks in succession:
\begin{enumerate}
    \item We prove $\Psi^*_f\approx f$ on $[-1,1]^d$,
    \item We define a ReLU GNN that outputs $\tilde{\Psi}_f$ on $[-1,1]^d\cap\tau\mathbb{Z}^d$,
    \item We show that $\tilde{\Psi}_f\approx \Psi^*_f$ on $[-1,1]^d\cap\tau\mathbb{Z}^d$.
\end{enumerate}
To achieve the first task, we will proceed along a sequence of approximations:
\begin{alignat}{2}
    \label{approx1} \Psi^*_f(\vec{x}) &\approx \Psi_{f}(\vec{x}) &&:= \sum_{\vec{n}\in [-M/2,M/2]^{d}}f(h\vec{n})\prod_{j=1}^{d} \rho_{hn_{j}}\ast(\mathfrak{s}_h\cdot\mathcal{G}_{M,r,h})''(x_{j}), \\
    \label{approx2} \Psi_{f}(\vec{x}) &\approx \widetilde{\mathcal{R}}_{M,r,h}f(\vec{x}) &&:= \sum_{\vec{n}\in [-M/2,M/2]^d} f(h\vec{n})\prod_{j=1}^d\mathfrak{s}_h\cdot\mathcal{G}_{M,r,h}(x_j-hn_j),\\
    \label{approx3} \widetilde{\mathcal{R}}_{M,r,h}f(\vec{x}) &\approx \mathcal{R}_{M,r,h}f(\vec{x}) &&= \sum_{\vec{n}\in\mathbb{Z}^d}f(h\vec{n})\prod_{j=1}^d\sinc{}_{h}\cdot\mathcal{G}_{M,r,h}(x_j-hn_j).
\end{alignat}
Note that we would be well-positioned to complete the task after \eqref{approx1}, \eqref{approx2}, \eqref{approx3} for it has been confirmed by Theorem~\ref{thm:samp} that $\mathcal{R}_{M,r,h}f\approx f$. Proceeding with \eqref{approx3}, we establish its validity through the verification of a high-dimensional counterpart of Lemma~\ref{lem:diff}.

\begin{lemma} \label{lem:diffhigh}
Let $f\in B^2_{[-\sigma,\sigma]^d}$. Suppose $h=4/M$, $r=\sqrt{M}$ where $M\in 4\mathbb{Z}_+$ is sufficiently large so that $\sigma<\pi/h = M\pi/4$. Then the following holds for $\vec{x}\in [-1,1]^d$
\begin{equation*} 
    |\widetilde{\mathcal{R}}_{M,r,h}f(\vec{x})-\mathcal{R}_{M,r,h}f(\vec{x})|\lesssim_d M^{d/2}\exp(-M/32)\|f\|_2.
\end{equation*}
\end{lemma}

\begin{proof}
Define, 
\begin{equation*}
    \mathfrak{F}_{M,r,h}(\vec{x}) := \widetilde{\mathcal{R}}_{M,r,h}f(\vec{x})-\mathcal{R}_{M,r,h}f(\vec{x}).
\end{equation*}
In a manner akin to the discussion outlined in \S\ref{sec:proxyGNNthm}, we derive that, for $\vec{x}\in [-1,1]^d$, 
\begin{equation} \label{thediffsimphigh}
    \mathfrak{F}_{M,r,h}(\vec{x})=\sum_{\vec{n}\in [-5M/4,5M/4]^{d}\setminus [-M/2,M/2]^d} \tilde{\delta}_{\vec{n},\vec{x}}f(h\vec{n})\prod_{j=1}^{d}\mathfrak{s}_h\cdot\mathcal{G}_{M,r,h}(x_{j}-hn_{j}).
\end{equation}
where $\tilde{\delta}_{\vec{n},\vec{x}}\in\{0,-1\}$ is a quantity whose values change with $\vec{x}$, $\vec{n}$. For each $j=1,\cdots,d$, we let 
\begin{equation*}
    \tilde{S}_j := \{\vec{n}=(n_1,\cdots,n_d)\in [-5M/4,5M/4]^d: n_j\in [-5M/4,-M/2]\cup [M/2,5M/4]\}.
\end{equation*}
Then it can be inferred from \eqref{thediffsimphigh} that, for $\vec{x}\in [-1,1]^d$,
\begin{equation} \label{Fbound1}
    |\mathfrak{F}_{M,r,h}(\vec{x})| \leq \sum_{j=1}^d\sum_{\vec{n}\in\tilde{S}_j} |f(h\vec{n})|\prod_{j=1}^{d}|\mathfrak{s}_h\cdot\mathcal{G}_{M,r,h}(x_{j}-hn_{j})|.
\end{equation}
Moreover, as illustrated in the proof of Lemma~\ref{lem:diff}, when both $x_j\in [-1,1]$ and $n_j\in [-5M/4,-M/2]\cup [M/2,5M/4]$, we have the following upper bound
\begin{equation} \label{Fbound2}
    |\mathfrak{s}_h\cdot\mathcal{G}_{M,r,h}(x_{j}-hn_{j})|\leq \exp(-(M/4)^2/2r^2) = \exp(-M/32).
\end{equation}
Therefore, inserting \eqref{Fbound2} back in \eqref{Fbound1} and invoking the Gaussian quadrature \eqref{Gaussquad} deliver us the conclusion
\begin{equation*}
    |\mathfrak{F}_{M,r,h}(\vec{x})| \lesssim_d \exp(-M/32)\bigg(\sum_{\vec{n}\in\tilde{S}_j} |f(h\vec{n})|\bigg)\lesssim_d M^{d/2}\exp(-M/32)\|f\|_2,
\end{equation*}
as desired.
\end{proof}

We now begin with some preparations for \eqref{approx1}, \eqref{approx2}. Following the reasoning presented at the end of \S\ref{sec:actualGNNpfdimone}, and observing \eqref{Jmajor}, \eqref{finest}, \eqref{discrep}, we deduce that, in one dimension and for any $j=1,\cdots,d$,
\begin{equation} \label{bound1}
    |\rho_{hn_j}\ast(\mathfrak{s}_h \cdot\mathcal{G}_{M,r,h})''(x_j) - {\bf F}_{n_j}(x_j)| \lesssim 2^{-M}\ll 1, \quad\forall M\geq 4.
\end{equation}
By a direct calculation, if $y\in (-hM,hM)$, then $|(\mathfrak{s}_h\cdot\mathcal{G}_{M,r,h})''(y)|\lesssim M^2$. Therefore, for all $x_j\in [-1,1]$, $n_j=-M/2,\cdots,M/2$,
\begin{equation*}
    \begin{split}
        |\rho_{hn_j}\ast(\mathfrak{s}_h \cdot\mathcal{G}_{M,r,h})''(x_j)| &= \bigg| \int_{hn}^{x_j+hM} (y-hn_j)(\mathfrak{s}_h\cdot\mathcal{G}_{M,r,h})'' (x_j-y)\,dy\bigg|\\
        &\lesssim \int_{hn}^{x_j+hM} |(\mathfrak{s}_h\cdot\mathcal{G}_{M,r,h})''(y)|\,dy\lesssim M^2
    \end{split}
\end{equation*}
which, along with \eqref{bound1}, renders 
\begin{equation} \label{allbounded}
    |\rho_{hn_j}\ast(\mathfrak{s}_h \cdot\mathcal{G}_{M,r,h})''(x_j)|, \quad |{\bf F}_{n_j}(x_j)| \lesssim  M^2.
\end{equation}
This information will be needed for the following lemma, in which we prove \eqref{approx1}, \eqref{approx2}.

\begin{lemma} \label{lem:approx1&2}
Let $f\in B^2_{[-\sigma,\sigma]^d}$. Suppose $h=4/M$, $r=\sqrt{M}$ where $M\in 4\mathbb{Z}_+$ is sufficiently large so that $\sigma<\pi/h = M\pi/4$. Then the followings hold for $\vec{x}\in [-1,1]^d$
\begin{align} 
    \label{approx1conc} |\Psi^*_f(\vec{x})-\Psi_{f}(\vec{x})|\lesssim_d M^{5d/2}2^{-M}\|f\|_2\\
    \label{approx2conc} |\Psi_{f}(\vec{x})- \widetilde{\mathcal{R}}_{M,r,h}f(\vec{x})|\lesssim_d M^{5d/2}\exp(-M/2)\|f\|_2.
\end{align}
\end{lemma}

\begin{proof}
We handle \eqref{approx2conc} first. It follows from the one-dimensional theory in \S\ref{sec:proxyGNNthm}, particularly in the proof of Proposition~\ref{prop:intbyparts} that, for each $j=1,\cdots,d$,
\begin{equation} \label{diff1}
    |\rho_{hn_{j}}\ast(\mathfrak{s}_h\cdot\mathcal{G}_{M,r,h})''(x_{j}) - \mathfrak{s}_h\cdot\mathcal{G}_{M,r,h}(x_j-hn_j)| \lesssim M\exp(-M/2).
\end{equation}
Hence, applying \eqref{allbounded}, \eqref{diff1} and the Gaussian quadrature \eqref{Gaussquad} simultaneously yields
\begin{multline*}
    \sum_{\vec{n}\in [-M/2,M/2]^{d}}|f(h\vec{n})|\cdot\bigg|\prod_{j=1}^{d} \rho_{hn_{j}}\ast(\mathfrak{s}_h\cdot\mathcal{G}_{M,r,h})''(x_{j})-\prod_{j=1}^d\mathfrak{s}_h\cdot\mathcal{G}_{M,r,h}(x_j-hn_j)\bigg|\\
    \lesssim_d M^{d/2} M^{2d-1}\exp(-M/2)\|f\|_2 \leq M^{5d/2}\exp(-M/2)\|f\|_2,
\end{multline*}
from which \eqref{approx2conc} ensues. 

For \eqref{approx1conc}, we temporarily fix $\vec{x}\in [-1,1]^d$ and further reduce the notations by setting
\begin{equation*}
    \rho_{hn_{j}}\ast(\mathfrak{s}_h \cdot\mathcal{G}_{M,r,h})''(x_{j}) \mapsto\mathfrak{L}_{n_{j}}(x_j).
\end{equation*}
Recall the $O$-notation introduced in \S\ref{sec:Preliminaries}. Then by \eqref{bound1}, $|{\bf F}_{n_j}(x_j)-\mathfrak{L}_{n_j}(x_j)|=O(2^{-M})$. Hence we can write
\begin{equation*}
    \begin{split}
        \bigg|\prod_{j=1}^{d}{\bf F}_{n_j}(x_j)-\prod_{j=1}^{d}\mathfrak{L}_{n_{j}}(x_j)\bigg| &=\bigg|\prod_{j=1}^{d}(\mathfrak{L}_{n_{j}}(x_j)+O(2^{-M}))-\prod_{j=1}^{d}\mathfrak{L}_{n_{j}}(x_j)\bigg| \\
        &\lesssim_d 2^{-dM} + \sum_{l=1}^{d-1}{d\choose l} M^{2l}2^{(d-l)M} \lesssim_{d} M^{2(d-1)}2^{-M},
    \end{split}
\end{equation*}
where \eqref{allbounded} has been used in the first inequality. Finally, through a calculation that is now routine, we get
\begin{equation*}
    |\Psi^*_f(\vec{x})-\Psi_{f}(\vec{x})| \lesssim_{d}  M^{5d/2}2^{-M}\|f\|_2,
\end{equation*}
implying \eqref{approx1conc}.
\end{proof}

Combining the results of Theorem~\ref{thm:samp}, Lemmas~\ref{lem:diffhigh},~\ref{lem:approx1&2}, and bearing in mind that $\sigma<\pi/(2h)=M\pi/8$, we conclude, for $x\in [-1,1]^d$,
\begin{align}
    \nonumber &|\Psi^*_f(\vec{x})-f(\vec{x})|\\
    \nonumber &\leq |\Psi^*_f(\vec{x})-\Psi_f(\vec{x})| + |\Psi_f(\vec{x})-\widetilde{\mathcal{R}}_{M,r,h}f(\vec{x})| + |\widetilde{\mathcal{R}}_{M,r,h}f(\vec{x})-\mathcal{R}_{M,r,h}f(\vec{x})| + |\mathcal{R}_{M,r,h}f(\vec{x})-f(\vec{x})|\\
    \nonumber &\lesssim_d \Big(M^{5d/2}2^{-M} + M^{5d/2}\exp(-M/2) + M^{d/2}\exp(-M/32) + \sigma^{d/2}\exp(-M/4)\Big)\|f\|_2\\
    \label{Psi*&f} &\lesssim_d M^{5d/2}\exp(-M/32)\|f\|_2,
\end{align}
thereby accomplishing the first task. We now turn to our second task, which is to draft a ReLU GNN that nearly outputs $\Psi^*_f$ across a dense lattice of $[-1,1]^d$. 

\paragraph{GNN realization.} We must devise a neural network architecture capable of encapsulating the product structure in \eqref{def:Psi*}. As a result, we rely on the following lemma, whose proof is detailed in \S\ref{appx:productlemmapf}.

\begin{lemma} \label{lem:product} Let $d\in\mathbb{Z}_+$ such that $d\geq 2$. Let $X_1,\cdots,X_{d}\in\mathbb{R}$ be such that $|X_{j}|\leq T$ for some $T\geq 1$. Let $m\in\mathbb{Z}_{+}$. Then there exists an ReLU NN structure $\Phi_{m,T}(X_1,\cdots,X_d)$ with $O((d-1)m)$ weights and layers, such that
\begin{equation*}
    |\Phi_{m,T}(X_1,\cdots,X_{d})-\prod_{j=1}^{d}X_{j}|\lesssim_{d} T^{d}e^{-m},
\end{equation*}
and moreover, if $X_{j}=0$ for any $j$, then $\Phi_{m,T}(X_1,\cdots,X_{d})=0$.  
\end{lemma}

We now prepare some concepts. We let ${\bf 1}_{n\times m}$ denote an $n\times m$-matrix whose any entry value is $1$, and in addition, if ${\bf M}$ is a matrix, then ${\bf M}^{\mathrm{row}_j}$ denotes the $j$th row of ${\bf M}$. We select our graph to be the lattice $[-5,5]^d \cap\tau\mathbb{Z}^d$, 
and the prediction is done on the sub-lattice $\mathcal{X}:=[-1,1]^d\cap\tau\mathbb{Z}^d$. For each $j=1,\cdots,d$, let ${\bf B}_j:= \begin{bmatrix} \vec{b}_{\vec{x},j} \end{bmatrix}_{\vec{x}\in\mathcal{X}}\in\mathbb{R}^{(M+1)\times |\mathcal{X}|}$ be a \textit{bias matrix}, where 
\begin{equation*} 
    \vec{b}_{\vec{x},j} = \vec{b} := \begin{bmatrix} h(-M/2) \\ h(-M/2+1) \\ \vdots \\ h(M/2-1) \\ h(M/2) \end{bmatrix} \in\mathbb{R}^{(M+1)\times 1}
\end{equation*}
is a bias vector. Next, set $m=M$ and, following \eqref{allbounded}, $T=CM^2$, for some appropriate $C>0$. Let $\vec{x}\in\mathcal{X}:= [-1,1]^d\cap\tau\mathbb{Z}^d$ be temporarily fixed; we define
\begin{equation} \label{Y1}
    Z_1(n_2,\cdots,n_{d};x_2,\cdots,x_{d}):= \Phi_{m,T}\bigg({\bf F}_{n_2}(x_2),\cdots,{\bf F}_{n_d}(x_d)\bigg).
\end{equation}
Then as a result of \eqref{allbounded} and Lemma~\ref{lem:product}, 
\begin{equation} \label{Y1bd}
    |Z_1(n_2,\cdots,n_{d};x_2,\cdots,x_{d})| \lesssim_{d} M^{2(d-1)}.
\end{equation}
We define
\begin{equation} \label{Y2}
    Z_2(n_1;x_2,\cdots,x_{d}) :=\sum_{n_2,\cdots,n_{d}}f(hn_1,\cdots,hn_{d})Z_1(n_2,\cdots,n_{d};x_2,\cdots,x_{d}).
\end{equation}
Then it follows from the definition and \eqref{Y1bd} that
\begin{equation} \label{Y2bd}
    |Z_2(n_1;x_2,\cdots,x_{d})| \lesssim_{d} M^{5(d-1)/2}\|f\|_2.
\end{equation}
Continuing, we now set $m=M$ and $T=C(d)M^{5(d-1)/2}$, for some appropriate $C(d)>0$, apply Lemma~\ref{lem:product} again, to form
\begin{equation} \label{Y3}
    Z_3(n_1;\vec{x}) :=\Phi_{m,T}\bigg(\|f\|_2^{-1}Z_2(n_1;x_2,\cdots,x_{d}), {\bf F}_{n_1}(x_1)\bigg),
\end{equation}
where, by \eqref{allbounded}, \eqref{Y2bd},
\begin{equation} \label{Y3bd}
    |Z_3(n_1;x_2,\cdots,x_{d})|\lesssim_d M^{5(d-1)}.
\end{equation}
After having provided all the necessary concepts, we assume as usual that we have sampled ${\bf S}_f := [f(h\vec{n})]_{\vec{n}\in [-M/2,M/2]^d}\in\mathbb{R}^{1\times (M+1)^d}$. Our ReLU GNN $\tilde{\Psi}_f$ will comprise of 
\begin{equation} \label{totalGNNhighdim}
    \Psi_5\circ\Psi_4\circ\Psi_3\circ\Psi_2\circ\Psi_1,
\end{equation}
when $d>1$, and
\begin{equation} \label{totalGNNdimone}
    \Psi'_2\circ\Psi_1, 
\end{equation}
when $d=1$. Here, $\Psi'_2, \Psi_3, \Psi_5$ are one-layered ReLU NNs, $\Psi_2$ is an $O_d(M^d)$-layered ReLU NN, being $\Phi_{m,T}$ defined in \eqref{Y1}, $\Psi_4$ an $O(M^3)$-layered ReLU NN, being $\Phi_{m,T}$ defined in \eqref{Y3}, and lastly $\Psi_1$ is a one-layer ReLU GNN. The composition notation $\circ$ in \eqref{totalGNNhighdim} signifies that the output of one network serves as the input for the next. The structures of $\Psi_2, \Psi_4$ will become more apparent during the proof of Lemma~\ref{lem:product} in \S\ref{appx:productlemmapf}, and $\Psi'_2, \Psi_3, \Psi_5$ involve only affine transformations. As $\Psi_1$ contains crucial information about our graph filtering process, we will delve into its structure in more detail below, while providing a brief overview of the others. The ensuing discussion draws upon the definitions provided immediately above for contextual reference.

\paragraph{ReLU GNN $\Psi_1$} (one hidden filter layer and only unital weights)

\begin{enumerate}
    \item {\bf Input:} The input ${\bf H}_{0,\Psi_1}\equiv {\bf X}_{\Psi_1}\in\mathbb{R}^{d\times |\mathcal{X}|}$ consists of vectors $\vec{x}\in\mathcal{X}$ as its columns.
    \item {\bf Hidden layer:} Given the input ${\bf H}_{0,\Psi_1}$, the output ${\bf H}_{1,\Psi_1}$ of this layer is given by bias-shifting with respect to the sampling locations, then an application of component-wise ReLU activation, followed by a convolution:
    \begin{equation*} 
        \begin{split}
            {\bf H}_{0,\Psi_1} &\equiv {\bf X}_{\Psi_1}\\ &\mapsto \begin{bmatrix} {\bf 1}_{(M+1)\times 1}{\bf X}^{\mathrm{row}_1}_{\Psi_1} \\ \vdots \\ {\bf 1}_{(M+1)\times 1}{\bf X}^{\mathrm{row}_d}_{\Psi_1} \end{bmatrix} - \begin{bmatrix} {\bf B}_1 \\ \vdots \\ {\bf B}_d \end{bmatrix} \xmapsto{\mathrm{act.}} \rho\Bigg(\begin{bmatrix} {\bf 1}_{(M+1)\times 1}{\bf X}^{\mathrm{row}_1}_{\Psi_1} \\ \vdots \\ {\bf 1}_{(M+1)\times 1}{\bf X}^{\mathrm{row}_d}_{\Psi_1} \end{bmatrix} - \begin{bmatrix} {\bf B}_1 \\ \vdots \\ {\bf B}_d \end{bmatrix}\Bigg) \\
            &\xmapsto{{\bf Filter}} \begin{bmatrix} \tilde{{\bf F}}_{x_1} \\
            \vdots \\ 
            \tilde{{\bf F}}_{x_d} \end{bmatrix}_{\vec{x}\in\mathcal{X}} \equiv {\bf H}_{1,\Psi_1},
        \end{split}    
    \end{equation*}
    where, for $j=1,\cdots,d$,
    \begin{equation} \label{filterinfo}
        \tilde{{\bf F}}_{x_j} := \begin{bmatrix} {\bf F}_{-M/2}(x_j) \\ {\bf F}_{-M/2+1}(x_j) \\ \vdots \\ {\bf F}_{M/2-1}(x_j) \\ {\bf F}_{M/2}(x_j) \end{bmatrix} \in\mathbb{R}^{(M+1)\times 1}.
    \end{equation}
    \item{\bf Output:} The output is ${\bf Y}_{\Psi_1}\equiv {\bf H}_{1,\Psi_1}\in\mathbb{R}^{d(M+1)\times |\mathcal{X}|}$. 
\end{enumerate}

When $d=1$, the output ${\bf Y}_{\Psi_1}\in\mathbb{R}^{(M+1)\times |\mathcal{X}|}$ becomes the input ${\bf X}_{\Psi'_2}$ for $\Psi'_2$, a ReLU NN that implements the linear transformation
\begin{equation*}
    \Psi'_2: \quad {\bf X}_{\Psi'_2} \mapsto {\bf S}_f{\bf X}_{\Psi'_2}.
\end{equation*}
The output $\tilde{\Psi}_f=\Psi'_2\circ\Psi_1$ has been previously shown in \eqref{dimoneapprox} to give robust prediction of $f$ on $[-1,1]\cap\tau\mathbb{Z}$. Additionally, it is apparent that the construction of $\tilde{\Psi}_f$ involves two layers, including one filter layer, and $M+1$ nonunital weights obtained from the sampled values ${\bf S}_f$ of $f$. Note that, when $d>1$, the filtering process illustrated in \eqref{filterinfo} encompasses one-dimensional filters applied to ReLU activations, aligning with the framework presented in \eqref{ourorder}. Further, in this case, we make use of ReLU NNs $\Psi_2, \Psi_4$ to emulate the product structure in \eqref{def:Psi*}. The existence of these NNs is assured by Lemma~\ref{lem:product}.

\paragraph{ReLU NN $\Psi_2$} ($O_d(M^d)$ hidden layers and $O_d(M^d)$ weights)

\begin{enumerate}
    \item {\bf Input:} The input ${\bf X}_{\Psi_2}$ is the output ${\bf Y}_{\Psi_1}\in\mathbb{R}^{d(M+1)\times |\mathcal{X}|}$ of $\Psi_1$.
    \item{\bf Output:} The output is ${\bf Y}_{\Psi_2} \equiv \begin{bmatrix} \tilde{{\bf F}}_{x_1} \\ {\bf M}_{x_2,\cdots,x_d}\end{bmatrix}_{\vec{x}\in\mathcal{X}}\in\mathbb{R}^{((M+1)^{d-1}+M+1)\times|\mathcal{X}|}$, where ${\bf M}_{x_2,\cdots,x_d}\in\mathbb{R}^{(M+1)^{d-1}\times 1}$ with rows indexed by $\{(n_2,\cdots,n_d): n_j=-M/2,\cdots,M/2\}$. The $(n_2,\cdots,n_d)$th row entry of ${\bf M}_{x_2,\cdots,x_d}$ is:
    \begin{equation*}
        [{\bf M}_{x_2,\cdots,x_d}]_{(n_2,\cdots,n_d)} 
        :=Z_1(n_2,\cdots,n_{d};x_2,\cdots,x_{d}).
    \end{equation*}
\end{enumerate}

\paragraph{ReLU NN $\Psi_3$} (one hidden layer and $O_d(M^d)$ weights, derived from the sampled values ${\bf S}_f$ of $f$)

\begin{enumerate}
    \item {\bf Input:} The input ${\bf X}_{\Psi_3}$ is the output ${\bf Y}_{\Psi_2}\in\mathbb{R}^{((M+1)^{d-1}+M+1)\times 1}$ of $\Psi_2$.
    \item{\bf Output:} The output is the matrix ${\bf Y}_{\Psi_3}$
    \begin{equation*}
        {\bf Y}_{\Psi_3} \equiv \begin{bmatrix} \tilde{{\bf F}}_{x_1} \\ {\bf Z}_{x_2,\cdots,x_d}\end{bmatrix}_{\vec{x}\in\mathcal{X}} \in\mathbb{R}^{2(M+1)\times |\mathcal{X}|}
    \end{equation*}
    where ${\bf Z}_{x_2,\cdots,x_d}\in\mathbb{R}^{(M+1)\times 1}$ whose rows indexed by $n_1=-M/2,\cdots,M/2$. The $n_1$th row entry of ${\bf Z}_{x_2,\cdots,x_d}$ is $[{\bf Z}_{x_2,\cdots,x_d}]_{n_1} := Z_2(n_1;x_2,\cdots,x_{d})$.
\end{enumerate}

\paragraph{ReLU NN $\Psi_4$} ($O(M^3)$ hidden layers and $O(M^3)$ weights)

\begin{enumerate}
    \item {\bf Input:} The input ${\bf X}_{\Psi_4}$ is the output ${\bf Y}_{\Psi_3}\in\mathbb{R}^{2(M+1)\times |\mathcal{X}|}$ of $\Psi_3$.
    \item{\bf Output:} The output is
    \begin{equation*}
        {\bf Y}_{\Psi_4} \equiv \begin{bmatrix} {\bf M}_{\vec{x}} \end{bmatrix}_{\vec{x}\in\mathcal{X}} \in \mathbb{R}^{(M+1)\times |\mathcal{X}|}
    \end{equation*}
    where ${\bf M}_{\vec{x}}\in\mathbb{R}^{(M+1)\times 1}$ with rows indexed by $\{n_1: n_1=-M/2,\cdots,M/2\}$. The $n_1$th row entry of ${\bf M}_{\vec{x}}$ is $[{\bf M}_{\vec{x}}]_{n_1}:=Z_3(n_1;\vec{x})$.
\end{enumerate}

\paragraph{ReLU NN $\Psi_5$} (one hidden layer and use $O(M)$ weights)

\begin{enumerate}
    \item {\bf Input:} The input ${\bf X}_{\Psi_5}$ is the output ${\bf Y}_{\Psi_4}\in\mathbb{R}^{(M+1)\times |\mathcal{X}|}$ of $\Psi_4$.
    \item{\bf Output:} The output is ${\bf Y}_{\Psi_5} \equiv \begin{bmatrix} z_{\vec{x}} \end{bmatrix}_{\vec{x}\in\mathcal{X}} \in \mathbb{R}^{1\times |\mathcal{X}|}$
    where $z_{\vec{x}}\in\mathbb{R}^{1\times 1}$ is a scalar that is, 
    \begin{equation} \label{def:lilz}
        z_{\vec{x}} := \sum_{n_1}\|f\|_2Z_3(n_1;\vec{x}).
    \end{equation}
\end{enumerate}

It is evident, through a straightforward summation, that $\tilde{\Psi}_f = \Psi_5\circ\Psi_4\circ\Psi_3\circ\Psi_2\circ\Psi_1$ costs in a total of $O_d(M^d)$ layers, including one filter layer, and $O_d(M^{d})$ weights. We advance to the final task, which entails verifying that the total output $\tilde{\Psi}_f(\vec{x})$ effectively approximates $\Psi^*_f(\vec{x})$ in \eqref{def:Psi*} for $\vec{x}\in\mathcal{X}=[-1,1]^d\cap\tau\mathbb{Z}^d$, for $d>1$, thereby implying a similar approximation of $\tilde{\Psi}_f$ to $f$ on the same domain. From \eqref{def:lilz}, this output is also $z_{\vec{x}}$. All the following calculations can be inferred from definitions \eqref{Y1}, \eqref{Y2}, \eqref{Y3}, \eqref{def:lilz}, as well as \eqref{Y1bd}, \eqref{Y2bd}, \eqref{Y3bd}, in conjunction with the application of Lemma~\ref{lem:product}. First, we have that
\begin{equation*}
    Z_1(n_2,\cdots,n_d;x_2,\cdots,x_d)=\prod_{j=2}^d {\bf F}_{n_j}(x_j) + O_d(M^{2(d-1)}e^{-M}),
\end{equation*}
suggesting
\begin{equation*}
    Z_2(n_1;x_2,\cdots,x_d)=\sum_{n_2,\cdots,n_d} f(hn_1,hn_2,\cdots,hn_d)\prod_{j=2}^d {\bf F}_{n_j}(x_j) + O_d(M^{5(d-1)/2}e^{-M}\|f\|_2).
\end{equation*}
Proceeding further, we obtain 
\begin{equation*}
    Z_3(n_1;\vec{x}) = \|f\|_2^{-1}\sum_{n_2,\cdots,n_d} f(hn_1,hn_2,\cdots,hn_d){\bf F}_{n_1}(x_1)\prod_{j=2}^d {\bf F}_{n_j}(x_j) + O_d(M^{5(d-1)}e^{-M}), 
\end{equation*}
and subsequently,
\begin{equation*}
    \tilde{\Psi}_f(\vec{x}) = z_{\vec{x}} = \sum_{\vec{n}\in [-M/2,M/2]^d} f(h\vec{n})\prod_{j=1}^d {\bf F}_{n_j}(x_j) + O_d(M^{5d}e^{-M}).
\end{equation*}
Referring back to \eqref{Psi*&f}, we can gather from the finding above that
\begin{equation*}
    |\tilde{\Psi}_f(\vec{x})-f(\vec{x})|\leq |\tilde{\Psi}_f(\vec{x})-\Psi^*_f(\vec{x})| + |\Psi^*_f(\vec{x})-f(\vec{x})|\lesssim_d M^{5d}\exp(-M/32)\|f\|_2,
\end{equation*}
and conclude Theorem~\ref{thm:actualGNN} for all dimensions. \qed

\section{Discussion} \label{sec:Discussion}

Our exploration into the expressiveness of GNNs has produced profound insights. Leveraging sampling theory, we devise a GNN capable of generalizing bandlimited functions over Euclidean centered unit cubes, showcasing performance comparable to the documented performance of traditional NNs but with fewer required weights. This GNN construction, grounded in an interpolation process using sampled functional values, is notable for its explicitness, offering a pre-trained network that circumvents additional learning steps. Furthermore, by realizing a regularized Whittaker-Kotel'nikov-Shannon sampling principle within the GNN framework, our approach enriches the theoretical toolkit for neural network construction, providing an alternative to methods like Taylor and Legendre approximations. Overall, our work contributes to the comprehension of GNN applications in function approximation and introduces a fresh perspective at the intersection of machine learning and mathematical analysis, with implications for both theoretical and applied aspects of neural network research. 


\section*{Acknowledgments}

AMN is supported by the Austrian Science Fund (FWF) Project P-37010. 
AMN was also supported by the NIH grant R01DE026728. 
YX was partially supported by the NIH grant U01DE029255 and the NSF grant IOS2107215.

AMN would like to express gratitude for the enlightening conversation on GNNs with Haitz Saez de Ocariz Borde.



\appendix 

\section{Basic sampling theory and proof of Lemma~\ref{lem:totaleng}} \label{appx:sampling} 

To provide context, we establish some groundwork. Consider a locally compact abelian group $G$ \cite[Chapters~VI, VII]{loomis2013introduction}, written additively, which also admits a multiplicative operation. Let $0,1$ be the additive and multiplicative identities, respectively. Let $\hat{G}$ be the dual of $G$, i.e., $G$ comprises group homomorphisms $\gamma: G\to\mathbb{C}$:
\begin{equation} \label{pair}
    \gamma: x\mapsto (\gamma,x).
\end{equation}
Here, $(\gamma,x)$ denotes the dual pairing between $G,\hat{G}$ \cite[Chapter VII, \S 34B]{loomis2013introduction}. Let $H\subset G$ be a discrete subgroup, and let $H^{\perp}\subset\hat{G}$ be the discrete annihilator of $H$:
\begin{equation*}
    H^{\perp} :=\{\gamma\in\hat{G}: (\gamma,h)=1, \forall h\in H\}.
\end{equation*}
Let $m_{X}$ denote the Haar measure on a group $X$. As $H$ is discrete, $\hat{G}/H^{\perp}$ is compact \cite[Theorem~1.2.5]{rudin2017fourier}, we can normalize $m_{\hat{G}/H^{\perp}}$ so that it has total measure one. Discrete groups such as $H, H^{\perp}$ are given counting measures. We also normalize $m_{\hat{G}}$ so that the following version of Fubini's theorem holds
\begin{equation*}
    \int_{\hat{G}} =\int_{\hat{G}/H^{\perp}}\int_{H^{\perp}},
\end{equation*}
or equivalently, as $m_{H^{\perp}}(\{\lambda\})=1$,
\begin{equation} \label{FUB}
    \int_{\hat{G}}f(\gamma)\,dm_{\hat{G}} = \int_{\hat{G}/H^{\perp}}\sum_{\lambda\in H^{\perp}}f(\gamma+\lambda)\,dm_{\hat{G}/H^{\perp}}.
\end{equation}
Finally, we normalize $m_{G}$ so that the Fourier transform and its inversion,
\begin{equation*}
    \hat{f}(\gamma) =\int_{G}f(x)(\gamma,-x)\,dm_{G} \quad\text{ and }\quad
    f(x) =\int_{\hat{G}}\hat{f}(\gamma)(\gamma,x)\,dm_{\hat{G}}
\end{equation*}
provide a unitary equivalence between $L^2(G)$ and $L^2(\hat{G})$. Then as a consequence, \cite[Chapter VI, \S 37E]{loomis2013introduction}
\begin{equation*}
    \int_{H}f\,dm_{H} =\int_{H^{\perp}}\hat{f}\,dm_{H^{\perp}}
\end{equation*}
which is needed for the upcoming theorem. Let $\Omega\subset\hat{G}/H^{\perp}$. One wants to envision that $\hat{G}/H^{\perp}$ ``bounds" $\Omega$; in a sense, $\{\gamma+H^{\perp}:\gamma\in\hat{G}\}\cap\Omega$ is a singleton set. Another way to state this is, let $\pi: \hat{G}\to\hat{G}/H^{\perp}$ be the canonical surjection, then $\pi|_{\Omega}$ is a bijection. We define the following reconstruction function
\begin{equation*}
    \varphi(x):=\int_{\hat{G}/H^{\perp}}(\gamma,x)\,dm_{\hat{G}}.
\end{equation*}
Clearly, $\varphi$ has interpolatory properties: $\varphi(x)=0$ if $0\not=x\in H$ and $\varphi(0)=1$. In this framework, the ensuing theorem is confirmed.

\begin{theorem}[Kluv\'anek sampling theorem] \label{Ksampthm} \cite{kluvanek1965sampling} (see also \cite[Theorem~1.5]{benedetto2012modern}) Let $f\in L^2(G)$ and let $\hat{f}$ be null outside $\Omega$. Then $f$ is a.e. equal to a continuous function which can be written as,
\begin{equation*} 
    f(x)=\sum_{y\in H}f(y)\varphi(x-y)
\end{equation*}
and the convergence is both uniform on $G$ and in the norm of $L^2(G)$, and the following Gaussian quadrature holds:
\begin{equation} \label{Kquad}
    \|f\|^2=\sum_{y\in H}|f(y)|^2.
\end{equation}    
\end{theorem}

Continuing, we demonstrate that Theorem~\ref{parzensampthm} can be derived from Theorem~\ref{Ksampthm}. Let $G=\mathbb{R}^{d}$. Then $\hat{G}=\widehat{\mathbb{R}^{d}}=\mathbb{R}^{d}$. Let the dual pairing \eqref{pair} between $\vec{x}\in\mathbb{R}^{d},\vec{y}\in\widehat{\mathbb{R}^{d}}$ be as follows
\begin{equation*}
    (\vec{y},\vec{x}) := e^{i\vec{x}\cdot\vec{y}}.
\end{equation*}
Let $\{\vec{v}\}_{j=1}^d$ be a basis for $\mathbb{R}^{d}$ and let $\{\vec{u}_{k}\}_{k=1}^d$ be its biorthogonal basis so that ${\vec{v}}_{j}\cdot \vec{u}_{k}=2\pi\delta_{jk},j\not=k$. Let, 
\begin{equation*}
    H=\{{\vec{v}}[\vec{n}]:\vec{n}\in\mathbb{Z}^{d}\} :=\{\sum_{j=1}^{d} n_{j}\vec{v}_{j}:\vec{n}\in\mathbb{Z}^{d}\}
\end{equation*}
be the sampling lattice generated by the ${\vec{v}}_{j}$'s. Evidently, $H$ is a discrete subgroup of $\mathbb{R}^{d}$ and $\mathbb{R}^{d}/H$ tessellates $\mathbb{R}^{d}$. The annihilating lattice $H^{\perp}$ of $H$ is
\begin{equation*}
    H^{\perp}=\{\vec{u}[\vec{n}]:\vec{n}\in\mathbb{Z}^{d}\} :=\{\sum_{j=1}^{d} n_{j}\vec{u}_{j}:\vec{n}\in\mathbb{Z}^{d}\}.
\end{equation*}
Set $\Omega=[-\sigma,\sigma]^{d}=\mathbb{R}^{d}/H^{\perp}$. We can dictate $\vec{u}_{j}$'s to be the vectors defining the edges of $[0,2\sigma]^{d}\cong\Omega$ - that means, $\vec{v}[\vec{n}] = \sum_{j=1}^{d}(\pi/\sigma)n_{j}\vec{e}_{j}$ - and the reconstruction function is 
\begin{equation*}
    \varphi(\vec{x})=\frac{1}{(2\sigma)^{d}}\int_{[-\sigma,\sigma]^{d}} e^{i \vec{x}\cdot\vec{t}}\,d\vec{t}.
\end{equation*}
Following Theorem~\ref{Ksampthm} and definition \eqref{sinc}, we obtain
\begin{equation} \label{hdsamp}
    f(\vec{x}) =\frac{1}{(2\sigma)^{d}}\sum_{\vec{y}\in H} f(\vec{y})\int_{[-\sigma,\sigma]^{d}}e^{i(\vec{x}-\vec{y})\cdot \vec{t}}\,d\vec{t} =\sum_{\vec{n}\in\mathbb{Z}^{d}}f((\pi/\sigma)\vec{n})\prod_{j=1}^{d}\frac{\sin(\sigma x_{j}-n_{j}\pi)}{(\sigma x_{j}-n_{j}\pi)},
\end{equation}
which is \eqref{parzen}, where the convergence is uniformly on $\mathbb{R}^{d}$ and in $L^2(\mathbb{R}^{d})$. 

\begin{proof}[Proof of Lemma~\ref{lem:totaleng}]
Lemma~\ref{lem:totaleng} is a straightforward corollary of Theorem~\ref{Ksampthm}. Specifically, by utilizing \eqref{normcompare}, \eqref{Kquad}, \eqref{hdsamp}, we derive
\begin{equation*} \label{three}
    \frac{1}{(2\pi)^{d}}\|\hat{f}\|_2^2=\|f\|_2^2=\sum_{\vec{n}\in\mathbb{Z}^{d}}|f(\vec{n}\pi/\sigma)|^2,
\end{equation*}
which is \eqref{Gaussquad}, as wanted.
\end{proof}

\subsection{Proof of Lemma~\ref{lem:intbyparts}} \label{sec:intbypartslemma}

We begin by citing the following theorem, which will be essential for our proof.

\begin{theorem}[Abel's identity] \cite[Theorem~4.2]{apostol1998introduction} \label{Abel}
Let $\{a(n)\}_{n\geq 0}$ be a real-valued sequence. Define, 
\begin{equation} \label{A} 
    A(x)=\sum_{n\leq x}a(n),
\end{equation}
and $A(x)=0$ if $x<0$. Then for every $g\in C^1[u,v]$, one has,
\begin{equation} \label{abel}
    \sum_{u<n\leq v}a(n)g(n)=A(v)g(v)-A(u)g(u)-\int_{u}^{v}A(x)g'(x)\,dx.
\end{equation}  
\end{theorem}

To apply Theorem~\ref{Abel}, we let $\{a(n)\}_{n\geq 0}$ be such that $a(0) = 1$ and $a(n)=0$ if $n\not= 0$. Define $A(x)$ as in \eqref{A}. Evidently,
\begin{equation} \label{ourA} 
    A(x)=\chi_{\{x\geq 0\}}(x).
\end{equation}
For $t\in\mathbb{R}$, define $\phi_{t-X}(x):=\phi(t-X-x)$. Let $u<0<v$. It then follows from \eqref{abel}, \eqref{ourA} that
\begin{align} 
    \nonumber \sum_{u<n\leq v}a(n)\phi_{t-X}(n) &=A(v)\phi_{t-X}(v)-\int_0^{v}\phi_{t-X}'(x)\,dx\\
    \label{1st}\Rightarrow\phi(t-X) &= \phi(t-X-v)-\int_0^{v}(d/dx)\phi(t-X-x)\,dx.
\end{align}
Note that $\mathrm{ReLU}(x)=x$ when $x>0$, implying $(d/dx)(\mathrm{ReLU}(x))=1$ on $(0,\infty)$. Hence
\begin{equation} \label{2nd} 
    \int_0^{v}(d/dx)\phi(t-X-x)\,dx =-\int_0^{v}(d^2/dx^2)\phi(t-X-x)x\,dx + (d/dx)\phi(t-X-x)x|_{x=v}.
\end{equation}
Combining \eqref{1st}, \eqref{2nd} leads to
\begin{equation} \label{3rd} 
    \phi(t-X) = \phi(t-X-v)+\int_0^{v}(d^2/dx^2)\phi(t-X-x)x\,dx - (d/dx)\phi(t-X-x)x|_{x=v}.
\end{equation}
Take $v\to\infty$ in \eqref{3rd}. By employing \eqref{lemm5cond} and Lebesgue's dominated convergence theorem \cite[Theorem~2.24]{folland1999real}, we get
\begin{equation*} 
    \delta_{X}\ast\phi(t)=\int_{\mathbb{R}}\phi(t-x)\,d\delta_{X}(x)=\phi(t-X)=\mathrm{ReLU}\ast\phi''(t-X)=\mathrm{ReLU}_{X}\ast\phi''(t),
\end{equation*}
which is the conclusion of the lemma. \qed

\subsection{Proof of Lemma~\ref{lem3}} \label{5.5}

\begin{figure}
    \centering
    \begin{minipage}{0.5\textwidth}
        \centering
        \includegraphics[width=0.99\textwidth]{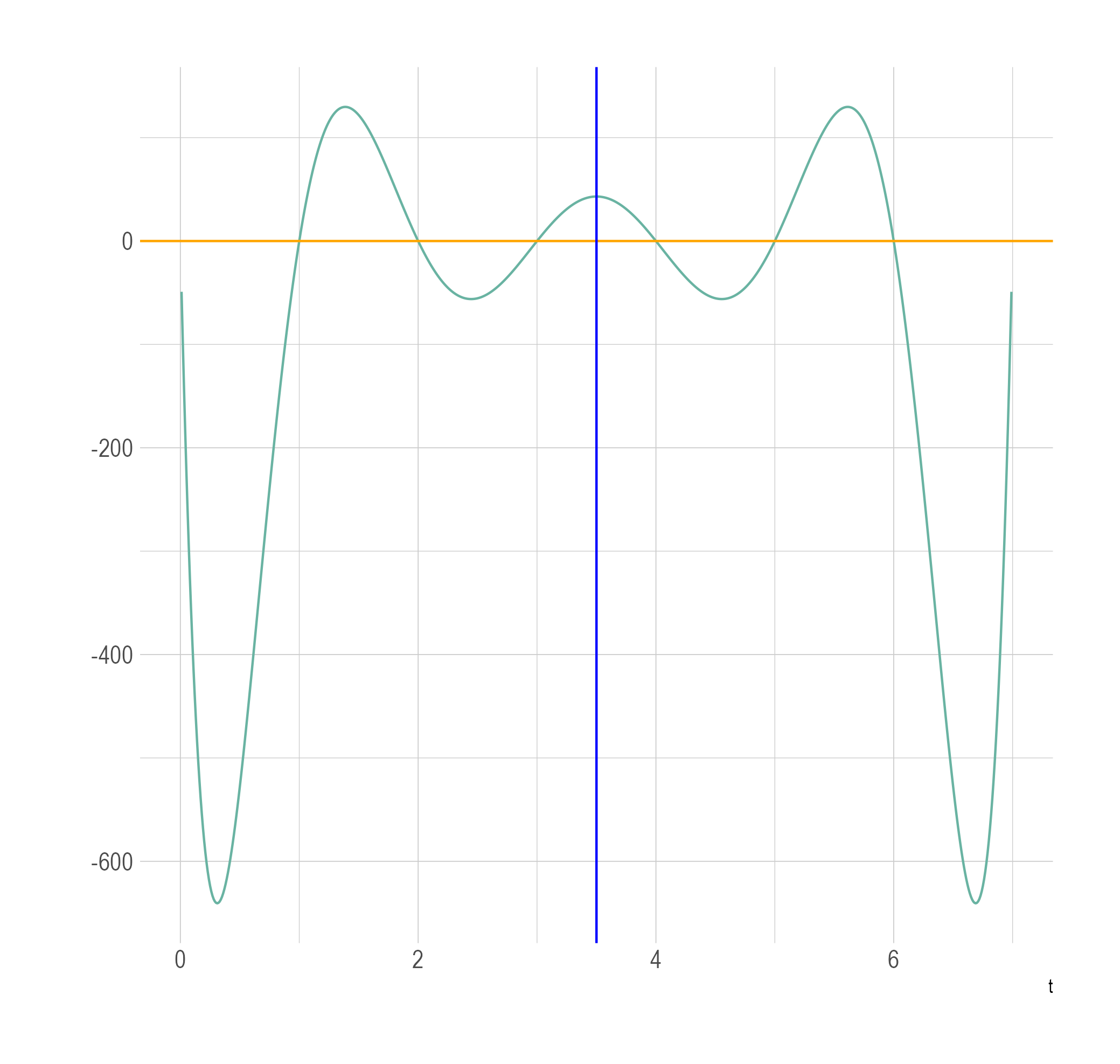}
        \caption{\small The graph of $\pi_7(t)$ on $[0,7]$ }
        \label{fig:L7}
    \end{minipage}\hfill
    \begin{minipage}{0.5\textwidth}
        \centering
        \includegraphics[width=0.99\textwidth]{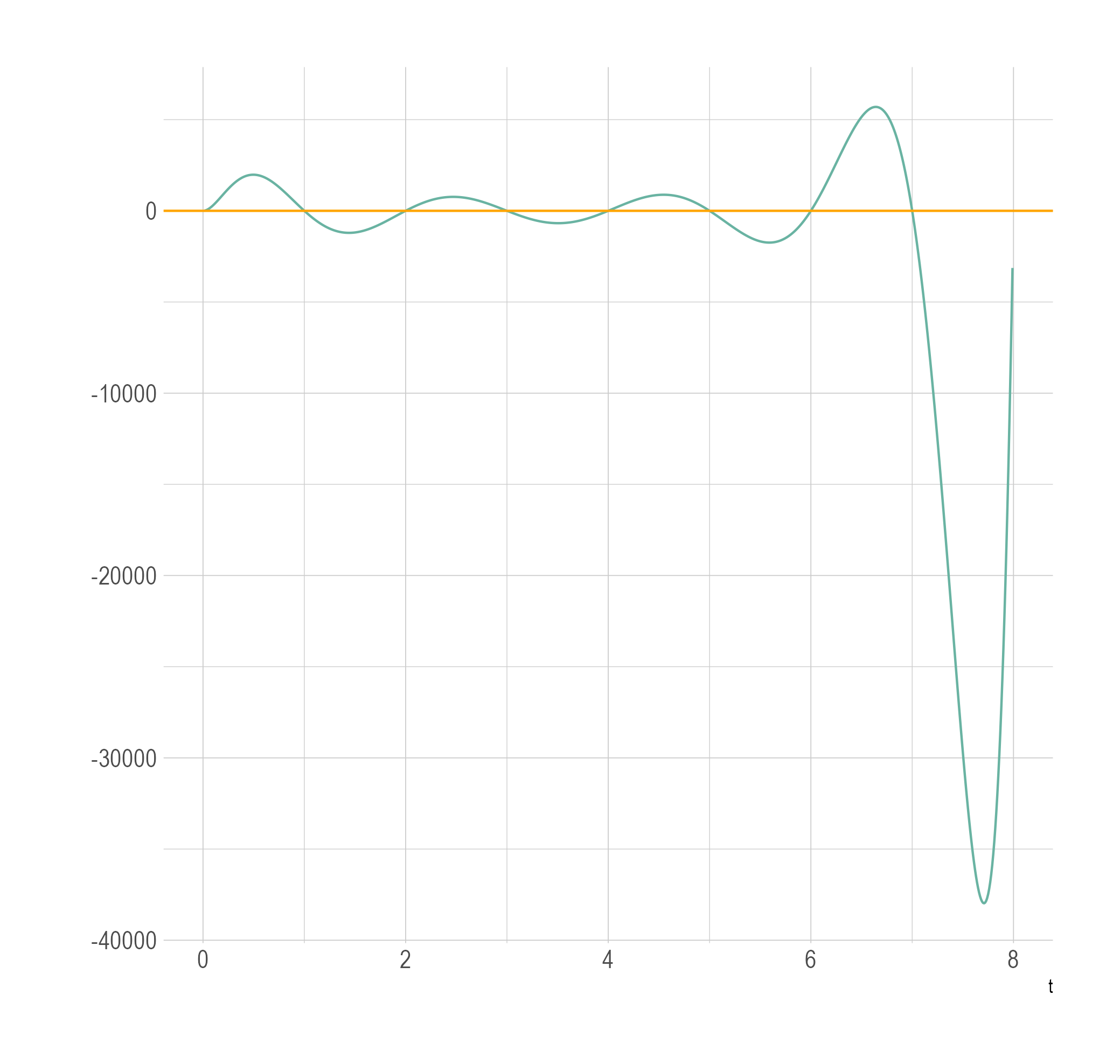}
        \caption{\small The graph of $t\pi_8(t)$ on $[0,8]$. }
        \label{fig:L8}
    \end{minipage}
\end{figure}

We want to prove the following statements
\begin{align}
    \label{firstpart} \text{if } L \text{ is even}, \quad &\bigg|\int_0^{L}u^2(u-1)(u-2)\cdots (u-L)\,du\bigg|\lesssim (L+1)!\\
    \label{secondpart} \text{if } L \text{ is odd}, \quad &\bigg|\int_0^{L}u(u-1)(u-2)\cdots (u-L)\,du\bigg|\lesssim L!.
\end{align} 
Before proceeding with the proof, we give a brief discussion about the functions
\begin{equation*}
    \pi_{L}(t) := t(t-1)(t-2)\cdots (t-L),
\end{equation*}
where $L\in\mathbb{Z}_{+}$. These are called \textit{factorial polynomials} in the literature \cite[Chapter 6, Subsection 3.1]{isaacson2012analysis}. A fair amount is known about them. For instance, \cite[Theorem~9.2]{quarteroni2010numerical}
\begin{equation} \label{factint}
    \text{if } L \text{ is odd},\quad \int_0^{L}\pi_{L}(t)\,dt<0
    \quad\text{ and }\quad\text{ if } L \text{ is even},\quad \int_0^{L}t\pi_{L}(t)\,dt<0.
\end{equation}
These functions possess a symmetry around $L/2$:
\begin{equation} \label{symmetry}
    \text{if } L \text{ is odd},\,\, \pi_{L}\bigg(\frac{L}{2}-t\bigg) =\pi_{L}\bigg(\frac{L}{2}+t\bigg) \,\,\text{ and }\,\,
    \text{if } L \text{ is even},\,\, \pi_{L}\bigg(\frac{L}{2}-t\bigg) =-\pi_{L}\bigg(\frac{L}{2}+t\bigg).
\end{equation}
This means that the behavior observed for $\pi_{L}$ on $[0,L/2]$ can be used to predict the one on $[L/2,L]$. We also know that for a non-integer $t$,
\begin{equation} \label{monotone}
    \text{ if } 0< t+1\leq L/2,\,\, |\pi_{L}(t+1)|<|\pi_{L}(t)|\quad\text{ and }\quad
    \text{ if } L/2\leq t<L,\,\, |\pi_{L}(t)|<|\pi_{L}(t+1)|.
\end{equation}
\eqref{symmetry} and \eqref{monotone} can be found in \cite[Chapter~6, Section~3, Lemma~1]{isaacson2012analysis} and \cite[Chapter~6, Section~3, Lemma~2]{isaacson2012analysis}, respectively. In light of \eqref{factint}, we see that when $L$ is odd, \eqref{secondpart} is equivalent to
\begin{equation} \label{oddoneside}
    -L!\lesssim\int_0^{L}\pi_{L}(t)\,dt<0.
\end{equation}
We proceed to prove \eqref{oddoneside}. Let $k\in\mathbb{N}$. It is easy to check that for $L$ odd and $2k+1\leq L$,
\begin{equation} \label{oddvalley}
    \pi_{L}\leq 0\,\,\text{ on }\,\, [2k, 2k+1] \quad\text{ and }\quad \pi_{L}\geq 0\,\,\text{ on }\,\, [2k+1, 2k+2].
\end{equation}
The graph of $\pi_{L}$ draws out an $L$ odd number of ``humps" over integer intervals $[k,k+1]$, where $k$ runs from $0$ to $L-1$, creating $\lceil L/2\rceil$ negative humps and $L-\lceil L/2\rceil$ positive humps; see Figure~\ref{fig:L7}. Let $\mathcal{A}_{k}:=\int_{k}^{k+1}\pi_{L}(t)\,dt$ be the integral area of $\pi_L$ over one such integer interval $[k,k+1]$; for simplicity, we have omitted the dependence on $L$ in the notation $\mathcal{A}_k$. It follows from \eqref{oddvalley} that
\begin{equation*}
    \mathcal{A}_{2k}\leq 0, \quad\text{ and }\quad \mathcal{A}_{2k+1}\geq 0,
\end{equation*}
for nonnegative integers $2k, 2k+1\leq L$. Our strategy is to assess the total contribution of the negative areas to the final integral, considering the positive ones can only help. From \eqref{monotone}, if $k\geq\lceil L/2\rceil$ and $k$ is even, then, $-\mathcal{A}_{k}=|\mathcal{A}_{k}|<|\mathcal{A}_{k+1}|=\mathcal{A}_{k+1}$. This means, if we sums all the signed areas $\mathcal{A}_{k}$'s, with $k\geq\lceil L/2\rceil$ up to $L-2$, we obtain
\begin{equation*}
    \int_{\lceil L/2\rceil}^{L-1}\pi_{L}(t)\,dt=\sum_{k=\lceil L/2\rceil}^{L-2}\mathcal{A}_{k}>0.
\end{equation*}
Similar argument and symmetry \eqref{symmetry} show that $\sum_{k=1}^{\lceil L/2\rceil -1}\mathcal{A}_{k}>0$. All that is left is $\mathcal{A}_0=\mathcal{A}_{L-1}<0$ and $\mathcal{A}_{\lceil L/2\rceil-1}$. Through \eqref{monotone} again, it is easy to see that $|\mathcal{A}_{\lceil L/2\rceil-1}|<|\mathcal{A}_{L-1}|=-\mathcal{A}_{L-1}$. A conservative estimate shows that, 
\begin{equation*}
    -\mathcal{A}_{L-1}=\int_{L-1}^{L}t(t-1)(t-2)\cdots (t-L+1)(L-t)\,dt\leq L!\int_{L-1}^{L}(t-L+1)(L-t)\,dt=\frac{5L!}{6},
\end{equation*}
which means $\mathcal{A}_{L-1}\gtrsim -L!$. Hence
\begin{equation*}
    \int_0^{L}\pi_{L}(t)\,dt = \sum_{k=0}^{L-1}\mathcal{A}_{k}\geq 3\mathcal{A}_{L-1}\gtrsim -L!,
\end{equation*}
and we are done with the second part of Lemma~\ref{lem3}.

Continuing our proof, we turn to the first part of Lemma~\ref{lem3}, which is expressed by \eqref{firstpart} and, by \eqref{factint}, can be equivalently stated as
\begin{equation} \label{evenoneside}
    -(L+1)!\lesssim\int_0^{L}t\pi_{L}(t)\,dt <0.
\end{equation}
Note that we no longer have any sort of symmetry or monotonic behaviors with the function $t\pi_{L}(t)$. However, \eqref{factint} holds, and more importantly, \eqref{symmetry}, \eqref{monotone} remain valid for $\pi_{L}(t)$ when $L$ is even, allowing us to employ the same strategy of summing up areas as before. Observe that, when $L$ is even and $2k+2\leq L$, $k\in\mathbb{N}$,
\begin{equation} \label{evenvalley}
    t\pi_{L}(t)\geq 0\,\,\text{ on }\,\, [2k, 2k+1] \quad\text{ and }\quad t\pi_{L}(t)\leq 0\,\,\text{ on }\,\, [2k+1, 2k+2].
\end{equation}
Let $\mathcal{B}_{k} :=\int_{k}^{k+1}t\pi_{L}(t)\,dt$. Then from \eqref{evenvalley},
\begin{equation*}
    \mathcal{B}_{2k}\geq 0, \quad\text{ and }\quad \mathcal{B}_{2k+1}\leq 0,
\end{equation*}
for nonnegative integers $2k, 2k+1\leq L$. Moreover, $t\pi_{L}(t) = t(t-L)\pi_{L-1}(t)$. We examine the contribution of
\begin{equation*}
    \int_0^{L-1} t(t-L)\pi_{L-1}(t)\, dt=\int_1^{L-1}t\pi_{L}(t)\,dt =\sum_{k=1}^{L-2}\mathcal{B}_{k}
\end{equation*}
to the total area $\int_0^{L}t\pi_{L}(t)\,dt$. Write,
\begin{align} 
    \nonumber \int_0^{L-1}t(t-L)\pi_{L-1}(t)\, dt &=\int_{\mathrm{neg}}t(t-L)\pi_{L-1}^{-}(t)\,dt + \int_{\mathrm{pos}} t(t-L)\pi_{L-1}^{+}(t)\,dt\\
    \label{areadecomp} &=-\int_{\mathrm{neg}} t(L-t)\pi_{L-1}^{-}(t)\,dt -\int_{\mathrm{pos}} t(L-t)\pi_{L-1}^{+}(t)\,dt,
\end{align}
where $\pi_{L-1}^{+}$ denotes the positive part of $\pi_{L-1}$ and $\pi_{L-1}^{-}$ the negative part of $\pi_{L-1}$, and
\begin{equation*}
    \mathrm{neg}:=\{t\in [0,L-1]:\pi_{L-1}(t)\leq 0\}\quad\text{ and }\quad \mathrm{pos}:=\{t\in [0,L-1]:\pi_{L-1}(t)\geq 0\}.
\end{equation*}
It is evident from \eqref{areadecomp}, the negative area contribution comes from $-\int_{\mathrm{pos}} t(L-t)\pi^{+}_{L-1}(t)\,dt$, and from \eqref{oddvalley}, 
\begin{equation} \label{keyobs}
    \mathrm{pos}=\bigcup_{k\in\mathbb{N}} [2k+1,2k+2],
\end{equation}
where $2k+2\leq L-2$. Now the two functions $t(L-t)$ and $\pi_{L-1}(t)$ have different symmetry axes at $t=L/2, t=(L-1)/2$, respectively. To reconcile this, we upper-bound
\begin{equation} \label{keyint}
    \int_{\mathrm{pos}} t(L-t)\pi^{+}_{L-1}(t)\,dt \leq 2\int_{\mathrm{pos}} t(L-1-t)\pi^{+}_{L-1}(t)\,dt,
\end{equation}
which is possible due to the description of $\mathrm{pos}$ in \eqref{keyobs}. We can now make use of \eqref{symmetry}; we assert that the last integral in \eqref{keyint} is no more than
\begin{equation*}
    \begin{split}
        2\sum_{k=\lceil (L-1)/2\rceil}^{L-3} 2\int_{k}^{k+1}t(L-1-t)|\pi_{L-1}(t)|\,dt &= 4\sum_{k=L/2}^{L-3}\int_{k}^{k+1}t(L-1-t)|\pi_{L-1}(t)|\,dt\\
        & \leq (2/3)\sum_{k=L/2}^{L-3} (k+1)(k+1)!(L-k)(L-k)!.
    \end{split} 
\end{equation*}
Direct calculations reveal that among the terms in the final sum above, the term with $k=L-3$ dominates. Hence
\begin{equation*}
    4\sum_{k=\lceil (L-1)/2\rceil}^{L-3}\int_{k}^{k+1}t(L-1-t)|\pi_{L-1}(t)|\,dt\lesssim L(L-2)(L-2)!\lesssim L!,
\end{equation*}
and so the total negative area in the interval $[0,L-1]$ cannot be less than,
\begin{equation} \label{L-1}
    -\int_{\mathrm{pos}} t(L-t)\pi_{L-1}^{+}(t)\,dt\gtrsim -L!.
\end{equation}
For the final interval $[L-1, L]$, a straightforward calculation yields
\begin{equation} \label{L}
    \int_{L-1}^{L} t|\pi_{L}(t)|\,dt =\int_{L-1}^{L}t^2(t-1)(t-2)\cdots (t-L+1)(L-t)\,dt\lesssim L\cdot L!.
\end{equation}
By combining \eqref{L-1}, \eqref{L}, we obtain \eqref{evenoneside}, which is \eqref{firstpart}, as desired. \qed

\begin{remark}
Experimental results obtained using MATLAB demonstrate that, for $L\in 2\mathbb{Z}_{+}$,
\begin{equation*}
    0>\int_0^{L}t\pi_{L}(t)\,dt\geq -5\cdot L!.
\end{equation*}
Hence if one only needs $L\leq 170$, perhaps the order $O(L!)$ is enough.
\end{remark}

\subsection{Proof of Lemma~\ref{lem:product}} \label{appx:productlemmapf}

Let $m, T$ be as in the premise of the lemma; in what follows, we simplify the notations by suppressing dependence on $m,T$. 

The case of $d=2$ for the lemma was established in \cite[Proposition~3]{yarotsky2017error}, or rather, its proof. Specifically, there exists a ReLU neural network $\widetilde{\times}_2$, such that if $|X|,|Y|\leq T$, then
\begin{equation} \label{d2}
    |\widetilde{\times}_2(X,Y)-XY|\lesssim T^2e^{-m}.
\end{equation}
Moreover, if $XY=0$ then so is $\widetilde{\times}_2(X,Y)$. Importantly, the numbers of weights and layers of $\widetilde{\times}_2$ are of order $O(m)$.

Now suppose $d>2$, and consider $|X_1|,\cdots,|X_{d}|\leq T$. We define
\begin{equation*}
    \Phi(X_1,\cdots,X_{d}):= \widetilde{\times}_2(X_{d},\widetilde{\times}_2(X_{d-1},\widetilde{\times}_2(\cdots\widetilde{\times}_2(X_2,X_1)))).
\end{equation*}
It is evident that the construction of $\Phi$ uses $O((d-1)m)$ weights and layers, and if $X_{j}=0$ for any $j$, then $\Phi(X_1,\cdots,X_{d})=0$. Further, repeated applications of \eqref{d2} reveal that
\begin{equation*} 
    \begin{split}
        \Phi(X_1,\cdots,X_{d}) &=\widetilde{\times}_2(X_{d},\widetilde{\times}_2(X_{d-1},\widetilde{\times}_2(\cdots\widetilde{\times}_2(X_2,X_1))))\\
        &=\prod_{j=1}^{d}X_{j} + O\bigg(e^{-m}\sum_{j=0}^d T^j\bigg) = \prod_{j=1}^{d}X_{j} + O_d(T^d e^{-m}),
    \end{split}
\end{equation*}
concluding the lemma. \qed

\end{document}